\pdfoutput=1
\documentclass[twoside,11pt]{article}

\usepackage{jmlr2e}
\usepackage{lmodern}
\usepackage[T1]{fontenc}
\usepackage{graphicx}
\usepackage{amssymb}
\usepackage{amsmath}
\usepackage[colorlinks,linkcolor=blue,citecolor=blue,urlcolor=blue]{hyperref}
\usepackage{booktabs}
\usepackage{framed}

\newif\ifworkingpaper
\workingpapertrue 
\ifworkingpaper
 \makeatletter
 \def\jmlrheading#1#2#3#4#5#6{\def\ps@jmlrtps{\let\@mkboth\@gobbletwo%
    \def\@oddhead{\scriptsize \@date \hfill}%
    \def\@oddfoot{\scriptsize \copyright #2 #6. \hfill}%
    \def\@evenhead{}\def\@evenfoot{}}%
  \thispagestyle{jmlrtps}}
 \def\@starteditor{\def\@editor{}%
    \begin{tabular}{@{}ll}
      {Initial version:} & September 12th, 2012.\\
      {This version:} & \@date.
    \end{tabular}}
 \def\@endeditor{}
 \makeatother
\fi

\graphicspath{{figures/}{}}

\let\bottomrule=\midrule
\def\eat#1{}
\def\eg.{\mbox{e.}\mbox{g.}}
\def\ie.{\mbox{i.}\mbox{e.}}
\def\argmax{\mathop{\rm arg\,max}}
\def\ArgMax{\mathop{\rm Arg\,Max}}

\def\calA{{\cal A}}
\def\calI{{\cal I}}
\def\calL{{\cal L}}
\def\calR{{\cal R}}
\def\d{{\mathrm{d}}}
\def\b{{b_{\hskip-.07em\star}}}
\def\bbbr{{\mathbb{R}}}

\def\bbbe{{\mathbb{E}}}
\def\bbbp{{\mathbb{P}}}
\def\bbbone{{\mathrm{1{\hskip-.26em}l}}}
\def\var{{\mathrm{var}}}
\def\cov{{\mathrm{cov}}}
\def\inBrace#1{{\left\{{#1}\right\}}}
\def\inBrack#1{{\left[{#1}\right]}}
\def\inPar#1{{\left({#1}\right)}}
\def\inCond#1#2{\!\left(\left.{#1}\,\right|{#2}\right)}
\def\P{{\rm P}}
\def\bbbP{{\text{\large$\bbbp$}}}
\def\Pa#1{{\P\inPar{#1}}}
\def\Pc#1#2{{\P\inCond{#1}{#2}}}
\def\Qa#1{{\P^*\inPar{#1}}}
\def\Qc#1#2{{\P^*\inCond{#1}{#2}}}
\def\normalDist#1#2{{{\cal N}({#1},{#2})}}
\def\lognormalDist#1#2{{{\it ln\cal N}({#1},{#2})}}

\def\pd#1#2{\frac{\partial{#1}}{\partial{#2}}}
\def\pdd#1#2#3{\frac{\partial^2{#1}}{\partial{#2}\,\partial{#3}}}

\def\pprime{{\prime\prime}}
\def\capw{\bar{w}}
\def\capY{\bar{Y}}
\def\estY{\widehat{Y}}
\def\capV{\bar{V}}
\def\estV{\widehat{V}}
\def\capW{\bar{W}}
\def\estW{{\hskip.2em\widehat{\phantom{Y}}\hskip-1em}W}
\def\capwz{{\bar{w}^{\text{\sc z}}_\theta}}
\def\capwm{{\bar{w}^{\text{\sc m}}_\theta}}
\def\capwmp{{\bar{w}^{\text{\sc m}^\prime}_\theta}}
\def\thetastar{{\theta^*\!}}
\def\user{{\text{\sf user}}}
\def\gbar{{\bar{g}}}

\def\d{{\rm d}}

\newenvironment{sem}{%
  \bgroup\boldmath
}{%
  \par\vspace{-1.5ex}
  \egroup}

\def\TheTitle{Counterfactual Reasoning and Learning Systems}
\def\TheAuthors{L\'eon Bottou, Jonas Peters, et al.}

\jmlrheading{}{}{}{10/2012}{?}{\TheAuthors}
\ShortHeadings{\TheTitle}{Bottou, Peters, et al.}
\firstpageno{1}

\begin{document}

\title{\TheTitle}

\author{\protect\setcounter{footnote}{1}\relax
        \name L\'eon Bottou\\ 
        \email leon@bottou.org\\
        \addr Microsoft Research, Redmond, WA.
        \AND
        \name Jonas Peters\thanks{%
              Jonas Peters has moved to ETH Z\"urich.}\\
        \email jonas.peters@tuebingen.mpg.de\\
        \addr Max Planck Institute, T\"ubingen.
        \AND
        \name Joaquin Qui\~{n}onero-Candela,$^{a}$\thanks{%
              Joaquin Qui\~{n}onero-Candela has joined Facebook.}\,
              Denis X. Charles,$^b$
              D. Max Chickering,$^b$\\
              Elon Portugaly,$^a$
              Dipankar Ray,$^c$
              Patrice Simard,$^b$
              Ed Snelson$^a$\\
        \addr $^a$ Microsoft Cambridge, UK.\\
              $^b$ Microsoft Research, Redmond, WA.\\
              $^c$ Microsoft Online Services Division, Bellevue, WA.}

\editor{Peter Spirtes}

\maketitle
\raggedbottom
\sloppy


\begin{abstract}%
This work shows how to leverage causal inference to understand the behavior of
complex learning systems interacting with their environment and predict the
consequences of changes to the system. Such predictions allow both humans and
algorithms to select the changes that would have improved the system
performance. This work is illustrated by experiments carried out on the ad
placement system associated with the Bing search engine.
\end{abstract}



\section{Introduction}

Statistical machine learning technologies in the real world are never
without a purpose. Using their predictions, humans or machines make
decisions whose circuitous consequences often violate the modeling
assumptions that justified the system design in the first place.

Such contradictions appear very clearly in the case of the learning
systems that power web scale applications such as search engines, ad
placement engines, or recommandation systems. For instance, the
placement of advertisement on the result pages of Internet search
engines depend on the bids of advertisers and on scores computed by
statistical machine learning systems. Because the scores affect the
contents of the result pages proposed to the users, they directly
influence the occurrence of clicks and the corresponding advertiser
payments.  They also have important indirect effects. Ad placement
decisions impact the satisfaction of the users and therefore their
willingness to frequent this web site in the future.  They also impact
the return on investment observed by the advertisers and therefore
their future bids. Finally they change the nature of the data
collected for training the statistical models in the future.

These complicated interactions are clarified by important theoretical
works. Under simplified assumptions, mechanism
design \citep{myerson-1981} leads to an insightful account of the
advertiser feedback loop \citep{varian-2007,edelman-2007}.  Under
simplified assumptions, multiarmed bandits theory
\citep{robbins-1952,auer-2002,langford-zhang-2007}
and reinforcement learning \citep{sutton-barto-1998} describe the
exploration/exploitation dilemma associated with the training feedback
loop. However, none of these approaches gives a complete account of
the complex interactions found in real-life systems.

This work is motivated by a very practical observation: in the data
collected during the operation of an ad placement engine, \emph{all
these fundamental insights manifest themselves in the form of
correlation/causation paradoxes.} Using the ad placement example as a
model of our problem class, we therefore argue that \emph{the language
and the methods of causal inference} provide flexible means to \emph{describe
such complex machine learning systems} and \emph{give sound
answers to the practical questions} facing the designer of such a
system.  Is it useful to pass a new input signal to the statistical
model?  Is it worthwhile to collect and label a new training set?
What about changing the loss function or the learning algorithm? In
order to answer such questions and improve the operational performance
of the learning system, one needs to unravel how the information
produced by the statistical models traverses the web of causes and
effects and eventually produces measurable performance metrics.

Readers with an interest in causal inference will find in this paper
($i$) a \emph{real world example demonstrating the value of causal
inference for large-scale machine learning applications},
($ii$) \emph{causal inference techniques applicable to continuously
valued variables with meaningful confidence intervals}, and ($iii$)
\emph{quasi-static analysis techniques for estimating how small 
interventions affect certain causal equilibria}. Readers with an
interest in real-life applications will find ($iv$) a selection
of \emph{practical counterfactual analysis techniques applicable to
many real-life machine learning systems}. Readers with an interest in
computational advertising will find a principled framework that
($v$) explains \emph{how to soundly use machine learning techniques
for ad placement}, and ($vi$) \emph{conceptually connects machine
learning and auction theory} in a compelling manner.

\smallskip

The paper is organized as follows.
Section~\ref{s-overview} gives an overview of the advertisement placement
problem which serves as our main example.  In particular, we stress some of
the difficulties encountered when one approaches such a problem without a
principled perspective.
Section~\ref{s-causalmodels} provides a condensed review of the essential
concepts of causal modeling and inference.
Section~\ref{s-counterfactuals} centers on formulating and answering
counterfactual questions such as ``how would the system have performed during
the data collection period if certain interventions had been carried out on
the system\,?''  We describe importance sampling methods for counterfactual
analysis, with clear conditions of validity and confidence intervals.
Section~\ref{s-structure} illustrates how the structure
of the causal graph reveals opportunities to exploit prior information
and vastly improve the confidence intervals.
Section~\ref{s-learning} describes how counterfactual analysis provides 
essential signals that can drive learning algorithms. Assume that we have
identified interventions that would have caused the system to perform well
during the data collection period. Which guarantee can we obtain on the
performance of these same interventions in the future?
Section~\ref{s-equilibrium} presents counterfactual differential techniques
for the study of equlibria. Using data collected when the system is at
equilibrium, we can estimate how a small intervention displaces the
equilibrium. This provides an elegant and effective way to reason about
long-term feedback effects.
Various appendices complete the main text with information that we think more
relevant to readers with specific backgrounds.


\section{Causation Issues in Computational Advertising}
\label{s-overview}

After giving an overview of the advertisement placement problem, which serves
as our main example, this section illustrates some of the difficulties that
arise when one does not pay sufficient attention to the causal structure of
the learning system.


\subsection{Advertisement Placement}
\label{s-adplacement}

\begin{figure}
\center
\includegraphics[width=.75\linewidth]{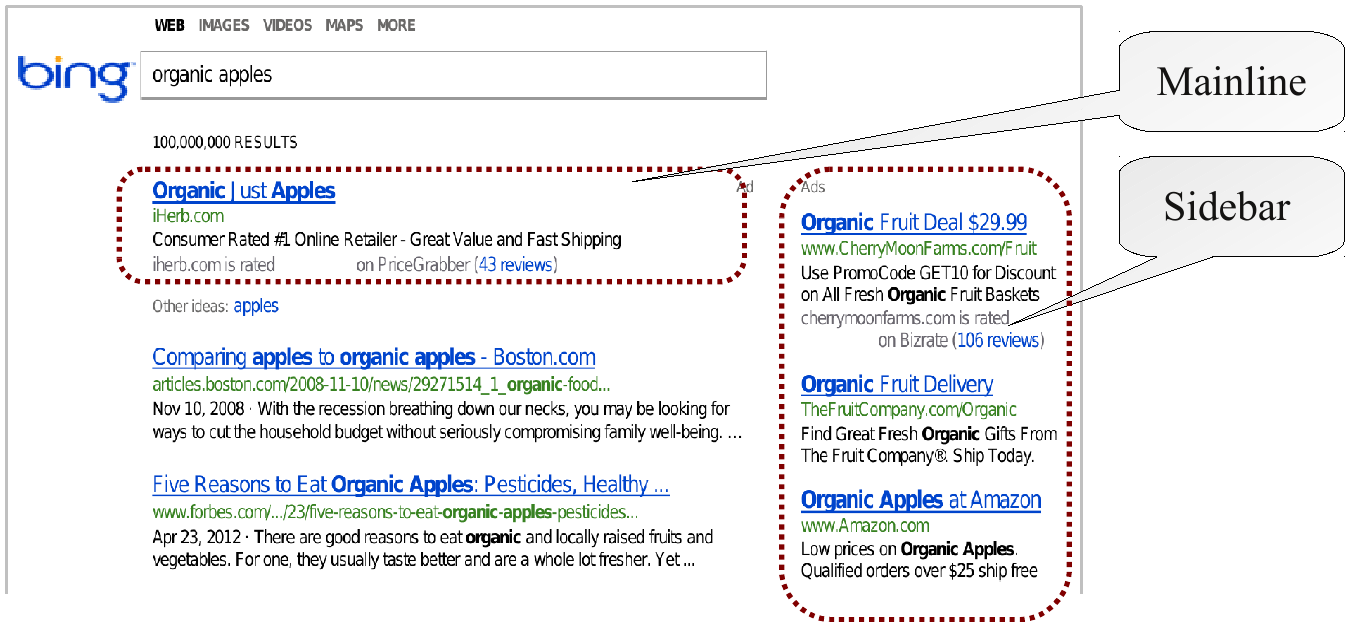}
\caption{Mainline and sidebar ads on a search result page. Ads placed in the
  mainline are more likely to be noticed, increasing both the chances of a
  click if the ad is relevant and the risk of annoying the user if the ad is
  not relevant.}
\label{fig-adlocations}
\end{figure}

All Internet users are now familiar with the advertisement messages
that adorn popular web pages. Advertisements are particularly
effective on search engine result pages because users who are
searching for something are good targets for advertisers who have
something to offer.  Several actors take part in this Internet
advertisement game:
\begin{itemize}
\item
  Advertisers create advertisement messages, and place bids that
  describe how much they are willing to pay to see their ads
  displayed or clicked.
\item
  Publishers provide attractive web services, such as, for instance,
  an Internet search engine. They display selected ads and expect to
  receive payments from the advertisers. The infrastructure to collect
  the advertiser bids and select ads is sometimes provided by an
  advertising network on behalf of its affiliated publishers. For the
  purposes of this work, we simply consider a publisher large enough
  to run its own infrastructure.
\item 
  Users reveal information about their current interests, for instance, by
  entering a query in a search engine. They are offered web pages that contain 
  a selection of ads (figure~\ref{fig-adlocations}). Users sometimes click on an
  advertisement and are transported to a web site controlled by the advertiser
  where they can initiate some business.
\end{itemize}
A conventional bidding language is necessary to precisely define under which
conditions an advertiser is willing to pay the bid amount. In the case of
Internet search advertisement, each bid specifies (a) the advertisement
message, (b) a set of keywords, (c) one of several possible matching criteria
between the keywords and the user query, and (d) the maximal price the
advertiser is willing to pay when a user clicks on the ad after entering a
query that matches the keywords according to the specified criterion.

Whenever a user visits a publisher web page, an advertisement placement engine
runs an auction in real time in order to select winning ads, determine where
to display them in the page, and compute the prices charged to advertisers,
should the user click on their ad. Since the placement engine is operated 
by the publisher, it is designed to further the interests of the publisher. 
Fortunately for everyone else, the publisher must balance short
term interests, namely the immediate revenue brought by the ads displayed on
each web page, and long term interests, namely the future revenues resulting
from the continued satisfaction of both users and advertisers.

Auction theory explains how to design a mechanism that optimizes the revenue
of the seller of a single object~{\citep{myerson-1981,milgrom-2004}} under
various assumptions about the information available to the buyers regarding
the intentions of the other buyers. In the case of the ad placement problem,
the publisher runs multiple auctions and sells opportunities to receive a click. 
When nearly identical auctions occur thousand of times per second, it
is tempting to consider that the advertisers have perfect information
about each other. This assumption gives support to the popular generalized
second price rank-score auction {\citep{varian-2007,edelman-2007}}:
\begin{itemize}
  \item
    Let $x$ represent the auction context information, such as the user
    query, the user profile, the date, the time, etc.  The ad placement
    engine first determines all eligible ads $a_1\dots a_n$ and the
    corresponding bids $b_1\dots b_n$ on the basis of the auction context $x$
    and of the matching criteria specified by the advertisers.
  \item
    For each selected ad $a_i$ and each potential position $p$ on the web
    page, a statistical model outputs the estimate $q_{i,p}(x)$ of the
    probability that ad $a_i$ displayed in position $p$ receives a user click.
    The rank-score $r_{i,p}(x) = b_i q_{i,p}(x)$ then represents the purported
    value associated with placing ad $a_i$ at position $p$.
  \item 
    Let $L$ represent a possible ad layout, that is, a set of positions that
    can simultaneously be populated with ads, and let $\calL$ be
    the set of possible ad layouts, including of course the empty layout. 
    The optimal layout and the corresponding ads are 
    obtained by maximizing the total rank-score
    \begin{equation}
      \label{eq-maxrankscore}
        \max_{L\in{\calL}} ~ \max_{i_1,i_2,\dots} 
           ~~ \sum_{p\in L} r_{i_p,p}(x) \,,
    \end{equation}
    subject to reserve constraints
    \begin{equation}
      \label{eq-reserves}
        \forall p\in L,~ r_{i_p,p}(x) \ge R_p(x) \,,
    \end{equation}
    and also subject to diverse policy constraints, such as, for instance,
    preventing the simultaneous display of multiple ads belonging to the same
    advertiser.  Under mild assumptions, this discrete maximization problem is
    amenable to computationally efficient greedy algorithms (see
    appendix~\ref{a-greedyadplacement}.)
  \item
    The advertiser payment associated with a user click is computed using the
    generalized second price (GSP) rule: the advertiser pays the smallest bid
    that it could have entered without changing the solution of the discrete
    maximization problem, all other bids remaining equal. In other words, the
    advertiser could not have manipulated its bid and obtained the same
    treatment for a better price.
\end{itemize}
Under the perfect information assumption, the analysis suggests that the
publisher simply needs to find which reserve prices $R_p(x)$ yield the best
revenue \emph{per auction}. However, the total revenue of the publisher also
depends on the traffic experienced by its web site.  Displaying an excessive
number of irrelevant ads can train users to ignore the ads, and can also
drive them to competing web sites.  Advertisers can artificially raise the
rank-scores of irrelevant ads by temporarily increasing the bids. Indelicate
advertisers can create deceiving advertisements that elicit many
clicks but direct users to spam web sites. Experience shows that the continued
satisfaction of the users is more important to the publisher than it is to the
advertisers.

Therefore the generalized second price rank-score auction has evolved.
Rank-scores have been augmented with terms that quantify the user satisfaction
or the ad relevance. Bids receive adaptive discounts in order to deal with
situations where the perfect information assumption is unrealistic.  These
adjustments are driven by additional statistical models.  The ad placement
engine should therefore be viewed as a complex learning system interacting
with both users and advertisers.


\subsection{Controlled Experiments}
\label{s-flighting}

The designer of such an ad placement engine faces the fundamental question of
testing whether a proposed modification of the ad placement engine results in
an improvement of the operational performance of the system.

The simplest way to answer such a question is to try the modification.  
The basic idea is to randomly split the users into treatment and control
groups {\citep{kohavi-2008}}.  Users from the control group see web pages
generated using the unmodified system.  Users of the treatment groups see web
pages generated using alternate versions of the system.  Monitoring various
performance metrics for a couple months usually gives sufficient information
to reliably decide which variant of the system delivers the most satisfactory
performance.

Modifying an advertisement placement engine elicits reactions from both the
users and the advertisers. Whereas it is easy to split users into treatment
and control groups, splitting advertisers into treatment and control groups
demands special attention because each auction involves multiple advertisers
{\citep{charles-2012a}}. Simultaneously controlling for both users and
advertisers is probably impossible. 

Controlled experiments also suffer from several drawbacks. They are
expensive because they demand a complete implementation of the
proposed modifications. They are slow because each experiment typically
demands a couple months. Finally, although there are elegant ways to
efficiently run overlapping controlled experiments on the same traffic
{\citep{tang-2010}}, they are limited by the volume of traffic
available for experimentation.

It is therefore difficult to rely on controlled experiments during the
conception phase of potential improvements to the ad placement engine.  It is
similarly difficult to use controlled experiments to drive the training
algorithms associated with click probability estimation
models{\eat{~\citep{simard-2012a}}}. Cheaper and faster statistical methods 
are needed to drive these essential aspects of the development of an ad 
placement engine. Unfortunately, interpreting cheap and fast data can be 
very deceiving.


\subsection{Confounding Data}
\label{s-simpson}

Assessing the consequence of an intervention using statistical data is
generally challenging because it is often difficult to determine whether 
the observed effect is a simple consequence of the intervention or 
has other uncontrolled causes.
 
For instance, the empirical comparison of certain kidney stone treatments
illustrates this difficulty \citep{charig-1986}. Table~\ref{tab-simpson}
reports the success rates observed on two groups of 350 patients treated with
respectively open surgery (treatment~A, with 78\% success) and percutaneous
nephrolithotomy (treatment~B, with 83\% success).  Although treatment~B seems
more successful, it was more frequently prescribed to patients suffering from
small kidney stones, a less serious condition.  Did treatment~B achieve a high
success rate because of its intrinsic qualities or because it was
preferentially applied to less severe cases?  Further splitting the data
according to the size of the kidney stones reverses the conclusion:
treatment~A now achieves the best success rate for both patients suffering
from large kidney stones and patients suffering from small kidney stones.
Such an inversion of the conclusion is called Simpson's paradox
\citep{simpson-1951}.

\begin{table}
\center
\caption{A classic example of Simpson's paradox.
The table reports the success rates of two treatments 
for kidney stones~\citep[tables I and II]{charig-1986}.
Although the overall success rate of treatment B seems better,
treatment B performs worse than treatment A on both patients 
with small kidney stones and patients with large kidney stones.
See section~\ref{s-simpson}.}
\par\bigskip
\label{tab-simpson}
\begin{tabular}{llll}
\toprule
& \hfil\parbox{6em}{%
     Overall} 
& \hfil\parbox{6em}{%
     Patients with\\[-.3ex] small stones}
& \hfil\parbox{6em}{%
     Patients with \\[-.3ex] large stones }
\\
\midrule
\parbox{12.5em}{%
  Treatment A:\\[-.3ex] {\small Open surgery}}
& 78\% (273/350) & {\bf 93\%} (81/87) & {\bf 73\%} (192/263) 
\\
\midrule
\parbox{12.5em}{%
  Treatment B:\\[-.3ex] {\small Percutaneous\:nephrolithotomy}}
& {\bf 83\%} (289/350) & 87\% (234/270) & 69\% (55/80) 
\\
\bottomrule
\end{tabular}
\par\smallskip
\end{table}

The stone size in this study is an example of a \emph{confounding variable},
that is an uncontrolled variable whose consequences pollute the effect of the
intervention.  Doctors knew the size of the kidney stones, chose to treat the
healthier patients with the least invasive treatment~B, and therefore caused
treatment~B to appear more effective than it actually was.  If we now decide
to apply treatment~B to all patients irrespective of the stone size, we break
the causal path connecting the stone size to the outcome, we eliminate the
illusion, and we will experience disappointing results.

When we suspect the existence of a confounding variable, we can split the
contingency tables and reach improved conclusions.  Unfortunately we cannot
fully trust these conclusions unless we are certain to have taken into account
all confounding variables. The real problem therefore comes from the
confounding variables we do not know.

Randomized experiments arguably provide the only correct solution to this
problem \citep[see][]{stigler-1992}.  The idea is to randomly chose whether
the patient receives treatment~A or treatment~B. Because this random choice is
independent from all the potential confounding variables, known and unknown,
they cannot pollute the observed effect of the treatments (see also
section~\ref{s-reweight}).  This is why controlled experiments in ad placement
(section~\ref{s-flighting}) randomly distribute users between treatment and
control groups, and this is also why, in the case of an ad placement engine,
we should be somehow concerned by the practical impossibility to randomly 
distribute both users and advertisers.


\subsection{Confounding Data in Ad Placement}
\label{s-simpson-in-ads}

Let us return to the question of assessing the value of passing a new 
input signal to the ad placement engine click prediction model.
Section~\ref{s-adplacement} outlines a placement method where the click
probability estimates $q_{i,p}(x)$ depend on the ad and the position we
consider, but do not depend on other ads displayed on the page.  We now
consider replacing this model by a new model that additionally uses the
estimated click probability of the top mainline ad to estimate the click
probability of the second mainline ad (figure~\ref{fig-adlocations}).  We
would like to estimate the effect of such an intervention using existing
statistical data.

We have collected ad placement data for Bing\footnote{\url{http://bing.com}}
search result pages served during three consecutive hours on a certain slice
of traffic. Let $q_1$ and $q_2$ denote the click probability estimates
computed by the existing model for respectively the top mainline ad and the
second mainline ad.  After excluding pages displaying fewer than two mainline
ads, we form two groups of 2000 pages randomly picked among those satisfying
the conditions $q_1<0.15$ for the first group and $q_1\ge0.15$ for the second
group.  Table~\ref{tab-simpson-in-ads} reports the click counts and
frequencies observed on the second mainline ad in each group.  Although the
overall numbers show that users click more often on the second mainline ad when
the top mainline ad has a high click probability estimate $q_1$, this
conclusion is reversed when we further split the data according to the click
probability estimate $q_2$ of the second mainline ad.
 
\begin{table}
\center
\caption{Confounding data in ad placement.  The table reports the
  click-through rates and the click counts of the second mainline ad. The
  overall counts suggest that the click-through rate of the second mainline ad
  increases when the click probability estimate $q_1$ of the top ad is high.
  However, if we further split the pages according to the click probability
  estimate $q_2$ of the second mainline ad, we reach the opposite conclusion.
  See section~\ref{s-simpson-in-ads}.}  
\par\bigskip
\label{tab-simpson-in-ads}
\begin{tabular}{llll}
\toprule
& \hfil Overall 
& \hfil $q_2$ low 
& \hfil $q_2$ high
\\
\midrule
$q_1$ low
&  ~6.2\% (124/2000)
&  {\bf ~5.1\%} (92/1823)
&  {\bf 18.1\%} (32/176)
\\
\midrule
$q_1$ high
& {\bf ~7.5\%} (149/2000)
& ~4.8\% (71/1500)
& 15.6\% (78/500)
\\
\bottomrule
\end{tabular}
\par\smallskip
\end{table}

Despite superficial similarities, this example is considerably more difficult
to interpret than the kidney stone example. The overall click counts show that
the actual click-through rate of the second mainline ad is positively correlated
with the click probability estimate on the top mainline ad. Does this mean
that we can increase the total number of clicks by placing regular ads below
frequently clicked ads?

Remember that the click probability estimates depend on the search query which
itself depends on the user intention. The most likely explanation is that
pages with a high $q_1$ are frequently associated with more commercial
searches and therefore receive more ad clicks on all positions. The observed
correlation occurs because the presence of a click and the magnitude of the
click probability estimate $q_1$ have a common cause: the user intention.
Meanwhile, the click probability estimate $q_2$ returned by the current model
for the second mainline ad also depend on the query and therefore the user
intention.  Therefore, assuming that this dependence has comparable strength,
and assuming that there are no other causal paths, splitting the counts
according to the magnitude of $q_2$ factors out the effects of this common
confounding cause. We then observe a negative correlation which now suggests
that a frequently clicked top mainline ad has a negative impact on the
click-through rate of the second mainline ad.  

If this is correct, we would probably increase the accuracy of the click
prediction model by switching to the new model. This would decrease the click
probability estimates for ads placed in the second mainline position on
commercial search pages. These ads are then less likely to clear the reserve
and therefore more likely to be displayed in the less attractive sidebar. The
net result is probably a loss of clicks and a loss of money despite the higher
quality of the click probability model. Although we could tune the reserve
prices to compensate this unfortunate effect, nothing in this data tells us
where the performance of the ad placement engine will land. Furthermore,
unknown confounding variables might completely reverse our conclusions.

Making sense out of such data is just too complex\,!


\subsection{A Better Way}
\label{s-better-way}

It should now be obvious that we need a more principled way to reason
about the effect of potential interventions. We provide one such more 
principled approach using the causal inference machinery 
(section~\ref{s-causalmodels}). The next
step is then the identification of a class of questions that are
sufficiently expressive to guide the designer of a complex learning
system, and sufficiently simple to be answered using data collected in
the past using adequate procedures (section~\ref{s-counterfactuals}).

A machine learning algorithm can then be viewed as an automated way to
generate questions about the parameters of a statistical model, obtain
the corresponding answers, and update the parameters accordingly
(section~\ref{s-learning}).  Learning algorithms derived in this
manner are very flexible: human designers and machine learning
algorithms can cooperate seamlessly because they rely on similar
sources of information.


\section{Modeling Causal Systems}
\label{s-causalmodels}

When we point out a causal relationship between two events, we
describe what we expect to happen to the event we call the
\emph{effect}, should an external operator manipulate the event we
call the \emph{cause}. Manipulability theories of causation
\citep{von-wright-1971,woodward-2005} raise this commonsense
insight to the status of a definition of the causal relation. Difficult
adjustments are then needed to interpret statements involving
causes that we can only observe through their effects, 
\emph{``because they love me,''} or that are not easily 
manipulated, \emph{``because the earth is round.''}

Modern statistical thinking makes a clear distinction between the
statistical model and the world. The actual mechanisms underlying the
data are considered unknown. The statistical models do not need to
reproduce these mechanisms to emulate the observable
data \citep{breiman-2001}. Better models are sometimes
obtained by deliberately avoiding to reproduce the true
mechanisms \citep[section~8.6]{vapnik-1982}. We can approach
the manipulability puzzle in the same spirit by viewing causation as a
reasoning model {\citep{bottou-2011}} rather than a property of the
world. Causes and effects are simply the pieces of an abstract
reasoning game. Causal statements that are not empirically testable
acquire validity when they are used as intermediate steps when one
reasons about manipulations or interventions amenable to
experimental validation.

This section presents the rules of this reasoning game. 
We largely follow the framework proposed by \citet{pearl-2009} 
because it gives a clear account of the connections between 
causal models and probabilistic models. 


\subsection{The Flow of Information}
\label{s-flow}

\begin{figure}[t]
\begin{sem}
\begin{equation*}
\begin{array}{rcl@{\qquad}l}
x & = & f_1(u,\varepsilon_1) 
& \text{\small\unboldmath 
  Query context $x$ from user intent $u$.} \\
a & = & f_2(x, v, \varepsilon_2)
& \text{\small\unboldmath 
  Eligible ads $(a_i)$ from query $x$ and inventory $v$.} \\
b & = & f_3(x, v, \varepsilon_3)
& \text{\small\unboldmath 
  Corresponding bids $(b_i)$.} \\
q & = & f_4(x, a, \varepsilon_4)
  & \text{\small\unboldmath 
  Scores $(q_{i,p}, R_p)$ from query $x$ and ads $a$.} \\
s & = & f_5(a, q, b, \varepsilon_5)
  & \text{\small\unboldmath 
  Ad slate $s$ from eligible ads $a$, scores $q$ and bids $b$.} \\
c & = & f_6(a, q, b, \varepsilon_6)
  & \text{\small\unboldmath 
  Corresponding click prices $c$.} \\
y & = & f_7(s, u, \varepsilon_7)
  & \text{\small\unboldmath 
  User clicks $y$ from ad slate $s$ and user intent $u$.} \\
z & = & f_8(y, c, \varepsilon_8)
  & \text{\small\unboldmath 
  Revenue $z$ from clicks $y$ and prices $c$.}
\end{array}
\end{equation*}
\end{sem}
\caption{A structural equation model for ad placement.  The sequence
  of equations describes the flow of information.  The functions $f_k$
  describe how effects depend on their direct causes.  The additional
  noise variables $\varepsilon_k$ represent independent sources of
  randomness useful to model probabilistic dependencies.}
\label{fig-sem}
\end{figure}

Figure~\ref{fig-sem} gives a deterministic description of the operation of the
ad placement engine. Variable $u$ represents the user and his or her intention
in an unspecified manner. The query and query context $x$ is then expressed as
an unknown function of the $u$ and of a noise variable $\varepsilon_1$. Noise
variables in this framework are best viewed as independent sources of
randomness useful for modeling a nondeterministic causal dependency.  We shall
only mention them when they play a specific role in the discussion. The set of
eligible ads $a$ and the corresponding bids $b$ are then derived from the
query $x$ and the ad inventory $v$ supplied by the advertisers. Statistical
models then compute a collection of scores $q$ such as the click probability
estimates $q_{i,p}$ and the reserves $R_p$ introduced in
section~\ref{s-adplacement}.  The placement logic uses these scores to
generate the ``ad~slate'' $s$, that is, the set of winning ads and their
assigned positions.  The corresponding click prices $c$ are computed. The set
of user clicks $y$ is expressed as an unknown function of the ad slate $s$ and
the user intent $u$. Finally the revenue $z$ is expressed as another 
function of the clicks $y$ and the prices $c$.

Such a system of equations is named \emph{structural equation
model} \citep{wright-1921}. Each equation asserts a functional
dependency between an effect, appearing on the left hand side of the
equation, and its direct causes, appearing on the right hand side as
arguments of the function.  Some of these causal dependencies are
\emph{unknown}. Although we postulate that the effect can be expressed as some
function of its direct causes, we do not know the form of this function. For
instance, the designer of the ad placement engine knows functions~$f_2$
to~$f_6$ and $f_8$ because he has designed them. However, he does not know the
functions~$f_1$ and~$f_7$ because whoever designed the user did not leave
sufficient documentation.

\begin{figure}[t]
\center
\includegraphics[width=.45\linewidth]{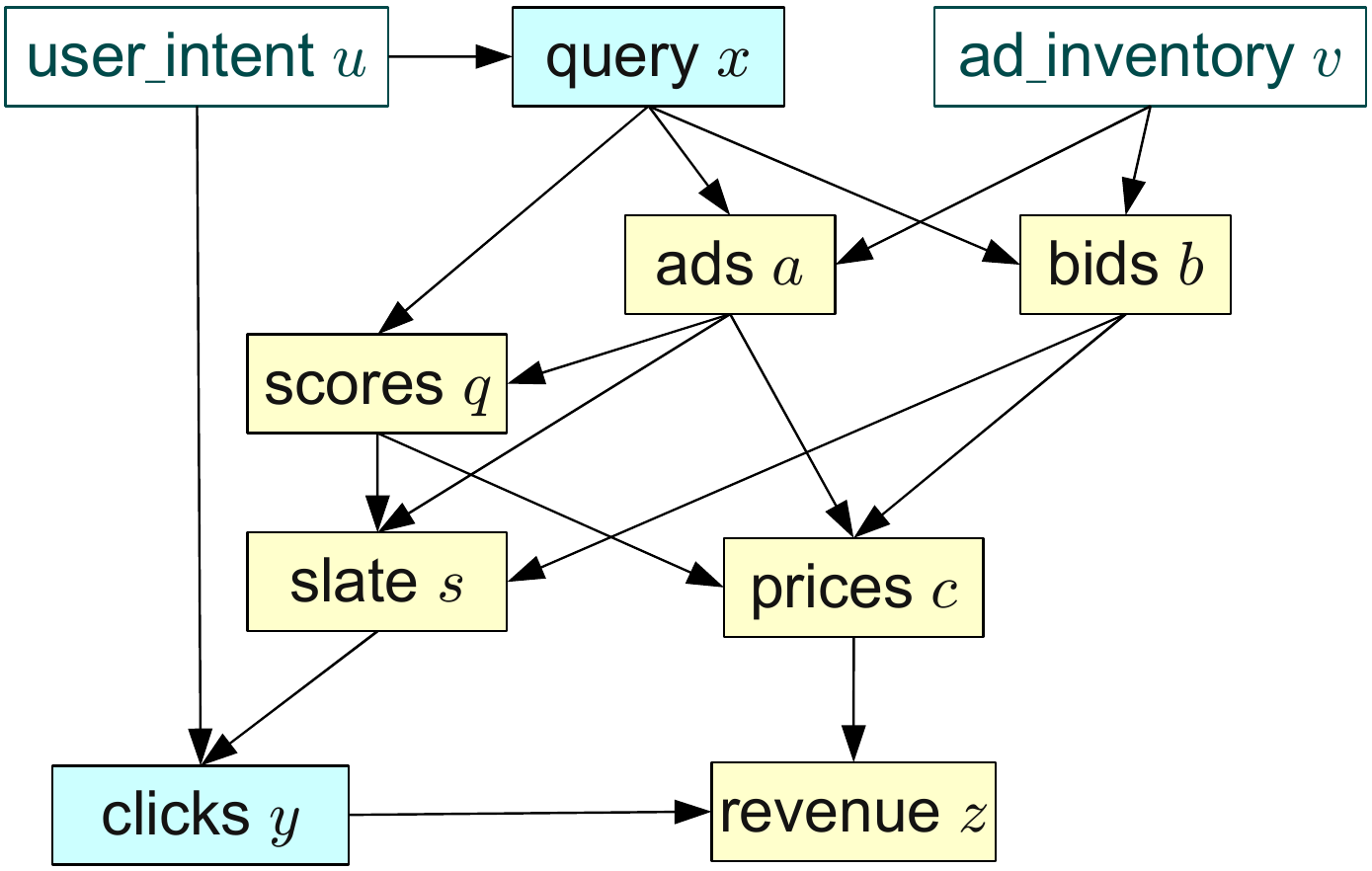}
\caption{Causal graph associated with the ad placement structural equation
  model (figure~\ref{fig-sem}). Nodes with yellow (as opposed to blue)
  background indicate bound variables with known functional
  dependencies. The mutually independent noise variables are implicit.}
\label{fig-causalgraph}
\end{figure}

Figure~\ref{fig-causalgraph} represents the directed causal graph associated
with the structural equation model. Each arrow connects a direct cause to its
effect. The noise variables are omitted for simplicity. The structure of this
graph reveals fundamental assumptions about our model. For instance, the user
clicks $y$ do not directly depend on the scores $q$ or the prices $c$
because users do not have access to this information.

We hold as a principle that causation obeys the \emph{arrow of time}:  
causes always precede their effects. Therefore the causal graph 
must be \emph{acyclic}. Structural equation models then support two
fundamental operations, namely simulation and intervention.
\begin{itemize}
\item
  \emph{Simulation} -- Let us assume that we know both the exact form of all
  functional dependencies and the value of all exogenous variables, that is, the
  variables that never appear in the left hand side of an equation.  We can
  compute the values of all the remaining variables by applying the equations
  in their natural time sequence.
\item
  \emph{Intervention} -- As long as the causal graph remains acyclic, we can
  construct derived structural equation models using arbitrary algebraic
  manipulations of the system of equations.  For instance, we can clamp a
  variable to a constant value by rewriting the right-hand side of the
  corresponding equation as the specified constant value.
\end{itemize}

\noindent
The algebraic manipulation of the structural equation models provides
a powerful language to describe interventions on a causal system.
This is not a coincidence. Many aspects of the mathematical notation
were invented to support causal inference in classical mechanics.
However, we no longer have to interpret the variable values as
physical quantities: the equations simply describe the flow of
information in the causal model \citep{wiener-1948}.


\subsection{The Isolation Assumption}
\label{s-isolation}

Let us now turn our attention to the exogenous variables, that is,
variables that never appear in the left hand side of an equation of
the structural model. Leibniz's \emph{principle of sufficient reason}
claims that there are no facts without causes. This suggests
that the exogenous variables are the effects of a network of causes
not expressed by the structural equation model. For instance, the
user intent $u$ and the ad inventory $v$ in 
figure~\ref{fig-causalgraph} have temporal correlations because both
users and advertisers worry about their budgets when the end of the
month approaches.  Any structural equation model should then be understood
in the context of a larger structural equation model potentially
describing all things in existence.

\begin{figure}
\center
\includegraphics[width=.8\linewidth]{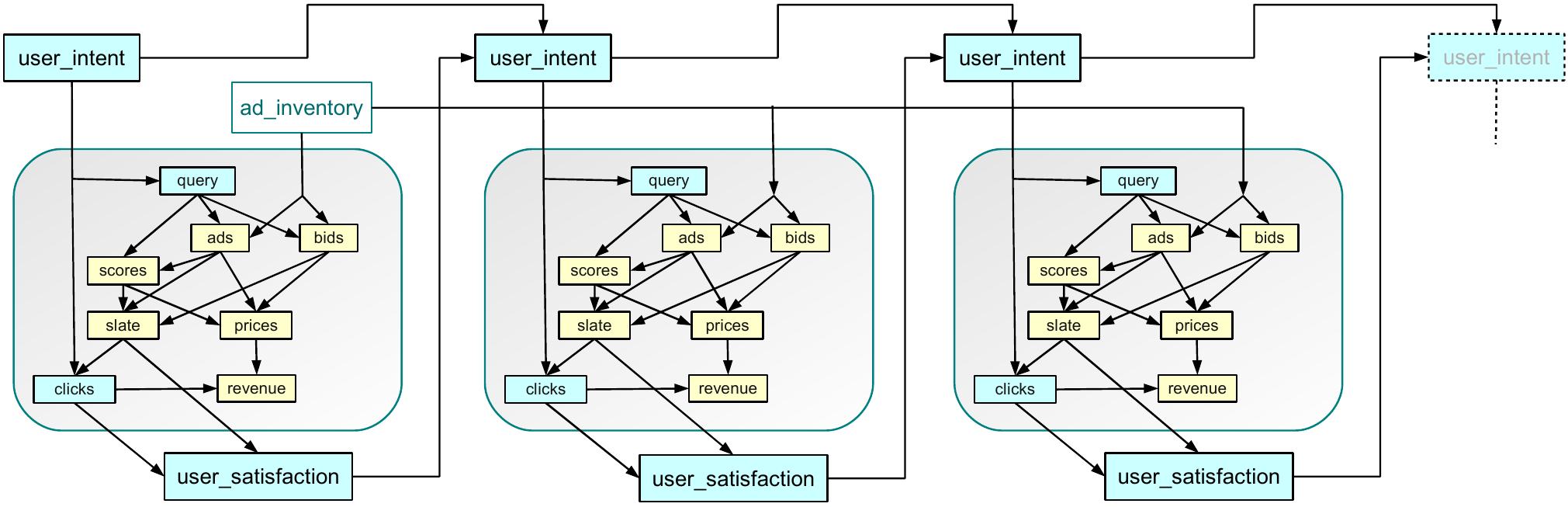}
\caption{Conceptually unrolling the user feedback loop by threading instances
  of the single page causal graph (figure~\ref{fig-causalgraph}).  Both the
  ad slate $s_t$ and user clicks $y_t$ have an indirect effect 
  on the user intent $u_{t+1}$ associated with the next query.}
\label{fig-manysem}
\end{figure}

Ads served on a particular page contribute to the continued satisfaction of
both users and advertisers, and therefore have an effect on their willingness
to use the services of the publisher in the future. The ad placement
structural equation model shown in figure~\ref{fig-sem} only describes the
causal dependencies for a single page and therefore cannot account for such
effects. Consider however a very large structural equation model containing a
copy of the page-level model for every web page ever served by the
publisher. Figure~\ref{fig-manysem} shows how we can thread the page-level
models corresponding to pages served to the same user. Similarly we could
model how advertisers track the performance and the cost of their
advertisements and model how their satisfaction affects their future bids.
The resulting causal graphs can be very complex. Part of this complexity
results from time-scale differences. Thousands of search pages are served in a
second. Each page contributes a little to the continued satisfaction of one
user and a few advertisers. The accumulation of these contributions produces
measurable effects after a few weeks.

Many of the functional dependencies expressed by the structural
equation model are left unspecified. Without direct knowledge of these
functions, we must reason using statistical data. The most fundamental
statistical data is collected from repeated trials that are assumed
independent.  When we consider the large structured equation model of
everything, we can only have one large trial producing a single data
point.\footnote{\relax See also the discussion on reinforcement
learning, section~\ref{s-bandits-and-rl}.  } It is therefore desirable
to identify repeated patterns of identical equations that can be
viewed as repeated independent trials. Therefore, when we study a
structural equation model representing such a pattern, we need to make
an additional assumption to expresses the idea that the oucome of one
trial does not affect the other trials. We call such an assumption
an \emph{isolation assumption} by analogy with
thermodynamics.\footnote{The concept of isolation is pervasive in
physics. An isolated system in thermodynamics \citep[section
2.D]{reichl-1998} or a closed system in mechanics \citep[\S
5]{landau-lifshitz-one} evolves without exchanging mass or energy with
its surroundings. Experimental trials involving systems that are
assumed isolated may differ in their initial setup and therefore have
different outcomes. Assuming isolation implies that the
outcome of each trial cannot affect the other trials.} This can be
achieved by assuming that \emph{the exogenous variables are
independently drawn from an unknown but fixed joint probability
distribution}. This assumption cuts the causation effects that could
flow through the exogenous variables.

The noise variables are also exogenous variables
acting as independent source of randomness. The noise variables are
useful to represent the conditional distribution
$\Pc{{\sf effect}}{{\sf causes}}$ using the equation
${\sf effect}\!=\!f({\sf causes},\varepsilon)$.
Therefore, we also assume joint independence between all the noise
variables and any of the named exogenous variable.\footnote{\relax
  Rather than letting two noise variables display measurable 
  statistical dependencies because they share a common cause, 
  we prefer to name the common cause and make the dependency 
  explicit in the graph.} 
For instance, in the case of the ad placement model 
shown in figure~\ref{fig-sem}, we assume that the joint
distribution of the exogenous variables factorizes as
\begin{equation}
\label{eq-isolation}
    \Pa{u,v,\varepsilon_1,\dots,\varepsilon_8} = 
    \Pa{u,v} \, \Pa{\varepsilon_1} \dots \Pa{\varepsilon_8}\,.
\end{equation}

Since an isolation assumption is only true up to a point, it should
be expressed clearly and remain under constant scrutiny.  We
must therefore measure additional performance metrics that reveal how
the isolation assumption holds. For instance, the ad placement
structural equation model and the corresponding causal graph
(figures~\ref{fig-sem} and~\ref{fig-causalgraph}) do not take user
feedback or advertiser feedback into account. Measuring the revenue is
not enough because we could easily generate revenue at the expense of
the satisfaction of the users and advertisers. When we evaluate
interventions under such an isolation assumption, we also need to
measure a battery of additional quantities that act as proxies for
the user and advertiser satisfaction. Noteworthy examples include
ad relevance estimated by human judges, and advertiser surplus
estimated from the auctions~\citep{varian-2009}.


\subsection{Markov Factorization}
\label{s-markov-factorization}

Conceptually, we can draw a sample of the exogenous variables using
the distribution specified by the isolation assumption, and we can
then generate values for all the remaining variables by simulating the
structural equation model.  

\begin{figure}[t]
\begin{sem}
\begin{equation*}
 \Pa{\begin{array}{c}
    u, v, x, a, b\\
    q, s, c, y, z
     \end{array} }
 \quad = \quad \left\{
\begin{array}{l@{\qquad}l} 
  \,\Pa{u,v} & \text{\small\unboldmath Exogenous vars.}\\
  \times~\Pc{x}{u} & \text{\small\unboldmath Query.} \\
  \times~\Pc{a}{x, v} & \text{\small\unboldmath Eligible ads.} \\
  \times~\Pc{b}{x, v} & \text{\small\unboldmath Bids.} \\
  \times~\Pc{q}{x, a} & \text{\small\unboldmath Scores.} \\
  \times~\Pc{s}{a, q, b} & \text{\small\unboldmath Ad slate.} \\
  \times~\Pc{c}{a, q, b} & \text{\small\unboldmath Prices.} \\
  \times~\Pc{y}{s, u} & \text{\small\unboldmath Clicks.} \\
  \times~\Pc{z}{y, c} & \text{\small\unboldmath Revenue.} 
\end{array} \right.
\end{equation*}
\end{sem}
\caption{Markov factorization of the structural equation model 
 of figure~\ref{fig-sem}.}
\label{fig-markov}
\bigskip
\par
\center
\includegraphics[width=.44\linewidth]{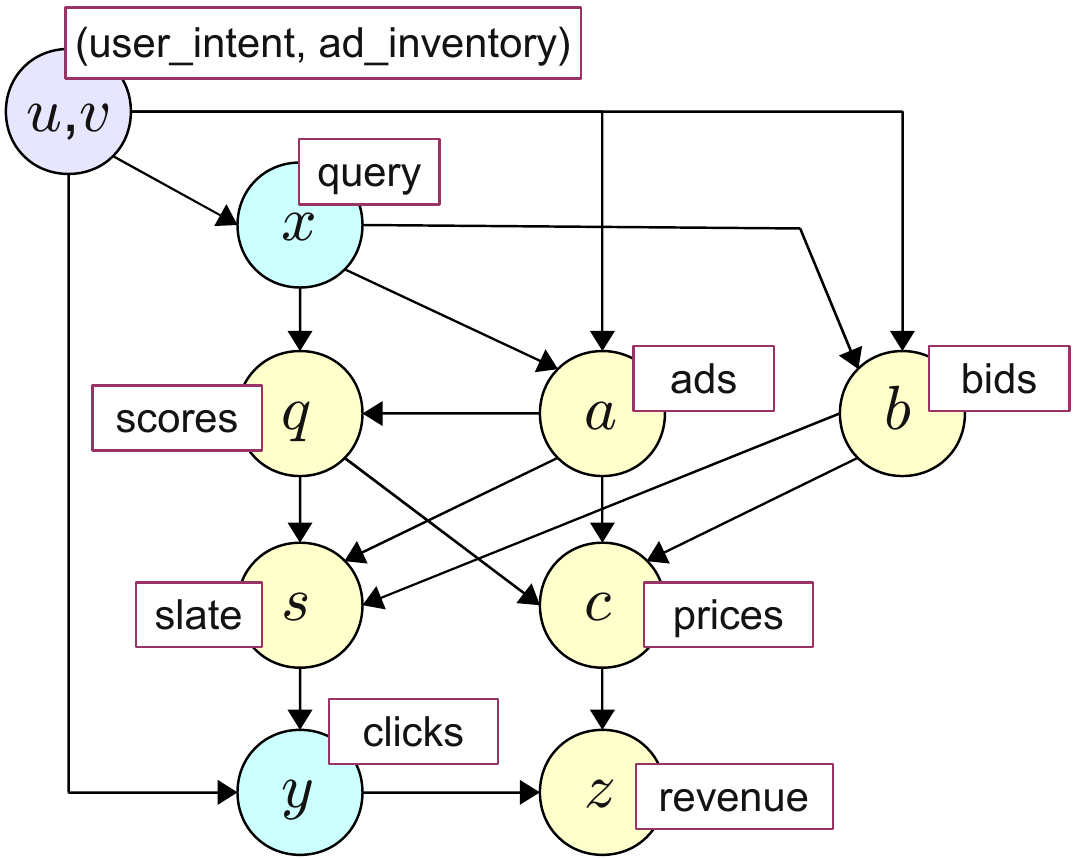}
\caption{Bayesian network associated with the Markov
  factorization shown in figure~\ref{fig-markov}.}
\label{fig-bayesnet}
\end{figure}

This process defines a \emph{generative probabilistic model}
representing the joint distribution of all variables in the structural
equation model.  The distribution readily factorizes as the product of
the joint probability of the named exogenous variables, and, for each
equation in the structural equation model, the conditional probability
of the effect given its direct causes \citep{spirtes-1993,pearl-2000}.
As illustrated by figures~\ref{fig-markov} and~\ref{fig-bayesnet},
this \emph{Markov factorization} connects the structural equation
model that describes causation, and the Bayesian network that
describes the joint probability distribution followed by the variables
under the isolation assumption.\footnote{Bayesian networks are
directed graphs representing the Markov factorization of a joint
probability distribution: the arrows no longer have a causal
interpretation.}

Structural equation models and Bayesian networks appear so intimately
connected that it could be easy to forget the differences. The
structural equation model is an algebraic object. As long as the
causal graph remains acyclic, algebraic manipulations are interpreted
as interventions on the causal system. The Bayesian network is a
generative statistical model representing a class of joint probability
distributions, and, as such, does not support algebraic
manipulations. However, the symbolic representation of its 
Markov factorization is an algebraic object, essentially 
equivalent to the structural equation model.

\subsection{Identification, Transportation, and Transfer Learning}

Consider a causal system represented by a structural equation model with some
unknown functional dependencies. Subject to the isolation assumption, data
collected during the operation of this system follows the distribution
described by the corresponding Markov factorization. Let us first assume that
this data is sufficient to identify the joint distribution of the subset of
variables we can observe. We can intervene on the system by clamping the value
of some variables. This amounts to replacing the right-hand side of the
corresponding structural equations by constants. The joint distribution of the
variables is then described by a new Markov factorization that shares many
factors with the original Markov factorization. Which conditional
probabilities associated with this new distribution can we express using only
conditional probabilities identified during the observation of the original
system? This is called the \emph{identifiability} problem. More generally, we
can consider arbitrarily complex manipulations of the structural equation
model, and we can perform multiple experiments involving different
manipulations of the causal system. Which conditional probabilities pertaining
to one experiment can be expressed using only conditional probabilities
identified during the observation of other experiments? This is called
the \emph{transportability} problem.

Pearl's \emph{do}-calculus completely solves the identifiability
problem and provides useful tools to address many instances of the
transportability problem \citep[see][]{pearl-2012}. Assuming that
we \emph{know} the conditional probability distributions involving
observed variables in the original structural equation
model, \emph{do}-calculus allows us to \emph{derive} conditional
distributions pertaining to the manipulated structural equation model.

Unfortunately, we must further distinguish the conditional
probabilities that we know (because we designed them) from those that
we estimate from empirical data. This distinction is important because
estimating the distribution of continuous or high cardinality
variables is notoriously difficult. Furthermore, \emph{do}-calculus
often combines the estimated probabilities in ways that amplify
estimation errors. This happens when the manipulated structural
equation model exercises the variables in ways that were rarely
observed in the data collected from the original structural equation
model.

Therefore we prefer to use much simpler causal inference techniques
(see sections~\ref{s-replay} and~\ref{s-reweight}).  Although these
techniques do not have the completeness properties
of \emph{do}-calculus, they combine estimation and transportation in a
manner that facilitates the derivation of useful confidence intervals.


\subsection{Special Cases}
\label{s-bandits-and-rl}

Three special cases of causal models are particularly relevant to this work.
\begin{itemize}
\item
  In the multi-armed bandit \citep{robbins-1952}, a user-defined policy
  function $\pi$ determines the distribution of action $a\in\{1\dots{K}\}$,
  and an unknown reward function $r$ determines the distribution of the
  outcome $y$ given the action $a$ (figure~\ref{fig-bandits}).  In order to
  maximize the accumulated rewards, the player must construct policies $\pi$
  that balance the exploration of the action space with the exploitation of
  the best action identified so far~\citep{auer-2002,audibert-2007,seldin-2012}.
\item
  The contextual bandit problem~\citep{langford-zhang-2007}
  significantly increases the complexity of multi-armed bandits by
  adding one exogenous variable $x$ to the policy function $\pi$ and
  the reward functions $r$ (figure~\ref{fig-contextual-bandits}).
\item
  Both multi-armed bandit and contextual bandit are special case of
  reinforcement learning \citep{sutton-barto-1998}. In essence, a
  Markov decision process is a sequence of contextual bandits where
  the context is no longer an exogenous variable but a state variable
  that depends on the previous states and actions
  (figure~\ref{fig-reinforcement-learning}).  Note that the policy
  function $\pi$, the reward function $r$, and the transition function
  $s$ are independent of time.  All the time dependencies are
  expressed using the states $s_t$.
\end{itemize}

\begin{figure}[t]
\begin{sem}
\begin{equation*}
\begin{array}{rcl@{\qquad}l}
a &=& \pi(\varepsilon)
  & \text{\small\unboldmath Action $a\in\{1\dots K\}$} \\
y & = & r(a,\: \varepsilon^\prime\:)
  & \text{\small\unboldmath Reward $y\in\bbbr$}
\end{array}
\end{equation*}
\end{sem}
\caption{\label{fig-bandits}\relax Structural equation model for the
  multi-armed bandit problem.  The policy $\pi$ selects a discrete
  action $a$, and the reward function $r$ determines the outcome $y$.
  The noise variables $\varepsilon$ and $\varepsilon^\prime$ represent
  independent sources of randomness useful to model probabilistic
  dependencies.}
\begin{sem}
\begin{equation*}
\begin{array}{rcl@{\qquad}l}
a &=& \pi(x,\: \varepsilon)
  & \text{\small\unboldmath Action $a\in\{1\dots K\}$}\\
y & = & r(x,\: a,\: \varepsilon^\prime)
  & \text{\small\unboldmath Reward $y\in\bbbr$}
\end{array}
\end{equation*}
\end{sem}
\caption{\label{fig-contextual-bandits}\relax
  Structural equation model for contextual bandit problem.
  Both the action and the reward depend on 
  an exogenous context variable $x$.}
\begin{sem}
\begin{equation*}
\begin{array}{rcl@{\qquad}l}
    a_t &=& \pi(s_{t-1},\:\varepsilon_t) 
          & \text{\small\unboldmath Action} \\
    y_t &=& r(s_{t-1},\:a_t,\:\varepsilon^\prime_t\,) 
          & \text{\small\unboldmath Reward $r_t\in\bbbr$}\\
    s_t &=& s(s_{t-1},\:a_t,\:\varepsilon^\pprime_t\,) 
          & \text{\small\unboldmath Next state} 
\end{array}
\end{equation*}
\end{sem}
\caption{\label{fig-reinforcement-learning}\relax
  Structural equation model for reinforcement learning.
  The above equations are replicated for all $t\in\{0\dots,T\}$.
  The context is now provided by a state variable $s_{t-1}$
  that depends on the previous states and actions.}
\end{figure}

\noindent
These special cases have increasing generality. Many simple structural
equation models can be reduced to a contextual bandit problem using
appropriate definitions of the context $x$, the action $a$ and the outcome
$y$. For instance, assuming that the prices $c$ are discrete, the ad placement
structural equation model shown in figure~\ref{fig-sem} reduces to a
contextual bandit problem with context $(u,v)$, actions $(s,c)$ and reward
$z$.  Similarly, given a sufficiently intricate definition of the state
variables $s_t$, all structural equation models with discrete variables can be
reduced to a reinforcement learning problem. Such reductions lose the fine 
structure of the causal graph. We show in section~\ref{s-structure} how this 
fine structure can in fact be leveraged to obtain more information 
from the same experiments.

Modern reinforcement learning algorithms \citep[see][]{sutton-barto-1998}
leverage the assumption that the policy function, the reward function, the
transition function, and the distributions of the corresponding noise
variables, are independent from time. This invariance property provides great
benefits when the observed sequences of actions and rewards are long in
comparison with the size of the state space. Only section~\ref{s-equilibrium}
in this contribution presents methods that take advantage of such an
invariance.  The general question of leveraging arbitrary functional
invariances in causal graphs is left for future work.


\section{Counterfactual Analysis}
\label{s-counterfactuals}

We now return to the problem of formulating and answering questions
about the value of proposed changes of a learning system.  Assume for
instance that we consider replacing the score computation model $M$ of
an ad placement engine by an alternate model $M^*$. 
We seek an answer to the conditional question:
\begin{quotation}
\noindent\llap{``}\emph{\relax
      How will the system perform if we replace model $M$ by model $M^*$\,?}''
\end{quotation}

\goodbreak
Given sufficient time and sufficient resources, we can obtain the
answer using a controlled experiment (section~\ref{s-flighting}).
However, instead of carrying out a new experiment, we would like to
obtain an answer using data that we have already collected in the past.
\begin{quotation}
\noindent\llap{``}\emph{\relax
  How would the system have performed if, when the data was collected,
  we had replaced model~$M$ by model~$M^*$?}''
\end{quotation}

\goodbreak
The answer of this \emph{counterfactual question} is of course a
\emph{counterfactual statement} that describes the system performance 
subject to a condition that did not happen.

Counterfactual statements challenge ordinary logic because they depend
on a condition that is known to be false. Although assertion $A\Rightarrow B$
is always true when assertion~$A$ is false, we certainly do not mean for
all counterfactual statements to be true. \citet{lewis-1973} navigates this
paradox using a modal logic in which a counterfactual statement describes the
state of affairs in an alternate world that resembles ours except for the
specified differences. Counterfactuals indeed offer many subtle ways to
qualify such alternate worlds. For instance, we can easily
describe isolation assumptions (section~\ref{s-isolation}) in a
counterfactual question:
\begin{quotation}
\hyphenpenalty2000
\noindent\llap{``}\emph{\relax
  How would the system have performed if, when the data was collected,
  we had replaced model~$M$ by model~$M^*$
  without incurring user or advertiser reactions?}''
\end{quotation}
The fact that we could not have changed the model without incurring the user
and advertiser reactions does not matter any more than the fact that we did
not replace model $M$ by model~$M^*$ in the first place.  This does not
prevent us from using counterfactual statements to reason about causes and
effects. Counterfactual questions and statements provide a natural framework
to express and share our conclusions.

The remaining text in this section explains how we can answer certain
counterfactual questions using data collected in the past.  More
precisely, we seek to estimate performance metrics that can be
expressed as expectations with respect to the distribution that would
have been observed if the counterfactual conditions had been in
force.\footnote{Although counterfactual expectations can be viewed as 
expectations of  unit-level counterfactuals
{\citep[definition~4]{pearl-2009}}, they elude the semantic subtleties
of unit-level counterfactuals and can be measured with 
randomized experiments (see section~\ref{s-reweight}.)}


\subsection{Replaying Empirical Data}
\label{s-replay}

\begin{figure}
\center
\includegraphics[width=.42\linewidth]{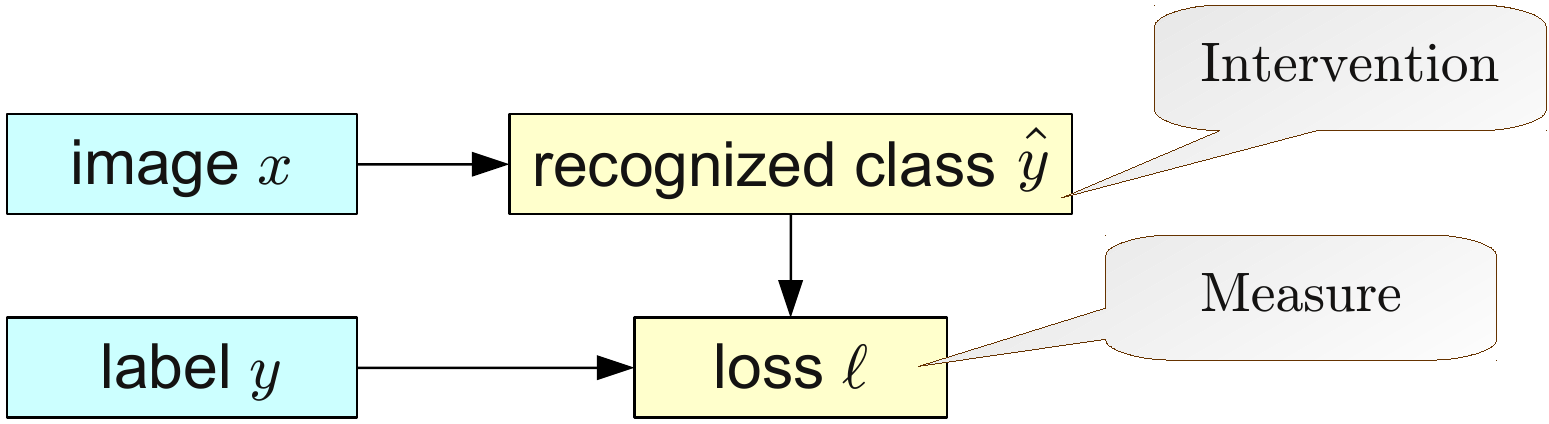}
\caption{Causal graph for an image recognition system. 
  We can estimate counterfactuals by 
  replaying data collected in the past.}
\label{fig-classifier}
\par
\includegraphics[width=.42\linewidth]{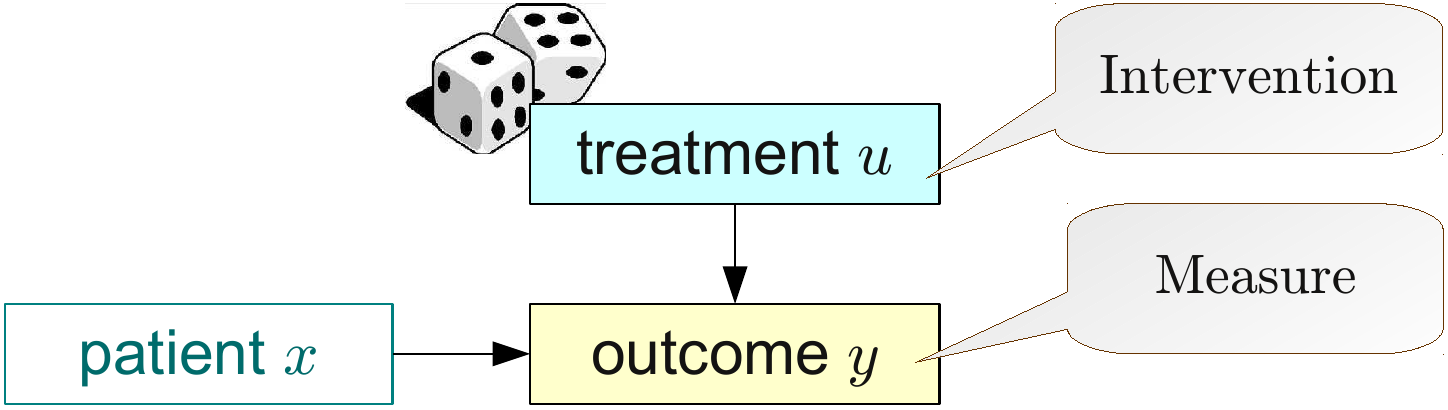}
\caption{Causal graph for a randomized experiment.  
  We can estimate certain counterfactuals by 
  reweighting data collected in the past.}
\label{fig-randomexp}
\end{figure}

Figure~\ref{fig-classifier} shows the causal graph associated with a simple
image recognition system. The classifier takes an image $x$ and produces a
prospective class label $\hat{y}$. The loss measures the penalty associated
with recognizing class $\hat{y}$ while the true class is $y$. 

To estimate the expected error of such a classifier, we collect a
representative data set composed of labeled images, run the classifier
on each image, and average the resulting losses. In other words, we
\emph{replay} the data set to estimate what
(counterfactual) performance would have been observed if we had used a
different classifier. We can then select in retrospect the classifier
that would have worked the best and hope that it will keep working
well. This is the counterfactual viewpoint on empirical risk
minimization \citep{vapnik-1982}.

Replaying the data set works because both the alternate classifier and the
loss function are known. More generally, to estimate a counterfactual
by replaying a data set, we need to know all the functional
dependencies associated with all causal paths connecting the
intervention point to the measurement point.  
This is obviously not always the case.


\subsection{Reweighting Randomized Trials}
\label{s-reweight}

Figure~\ref{fig-randomexp} illustrates the randomized experiment
suggested in section~\ref{s-simpson}. The patients are randomly split
into two equally sized groups receiving respectively treatments $A$
and $B$. The overall success rate for this experiment is therefore
$Y=(Y_A+Y_B)/2$ where $Y_A$ and $Y_B$ are the success rates observed
for each group.  We would like to estimate which (counterfactual)
overall success rate $Y^*$ would have been observed if we had selected
treatment $A$ with probability $p$ and treatment $B$ with probability
$1-p$.

Since we do not know how the outcome depends on the treatment and the
patient condition, we cannot compute which outcome $y^*$ would have
been obtained if we had treated patient $x$ with a different treatment
$u^*$.  Therefore we cannot answer this question by replaying the data
as we did in section~\ref{s-replay}.

However, observing different success rates $Y_A$ and $Y_B$ for the
treatment groups reveals an empirical correlation between the
treatment $u$ and the outcome $y$. Since the only cause of the
treatment~$u$ is an independent roll of the dices, this correlation
cannot result from any known or unknown confounding common
cause.\footnote{See also the discussion of Reichenbach's common cause
principle and of its limitations in 
\citep{spirtes-1993,spirtes-scheines-2004}.}  Having eliminated
this possibility, we can \emph{reweight} the observed outcomes and
compute the estimate $Y^*\approx{p\,Y_A}+{(1-p)\,Y_B}$\,.


\subsection{Markov Factor Replacement}
\label{s-markov-replacement}

The reweighting approach can in fact be
applied under much less stringent conditions. Let us return to the ad
placement problem to illustrate this point.

The average number of ad clicks per page is often called \emph{click yield}.
Increasing the click yield usually benefits both the advertiser and the
publisher, whereas increasing the revenue per page often benefits the
publisher at the expense of the advertiser. Click yield is therefore a very
useful metric when we reason with an isolation assumption that ignores the
advertiser reactions to pricing changes.

Let $\omega$ be a shorthand for all variables appearing 
in the Markov factorization of the ad placement 
structural equation model,
\begin{eqnarray}
\label{eq-ad-markov}
   \Pa{\omega} &=& \Pa{u,v}\,\Pc{x}{u}\,
                 \Pc{a}{x,v}\,\Pc{b}{x,v}\,\Pc{q}{x,a} 
                 \nonumber\\
             & & \qquad\times~
                 \Pc{s}{a,q,b}\,\Pc{c}{a,q,b}\,
                 \Pc{y}{s,u}\,\Pc{z}{y,c} ~.
\end{eqnarray}

Variable~$y$ was defined in section~\ref{s-flow} as the set of user clicks.
In the rest of the document, we slightly abuse this notation 
by using the same letter~$y$ to represent the number of clicks. 
We also write the expectation $Y=\bbbe_{\omega\sim\Pa{\omega}}[y]$ 
using the integral notation
\[ 
    Y=\int_\omega y ~ \Pa{\omega} ~. 
\]

\begin{figure}
\center
\includegraphics[width=.55\linewidth]{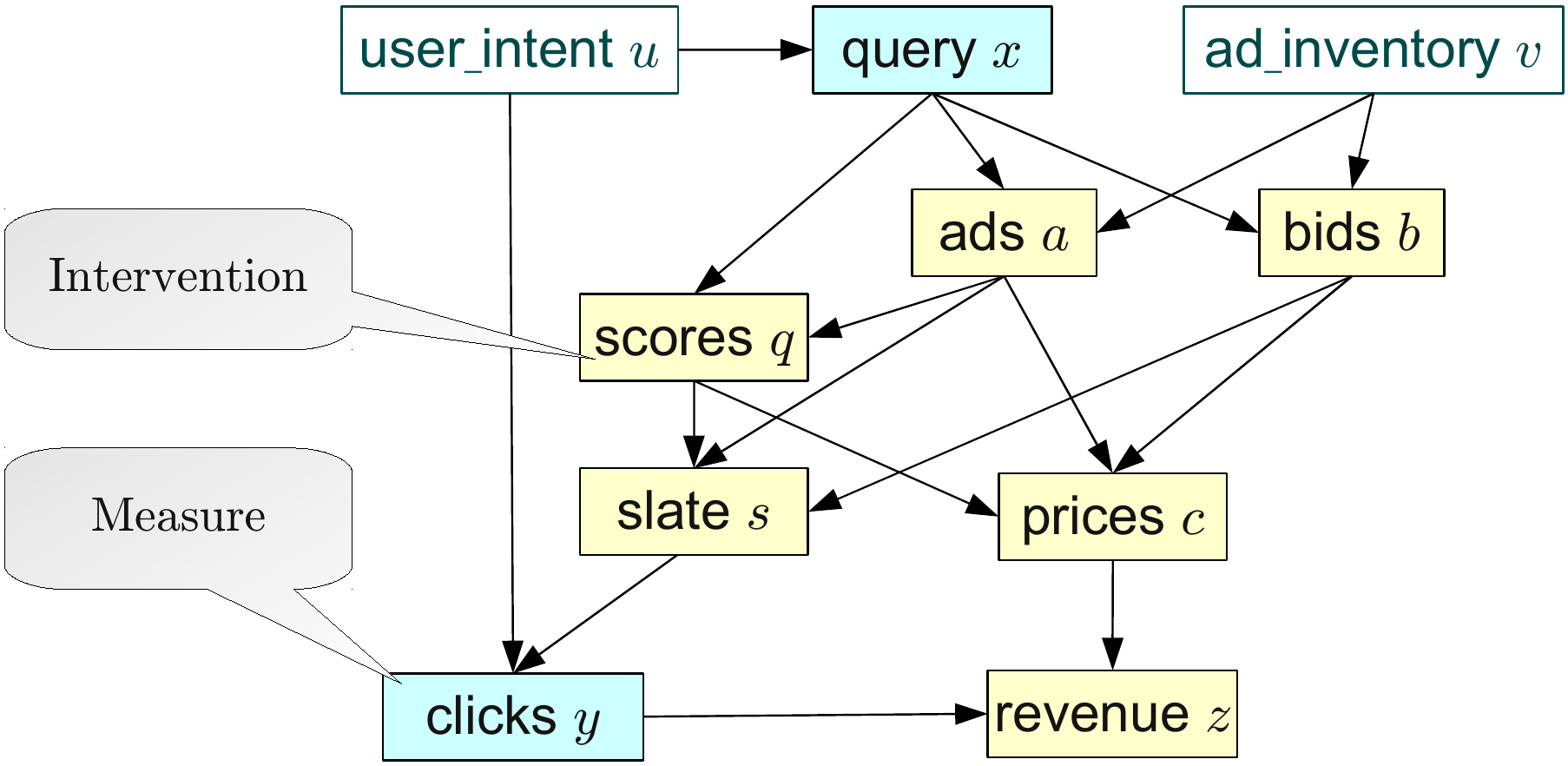}
\caption{Estimating which average number of clicks per page would have
been observed if we had used a different scoring model.}
\label{fig-semtwo}
\end{figure}

We would like to estimate what the expected click yield~$Y^*$ would have been
if we had used a different scoring function (figure~\ref{fig-semtwo}). This
intervention amounts to replacing the actual factor $\Pc{q}{x,a}$ by a
counterfactual factor $\Qc{q}{x,a}$ in the Markov factorization.
\begin{eqnarray}
\label{eq-ad-counterfactual-markov}
   \Qa{\omega} &=& \Pa{u,v}\,\Pc{x}{u}\,
                 \Pc{a}{x,v}\,\Pc{b}{x,v}\,
                 \text{\boldmath$\Qc{q}{x,a}$} 
                 \nonumber\\
               & & \qquad\times~
                 \Pc{s}{a,q,b}\,\Pc{c}{a,q,b}\,
                 \Pc{y}{s,u}\,\Pc{z}{x,c} ~.
\end{eqnarray}

Let us assume, for simplicity, that the actual factor~$\Pc{q}{x,a}$ is
nonzero everywhere. We can then estimate the counterfactual expected 
click yield $Y^*$ using the transformation
\begin{equation}
\label{eq-ad-reweighting}
   Y^* ~=~ \int_\omega y ~ \Qa{\omega} 
       ~=~ \int_\omega y ~ \frac{\Qc{q}{x,a}}{\Pc{q}{x,a}} ~ \Pa{\omega} 
 ~\approx~ \frac{1}{n} ~ \sum_{i=1}^{n} y_i ~
             \frac{\Qc{q_i}{x_i,a_i}}{\Pc{q_i}{x_i,a_i}} ~,
\end{equation}
where the data set of tuples $(a_i,x_i,q_i,y_i)$ is distributed according to
the actual Markov factorization instead of the counterfactual Markov
factorization. This data could therefore have been collected during the normal
operation of the ad placement system. Each sample is reweighted to
reflect its probability of occurrence under the counterfactual conditions.

\smallskip
In general, we can use \emph{importance sampling} to 
estimate the counterfactual expectation  
of any quantity~$\ell(\omega)$\,:
\begin{equation}
\label{eq-general-reweighting}
   Y^*
       ~=~ \int_\omega \ell(\omega) ~ \Qa{\omega}
       ~=~ \int_\omega \ell(\omega) ~ 
                \frac{\Qa{\omega}}{\Pa{\omega}} ~ \Pa{\omega} 
 ~\approx~ \frac{1}{n} \sum_{i=1}^{n} \ell(\omega_i) \: w_i
\end{equation}
with weights
\begin{equation}
\label{eq-general-weights}
  w_i ~=~ w(\omega_i) ~=~ \frac{\Qa{\omega_i}}{\Pa{\omega_i}} ~=~
  \frac{\text{factors appearing in $\Qa{\omega_i}$ 
              but not in $\Pa{\omega_i}$}}
       {\text{factors appearing in $\Pa{\omega_i}$ 
              but not in $\Qa{\omega_i}$}}~.
\end{equation}

Equation~\eqref{eq-general-weights} emphasizes the simplifications resulting
from the algebraic similarities of the actual and counterfactual Markov
factorizations. Because of these simplifications, the evaluation of the
weights only requires the knowledge of the few factors that differ between
$\Pa{\omega}$ and $\Qa{\omega}$. Each data sample needs to provide the value
of $\ell(\omega_i)$ and the values of all variables needed to evaluate
the factors that do not cancel in the ratio~\eqref{eq-general-weights}.

In contrast, the replaying approach (section~\ref{s-replay}) demands the
knowledge of all factors of $\Qa{\omega}$ connecting the point of intervention
to the point of measurement~$\ell(\omega)$.  On the other hand, it does not
require the knowledge of factors appearing only in~$\Pa{\omega}$.

\smallskip
Importance sampling relies on the assumption that all the factors appearing in
the denominator of the reweighting ratio~\eqref{eq-general-weights} are
nonzero whenever the factors appearing in the numerator are nonzero.  Since
these factors represents conditional probabilities resulting from the effect
of an independent noise variable in the structural equation model, this
assumption means that the data must be collected with an experiment involving
active randomization. We must therefore design cost-effective randomized
experiments that yield enough information to estimate many interesting
counterfactual expectations with sufficient accuracy.  This problem cannot be
solved without answering the confidence interval question: given data
collected with a certain level of randomization, with which accuracy can we
estimate a given counterfactual expectation?


\subsection{Confidence Intervals}
\label{s-confidence}

At first sight, we can invoke the law of large numbers and write
\begin{equation}
\label{eq-naive-mc}
   Y^* = \int_\omega \ell(\omega)\,w(\omega)~\Pa{\omega}
     ~~\approx~~  \frac{1}{n} \sum_{i=1}^{n} \ell(\omega_i)\,w_i \,.
\end{equation}
For sufficiently large~$n$, the central limit theorem provides 
confidence intervals whose width grows with the standard deviation
of the product $\ell(\omega)\,w(\omega)$.

Unfortunately, when $\Pa{\omega}$ is small, the reweighting
ratio~$w(\omega)$ takes large values with low probability.  This heavy
tailed distribution has annoying consequences because the variance of
the integrand could be very high or infinite.  When the variance is
infinite, the central limit theorem does not hold.  When the variance
is merely very large, the central limit convergence might occur too
slowly to justify such confidence intervals. Importance sampling works
best when the actual distribution and the counterfactual distribution
overlap.

When the counterfactual distribution has significant mass in domains
where the actual distribution is small, the few samples available in
these domains receive very high weights. Their noisy contribution
dominates the reweighted estimate~\eqref{eq-naive-mc}. We can obtain
better confidence intervals by eliminating these few samples drawn in
poorly explored domains. The resulting bias can be bounded using prior
knowledge, for instance with an assumption about the range of values
taken by $\ell(\omega)$,
\begin{equation}
\label{eq-bounded-integrand}
  \forall \omega \quad \ell(\omega)~\in~\inBrack{\,0,\,M}\,.
\end{equation}

Let us choose the maximum weight value $R$ deemed acceptable for the
weights. We have obtained very consistent results in practice with~$R$
equal to the fifth largest reweighting ratio observed on the empirical
data.\footnote{This is in fact a slight abuse because the theory calls
for choosing $R$ before seing the data.}  We can then rely
on \emph{clipped weights} to eliminate the contribution of
the poorly explored domains,
\[ 
   \capw(\omega) 
      ~=~ \left\{\begin{array}{ll} 
          w(\omega) & \text{if $\Qa{\omega} < R\:\Pa{\omega}$} \\
          0 & \text{otherwise.}
      \end{array}\right.
\]
The condition $\Qa{\omega}<R\:\Pa{\omega}$ ensures that the ratio 
has a nonzero denominator $\Pa{\omega}$ and is smaller than $R$.
Let $\Omega_R$ be the set of all values of $\omega$ 
associated with acceptable ratios:
\[
 \Omega_R ~=~ \left\{ \, \omega: ~ \Qa{\omega} < R\:\Pa{\omega} \, \right\} \,.
\]
We can decompose $Y^*$ in two terms:
\begin{equation}
\label{eq-decomp-ystar}
  Y^* ~=~ \int_{\omega\in\Omega_R} \hskip-1em\ell(\omega)\,\Qa{\omega}
      ~+~ \int_{\omega\in\Omega\setminus\Omega_R} \hskip-1.5em\ell(\omega)\,\Qa{\omega} 
      ~=~ \capY^* + \inPar{Y^*-\capY^*} \,.
\end{equation}

The first term of this decomposition is the \emph{clipped expectation}
$\capY^*$.  Estimating the clipped expectation $\capY^*$ is much
easier than estimating $Y^*$ from~\eqref{eq-naive-mc} because the
clipped weights $\capw(\omega)$ are bounded by $R$.
\begin{equation}
\label{eq-capped-mc}
   \capY^* ~=~ \int_{\omega\in\Omega_R} \hskip-1em\ell(\omega)\,\Qa{\omega}
           ~=~ \int_\omega \ell(\omega)\,\capw(\omega)~\Pa{\omega} ~~\approx~~ 
    \estY^* ~=~ \frac{1}{n} \sum_{i=1}^{n} \ell(\omega_i)\,\capw(\omega_i) \,.
\end{equation}

The second term of equation~\eqref{eq-decomp-ystar} can be bounded
by leveraging assumption \eqref{eq-bounded-integrand}. The resulting 
bound can then be conveniently estimated using only the clipped weights.
\[
  Y^*-\capY^* 
     ~=~ \int_{\omega\in\Omega\setminus\Omega_R} \hskip-1.5em\ell(\omega)\,\Qa{\omega}
     ~~\in~~ \Big[\,0,~M \, \Qa{\Omega\setminus\Omega_R}\,\Big]
     ~=~ \Big[\,0,~M\,\inPar{1-\capW^*}\,\Big] \quad\text{with}
\]
\begin{equation}
\label{eq-capped-w}
   \capW^* ~=~ \Qa{\Omega_R} 
           ~=~ \int_{\omega\in\Omega_R} \hskip-1em\Qa{\omega}
           ~=~ \int_\omega \capw(\omega)\,\Pa{\omega} ~~\approx~~
   \estW^* ~=~ \frac{1}{n} \sum_{i=1}^{n} \capw(\omega_i) \,.
\end{equation}

Since the clipped weights are bounded, the estimation errors
associated with~\eqref{eq-capped-mc} and~\eqref{eq-capped-w} are well
characterized using either the central limit theorem or using empirical
Bernstein bounds (see appendix~\ref{a-confidence} for details).
Therefore we can derive an \emph{outer confidence interval}
of the form
\begin{equation}
  \label{eq-conf-outer}
  \bbbP\inBrace{ ~ \estY^* -\epsilon_R ~\le~ \capY^* \:\le~ 
    \estY^*+\epsilon_R ~ } ~\ge~ 1-\delta 
\end{equation}
and an \emph{inner confidence interval} of the form
\begin{equation}
\label{eq-conf-inner}
  \bbbP\inBrace{ ~ \capY^* ~\le~ Y^* \:\le~ \capY^* + M (1 - \estW^* + \xi_R) ~ } 
     ~\ge~ 1-\delta \,.
\end{equation}
The names \emph{inner} and \emph{outer} are in fact related to our
prefered way to visualize these intervals (\eg., figure~\ref{fig-exp-mlr}).
Since the bounds on $Y^*-\capY^*$ can be written as
\begin{equation}
\label{eq-bias-bound}
  \capY^* ~\le~ Y^* ~\le~ \capY^* + M\,\inPar{1-\capW^*}\,,
\end{equation}
we can derive our final confidence interval,
\begin{equation}
\label{eq-conf-dual}
   \bbbP\inBrace{ ~ \estY^* - \epsilon_R ~\le~ Y^* \:\le~
       \estY^* + M(1 - \estW^* + \xi_R) + \epsilon_R ~ } \ge 1-2\delta \,.
\end{equation}

In conclusion, replacing the unbiased importance sampling
estimator~\eqref{eq-naive-mc} by the clipped importance sampling
estimator~\eqref{eq-capped-mc} with a suitable choice of $R$ leads to
improved confidence intervals. Furthermore, since the derivation of
these confidence intervals does not rely on the assumption that
$\Pa{\omega}$ is nonzero everywhere, the clipped importance sampling
estimator remains valid when the distribution $\Pa{\omega}$ has a
limited support.  This relaxes the main restriction associated with
importance sampling.


\subsection{Interpreting the Confidence Intervals}
\label{s-interpreting-confidence}

The estimation of the counterfactual expectation~$Y^*$ can be
inaccurate because the sample size is insufficient or because the
sampling distribution~$\Pa{\omega}$ does not sufficiently explore the
counterfactual conditions of interest. 

By construction, the clipped expectation~$\capY^*$ ignores the domains
poorly explored by the sampling distribution~$\Pa{\omega}$.  The
difference $Y^*-\capY^*$ then reflects the inaccuracy resulting from a
lack of exploration.  Therefore, assuming that the bound~$R$ has been
chosen competently, the relative sizes of the outer and inner
confidence intervals provide precious cues to determine whether we can
continue collecting data using the same experimental setup or should
adjust the data collection experiment in order to obtain a better
coverage.
\begin{itemize}
\item
  The \emph{inner confidence interval} \eqref{eq-conf-inner} witnesses
  the uncertainty associated with the domain~$G_R$ insufficiently
  explored by the actual distribution. A large inner confidence
  interval suggests that the most practical way to improve the
  estimate is to adjust the data collection experiment in order to
  obtain a better coverage of the counterfactual conditions of
  interest.
\item
  The \emph{outer confidence interval} \eqref{eq-conf-outer}
  represents the uncertainty that results from the limited sample
  size. A large outer confidence interval indicates that the sample is
  too small. To improve the result, we simply need to continue
  collecting data using the same experimental setup.
\end{itemize}


\subsection{Experimenting with Mainline Reserves}
\label{s-exp-mlr}

We return to the ad placement problem to illustrate the reweighting approach
and the interpretation of the confidence intervals.  Manipulating the reserves
$R_p(x)$ associated with the mainline positions (figure~\ref{fig-adlocations})
controls which ads are prominently displayed in the mainline or displaced into
the sidebar. 

\smallskip

We seek in this section to answer counterfactual questions of the form: 
\begin{quotation}
\hyphenpenalty2000
\noindent\llap{``}\emph{\relax
  How would the ad placement system have performed if
  we had scaled the mainline reserves by a constant factor $\rho$,
  without incurring user or advertiser reactions?}''
\end{quotation}

Randomization was introduced using a modified version of the ad placement
engine. Before determining the ad layout (see section~\ref{s-adplacement}), a
random number~$\varepsilon$ is drawn according to the standard normal
distribution $\normalDist{0}{1}$, and all the mainline reserves are multiplied
by~$m=\rho\,e^{-\sigma^2/2+\sigma\varepsilon}$. Such multipliers follow a
log-normal distribution\footnote{More precisely,
  $\lognormalDist{\mu}{\sigma^2}$ with $\mu=\sigma^2/2+\log\rho$.}
whose mean is~$\rho$ and whose width is controlled by $\sigma$. 
This effectively provides a parametrization of the conditional score
distribution~$\Pc{q}{x,a}$ (see figure~\ref{fig-markov}.)

The Bing search platform offers many ways to select traffic for controlled
experiments (section~\ref{s-flighting}). In order to match our isolation
assumption, individual page views were randomly assigned to traffic buckets
without regard to the user identity. The main treatment bucket was processed
with mainline reserves randomized by a multiplier drawn as explained above
with $\rho\!=\!1$ and $\sigma\!=\!0.3$. With these parameters, the mean
multiplier is exactly~1, and~$95\%$ of the multipliers are in
range~$[0.52,1.74]$. Samples describing 22 million search result pages were
collected during five consecutive weeks.

\begin{figure}[p]
\center
\includegraphics[width=.54\linewidth]{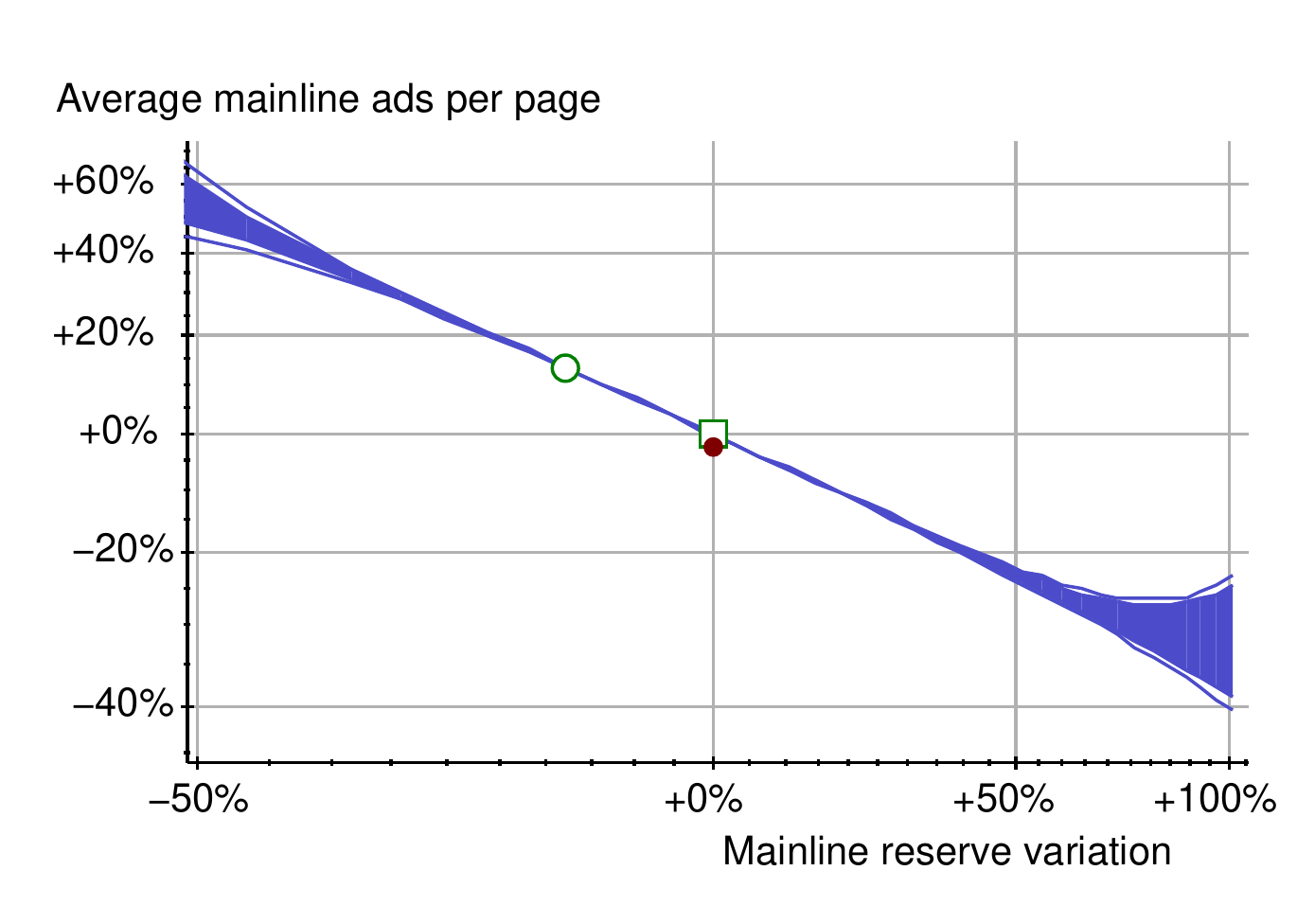}\\[-2.2ex]
\includegraphics[width=.54\linewidth]{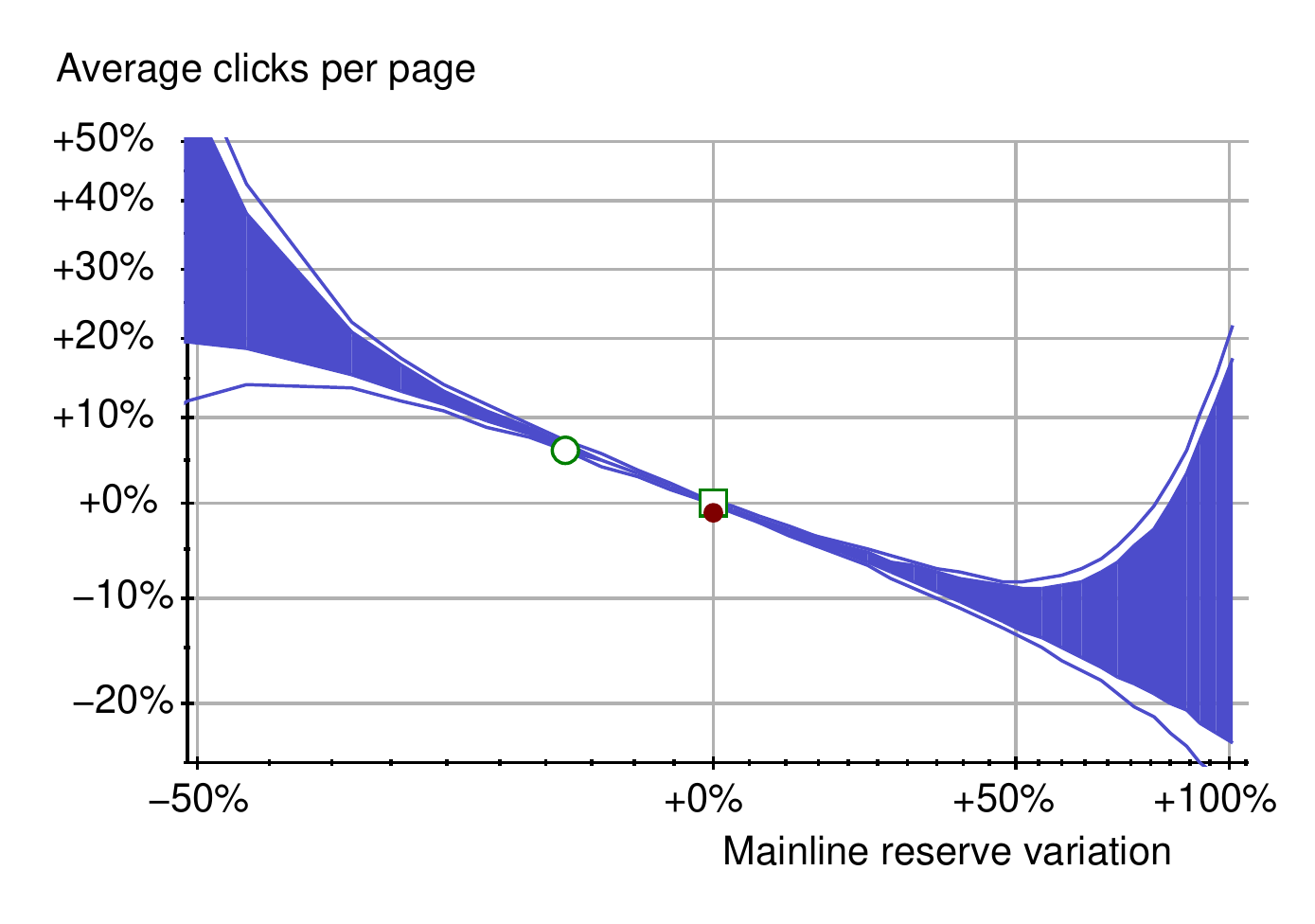}\\[-2.2ex]
\includegraphics[width=.54\linewidth]{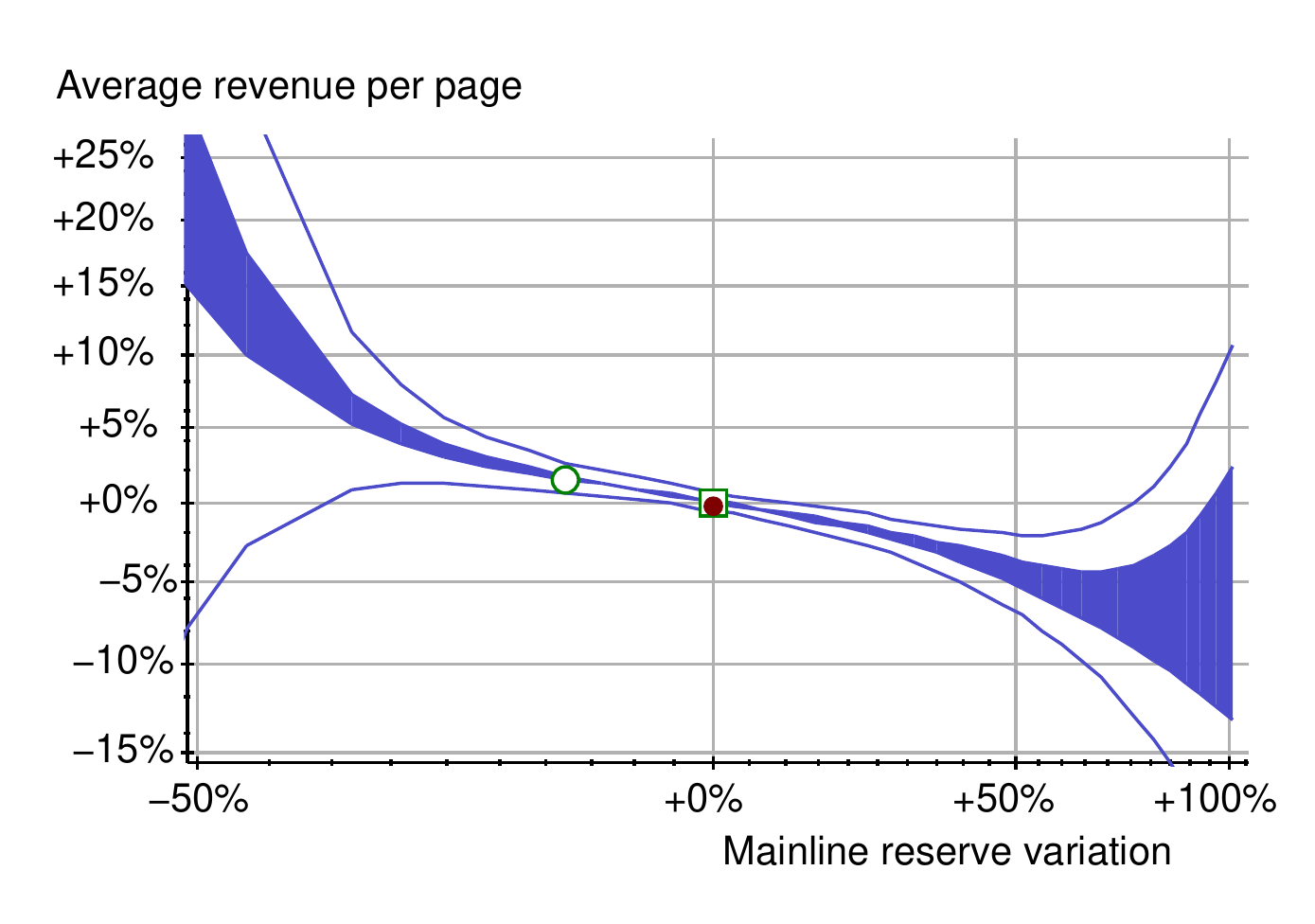}\\[-2.2ex]
\caption{\label{fig-exp-mlr} Estimated variations of three
  performance metrics in response to mainline reserve changes.
  The curves delimit $95\%$ confidence intervals for the
  metrics we would have observed if we had increased the mainline reserves by
  the percentage shown on the horizontal axis. The filled areas represent the
  inner confidence intervals.  The hollow squares represent the metrics
  measured on the experimental data. The hollow circles
  represent metrics measured on a second experimental bucket
  with mainline reserves reduced by~18\%. The filled circles represent
  the metrics effectively measured on a control bucket running without
  randomization.}
\end{figure}

We then use this data to estimate what would have been measured if the
mainline reserve multipliers had been drawn according to
a distribution determined by parameters~$\rho^*$ and~$\sigma^*$.
This is achieved by reweighting each sample $\omega_i$ with
\[ 
   w_i  = \frac{\Qc{q_i}{x_i,a_i}}{\Pc{q_i}{x_i,a_i}} 
        = \frac{p(m_i\,;\,\rho^*,\,\sigma^*)}{p(m_i\,;\,\rho,\,\sigma)} \,, 
\]
where~$m_i$ is the multiplier drawn for this sample during the data 
collection experiment, and~$p(t\,;\,\rho,\sigma)$ is the density 
of the log-normal multiplier distribution.

Figure~\ref{fig-exp-mlr} reports results obtained by varying $\rho^*$
while keeping $\sigma^*\!=\!\sigma$. This amounts to estimating what
would have been measured if all mainline reserves had been multiplied
by~$\rho^*$ while keeping the same randomization. The curves bound
95\% confidence intervals on the variations of the average number of
mainline ads displayed per page, the average number of ad clicks per
page, and the average revenue per page, as functions of $\rho^*$. The
inner confidence intervals, represented by the filled areas, grow
sharply when $\rho^*$ leaves the range explored during the data
collection experiment. The average revenue per page has more variance
because a few very competitive queries command high prices.

In order to validate the accuracy of these counterfactual estimates, a second
traffic bucket of equal size was configured with mainline reserves reduced by
about $18\%$. The hollow circles in figure~\ref{fig-exp-mlr} represent
the metrics effectively measured on this bucket during the same time period.
The effective measurements and the counterfactual estimates match with high
accuracy.

Finally, in order to measure the cost of the randomization, we also ran the
unmodified ad placement system on a control bucket. The brown filled circles
in figure~\ref{fig-exp-mlr} represent the metrics effectively measured on the
control bucket during the same time period. The randomization caused a small
but statistically significant increase of the number of mainline ads per
page. The click yield and average revenue differences are not significant.

This experiment shows that we can obtain accurate counterfactual estimates
with affordable randomization strategies. However, this nice conclusion does
not capture the true practical value of the counterfactual estimation
approach.


\subsection{More on Mainline Reserves}

The main benefit of the counterfactual estimation approach is the ability to
\emph{use the same data} to answer a \emph{broad range of counterfactual}
\emph{questions}. Here are a few examples of counterfactual questions
that can be answered using data collected using the simple mainline reserve
randomization scheme described in the previous section:

\begin{itemize}
\item 
  \emph{Different variances} -- Instead of estimating what would have been
  measured if we had increased the mainline reserves without changing the
  randomization variance, that is, letting $\sigma^*=\sigma$, we can use the
  same data to estimate what would have been measured if we had also changed
  $\sigma$.  This provides the means to determine which level of randomization
  we can afford in future experiments.
\item
  \emph{Pointwise estimates} -- We often want to estimate what would have been
  measured if we had set the mainline reserves to a specific value without
  randomization. Although computing estimates for small values of $\sigma$
  often works well enough, very small values lead to large confidence
  intervals.

  Let $Y_\nu(\rho)$ represent the expectation we would have observed 
  if the multipliers $m$ had mean $\rho$ and variance $\nu$. 
  We have then $Y_\nu(\rho)=\bbbe_m[\:\bbbe[y|m]\:] = \bbbe_m[Y_0(m)]$.
  Assuming that the  pointwise value $Y_0$ is smooth enough 
  for a second order development,
  \[
      \def\mmr{{(m\!-\!\rho)}}
      Y_\nu(\rho) 
      ~\approx~ \bbbe_m\big[\:Y_0(\rho)
                  +\mmr Y^\prime_0(\rho) 
                  +\mmr^2 Y^\pprime_0(\rho)/2\:\big]
      ~=~ Y_0(\rho) + \nu Y^\pprime_0(\rho)/2 ~.
  \]
  Although the reweighting method cannot estimate the point-wise value
  $Y_0(\rho)$ directly, we can use the reweighting method to estimate
  both~$Y_\nu(\rho)$ and~$Y_{2\nu}(\rho)$ with acceptable confidence intervals
  and write $Y_0(\rho) \approx 2Y_\nu(\rho)-Y_{2\nu}(\rho)$
  ~\citep{goodwin-2011}.
  
\item
  \emph{Query-dependent reserves} -- Compare for instance the queries
  ``car~insurance'' and ``common~cause~principle'' in a web search engine.
  Since the advertising potential of a search varies considerably with the
  query, it makes sense to investigate various ways to define query-dependent
  reserves~\citep{charles-2012b}.

  The data collected using the simple mainline reserve randomization can also
  be used to estimate what would have been measured if we had increased all
  the mainline reserves by a query-dependent multiplier $\rho^*(x)$. This is
  simply achieved by reweighting each sample $\omega_i$ with
  \[ 
    w_i = \frac{\Qc{q_i}{x_i,a_i}}{\Pc{q_i}{x_i,a_i}} 
        = \frac{p(m_i\,;\,\rho^*(x_i)\,,\,\sigma)}{p(m_i\,;\,\mu,\,\sigma)}\,.
  \]
\end{itemize}

Considerably broader ranges of counterfactual questions can be
answered when data is collected using randomization schemes that
explore more dimensions. For instance, in the case of the ad placement
problem, we could apply an independent random multiplier for each
score instead of applying a single random multiplier to the mainline
reserves only. However, the more dimensions we randomize, the more
data needs to be collected to effectively explore all these
dimensions. Fortunately, as discussed in section \ref{s-structure},
the structure of the causal graph reveals many ways to leverage a
priori information and improve the confidence intervals.


\subsection{Related Work}

Importance sampling is widely used to deal with covariate
shifts~\citep{shimodaira-2000,sugiyama-2007}. Since manipulating the
causal graph changes the data distribution, such an intervention can
be viewed as a covariate shift amenable to importance sampling.
Importance sampling techniques have also been proposed without causal
interpretation for many of the problems that we view as causal
inference problems. In particular, the work presented in this section
is closely related to the Monte-Carlo approach of reinforcement
learning~\citep[chapter~5]{sutton-barto-1998} and to the offline
evaluation of contextual bandit policies~\citep{li-2010,li-2011}.

Reinforcement learning research traditionally focuses on control
problems with relatively small discrete state spaces and long
sequences of observations. This focus reduces the need for
characterizing exploration with tight confidence intervals. For
instance, \citeauthor{sutton-barto-1998} suggest to normalize the
importance sampling estimator by~$1/\sum_i\:w(\omega_i)$ instead
of~$1/n$. This would give erroneous results when the data collection
distribution leaves parts of the state space poorly
explored. Contextual bandits are traditionally formulated with a
finite set of discrete actions. For instance, Li's
(\citeyear{li-2011}) unbiased policy evaluation assumes that the data
collection policy always selects an arbitrary policy with probability
greater than some small constant. This is not possible when the action
space is infinite.

Such assumptions on the data collection distribution are often
impractical.  For instance, certain ad placement policies are not
worth exploring because they cannot be implemented efficiently or are
known to elicit fraudulent behaviors. There are many practical
situations in which one is only interested in limited aspects of the
ad placement policy involving continuous parameters such as click
prices or reserves. Discretizing such parameters eliminates useful a
priori knowledge: for instance, if we slightly increase a reserve, we
can reasonable believe that we are going to show slightly less
ads. 

Instead of making assumptions on the data collection distribution, we
construct a biased estimator~\eqref{eq-capped-mc} and bound its
bias. We then interpret the inner and outer confidence intervals as
resulting from a lack of exploration or an insufficient sample size.

Finally, the causal framework allows us to easily formulate
counterfactual questions that pertain to the practical ad placement
problem and yet differ considerably in complexity and exploration
requirements. We can address specific problems identified by the
engineers without incurring the risks associated with a complete
redesign of the system. Each of these incremental steps helps
demonstrating the soundness of the approach.


\section{Structure}
\label{s-structure}

This section shows how the structure of the causal graph reveals many
ways to leverage a priori knowledge and improve the accuracy of our
counterfactual estimates. Displacing the reweighting point
(section~\ref{s-slates}) improves the inner confidence interval and
therefore reduce the need for exploration. Using a prediction function
(section~\ref{s-predictors}) essentially improve the outer confidence
interval and therefore reduce the sample size requirements.


\subsection{Better Reweighting Variables}
\label{s-slates}

Many search result pages come without eligible ads. We then know with
certainty that such pages will have zero mainline ads, receive zero
clicks, and generate zero revenue. This is true for the randomly
selected value of the reserve, and this would have been true for any
other value of the reserve. We can exploit this knowledge by
pretending that the reserve was drawn from the counterfactual
distribution $\Qc{q}{x_i,a_i}$ instead of the actual distribution
$\Pc{q}{x_i,a_i}$. The ratio~$w(\omega_i)$ is therefore forced to
the unity. This does not change the estimate but reduces the size of
the inner confidence interval. The results of figure~\ref{fig-exp-mlr}
were in fact helped by this little optimization.

There are in fact many circumstances in which the observed outcome
would have been the same for other values of the randomized
variables. This prior knowledge is in fact encoded in the structure of
the causal graph and can be exploited in a more systematic manner.
For instance, we know that users make click decisions without knowing
which scores were computed by the ad placement engine, and without
knowing the prices charged to advertisers. The ad placement causal
graph encodes this knowledge by showing the clicks $y$ as direct
effects of the user intent $u$ and the ad slate $s$. This implies that
the exact value of the scores $q$ does not matter to the clicks $y$ as
long as the ad slate $s$ remains the same.

Because the causal graph has this special structure, we can simplify
both the actual and counterfactual Markov factorizations
\eqref{eq-ad-markov}\,\eqref{eq-ad-counterfactual-markov} without
eliminating the variable $y$ whose expectation is sought.
Successively eliminating variables $z$, $c$, and $q$ gives:
\begin{eqnarray*}
   \Pa{u,v,x,a,b,s,y} &=& \Pa{u,v}\,\Pc{x}{u}\,\Pc{a}{x,v}\,
                          \Pc{b}{x,v}\,\Pc{s}{x,a,b}\,\Pc{y}{s,u} ~,\\
   \Qa{u,v,x,a,b,s,y} &=& \Pa{u,v}\,\Pc{x}{u}\,\Pc{a}{x,v}\,
                          \Pc{b}{x,v}\,\Qc{s}{x,a,b}\,\Pc{y}{s,u} ~.
\end{eqnarray*}
The conditional distributions $\Pc{s}{x,a,b}$ and $\Qc{s}{x,a,b}$
did not originally appear in the Markov factorization.
They are defined by marginalization as a consequence
of the elimination of the variable $q$ representing the scores.
\[
    \Pc{s}{x,a,b}=\int_q \Pc{s}{a,q,b}\:\Pc{q}{x,a} ~,\quad
    \Qc{s}{x,a,b}=\int_q \Pc{s}{a,q,b}\:\Qc{q}{x,a} \,.
\]

\begin{figure}[t]
\center
\includegraphics[width=.54\linewidth]{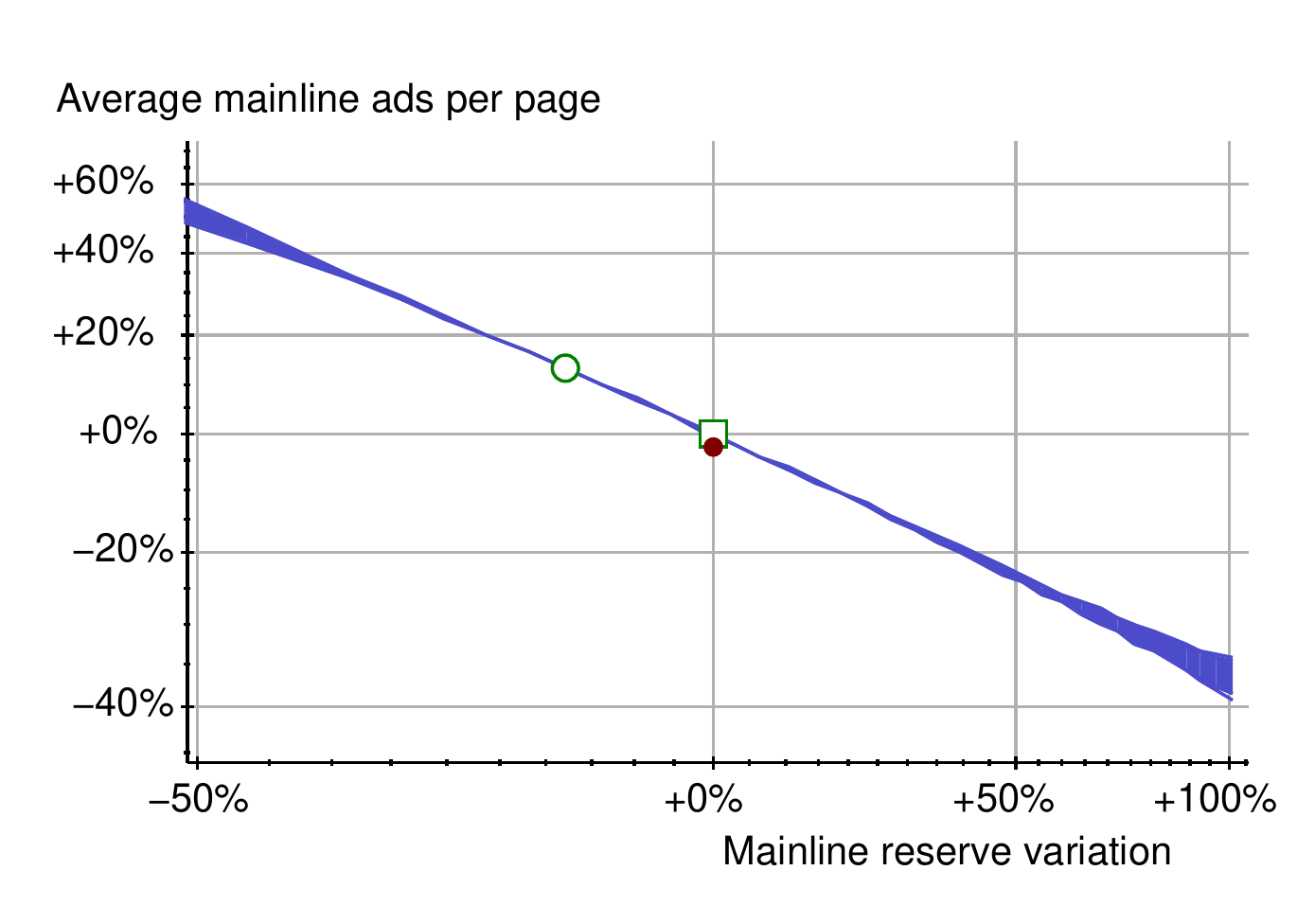}\\[-2.2ex]
\includegraphics[width=.54\linewidth]{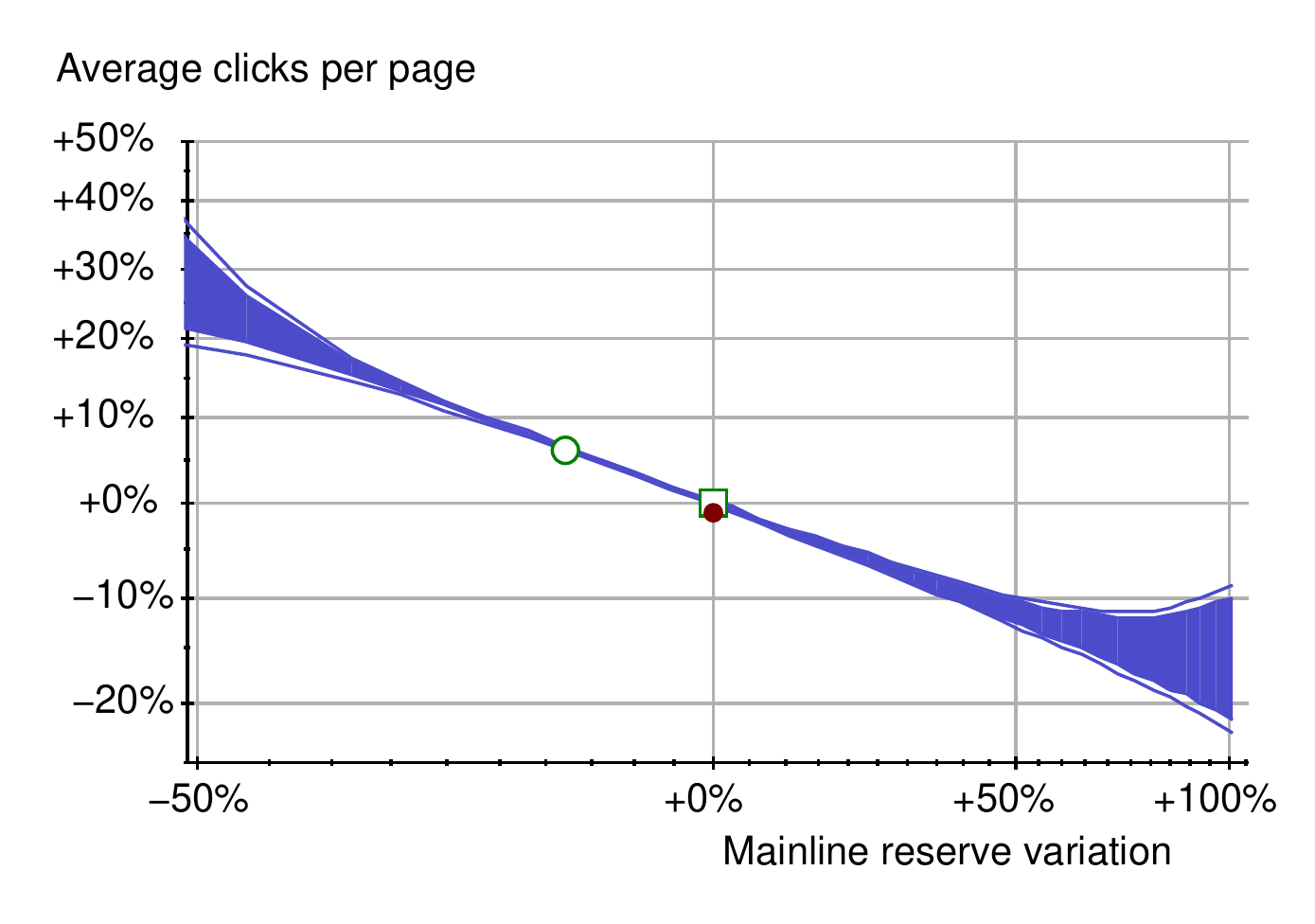}
\caption{\label{fig-exp-mlrtwo} Estimated variations of two performance
  metrics in response to mainline reserve changes. These estimates were
  obtained using the ad slates $s$ as reweighting variable. Compare the inner
  confidence intervals with those shown in figure~\ref{fig-exp-mlr}. }
\end{figure}

We can estimate the counterfactual click yield~$Y^*$ 
using these simplified factorizations:
\begin{eqnarray}
\label{eq-slate-reweighting}
Y^* &=& \int 
             y~\Qa{u,v,x,a,b,s,y}
    ~=~ \int 
             y~\frac{\Qc{s}{x,a,b}}{\Pc{s}{x,a,b}}~
               \Pa{u,v,x,a,b,s,y}\nonumber\\
    &\approx& \frac{1}{n}\sum_{i=1}^{n} ~ y_i\,
               \frac{\Qc{s_i}{x_i,a_i,b_i}}{\Pc{s_i}{x_i,a_i,b_i}}~.
\end{eqnarray}

We have reproduced the experiments described in section~\ref{s-exp-mlr} with 
the counterfactual estimate~\eqref{eq-slate-reweighting} instead
of~\eqref{eq-ad-reweighting}. For each example $\omega_i$, we determine
which range $[m^{\max}_i,m^{\min}_i]$ of mainline reserve multipliers could
have produced the observed ad slate~$s_i$, and then compute the reweighting 
ratio using the formula:
\[
  w_i = \frac{\Qc{s_i}{x_i,a_i,b_i}}{\Pc{s_i}{x_i,a_i,b_i}}
  = \frac{\Psi(m^{\max}_i;\,\rho^*,\sigma^*)-\Psi(m^{\min}_i;\,\rho^*,\sigma^*)}
         {\Psi(m^{\max}_i;\,\rho,\sigma)-\Psi(m^{\min}_i;\,\rho,\sigma)}~,
\]
where $\Psi(m;\rho,\sigma)$ is the cumulative of the log-normal
multiplier distribution. Figure~\ref{fig-exp-mlrtwo} shows
counterfactual estimates obtained using the same data as
figure~\ref{fig-exp-mlr}. The obvious improvement of the inner
confidence intervals significantly extends the range of mainline
reserve multipliers for which we can compute accurate counterfactual
expectations using this same data.

\begin{figure}[t]
\center
\includegraphics[width=.55\linewidth]{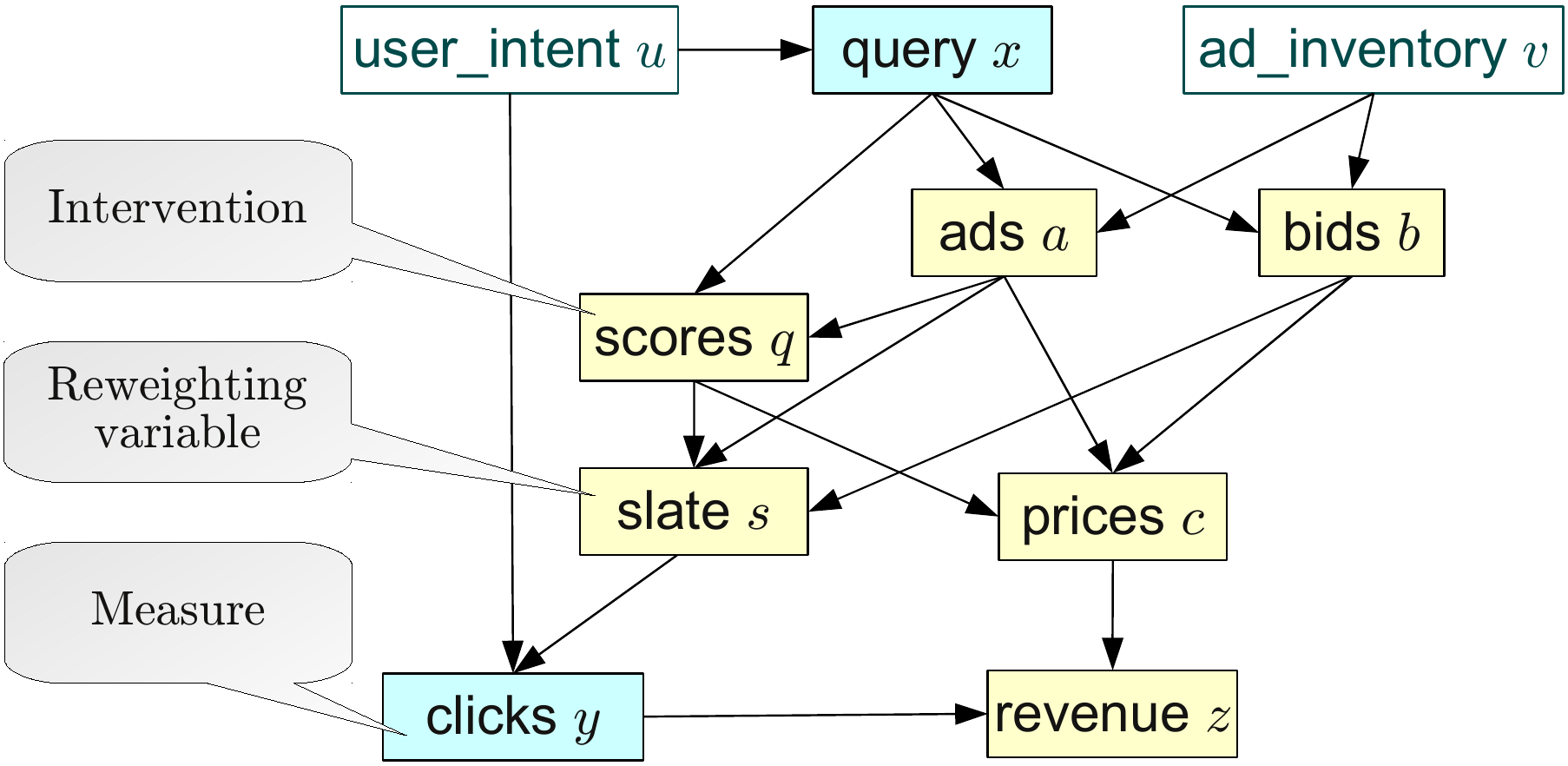}
\caption{The reweighting variable(s) must intercept
  all causal paths from the point of intervention
  to the point of measurement.}
\label{fig-semthree}
\end{figure}

Comparing~\eqref{eq-ad-reweighting} and~\eqref{eq-slate-reweighting}
makes the difference very clear: instead of computing the ratio of the
probabilities of the observed scores under the counterfactual and
actual distributions, we compute the ratio of the probabilities of the
observed ad slates under the counterfactual and actual
distributions. As illustrated by figure~\ref{fig-semthree}, we now
distinguish the reweighting variable (or variables) from the
intervention. In general, the corresponding manipulation of the Markov
factorization consists of marginalizing out all the variables that
appear on the causal paths connecting the point of intervention to the
reweighting variables and factoring all the independent terms out of
the integral. This simplification works whenever the reweighting
variables intercept all the causal paths connecting the point of
intervention to the measurement variable. In order to compute the new
reweighting ratios, all the factors remaining inside the integral,
that is, all the factors appearing on the causal paths connecting the
point of intervention to the reweighting variables, have to be known.

\begin{figure}
\center
\includegraphics[width=.75\linewidth]{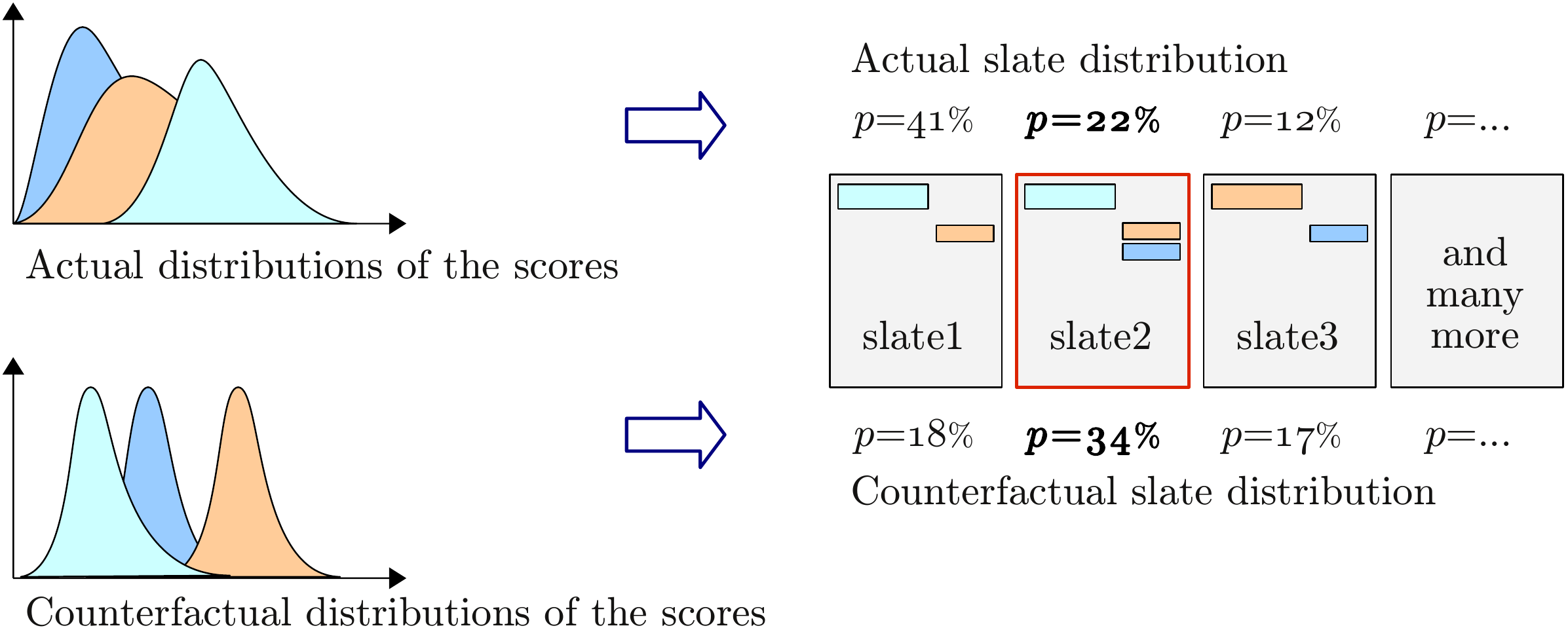}
\caption{A distribution on the scores $q$ induce 
  a distribution on the possible ad slates $s$. 
  If the observed slate is \texttt{slate2}, 
  the reweighting ratio is $34/22$.}
\label{fig-slatedist}
\end{figure}

Figure~\ref{fig-exp-mlrtwo} does not report the average revenue per
page because the revenue~$z$ also depends on the scores~$q$ through
the click prices~$c$. This causal path is not intercepted by the ad
slate variable~$s$ alone. However, we can introduce a new variable
$\tilde{c}=f(c,y)$ that filters out the click prices computed for ads
that did not receive a click. Markedly improved revenue estimates are
then obtained by reweighting according to the joint variable
$(s, \tilde{c})$.

Figure~\ref{fig-slatedist} illustrates the same approach applied to
the simultaneous randomization of all the scores $q$ using independent
log-normal multipliers. The weight $w(\omega_i)$ is the ratio of
the probabilities of the observed ad slate $s_i$ under the
counterfactual and actual multiplier distributions. Computing these
probabilities amounts to integrating a multivariate Gaussian
distribution~\citep{genz-1992}. Details will be provided in a
forthcoming publication.


\subsection{Variance Reduction with Predictors}
\label{s-predictors}

Although we do not know exactly how the variable of interest
$\ell(\omega)$ depends on the measurable variables and are affected by
interventions on the causal graph, we may have strong a priori
knowledge about this dependency. For instance, if we augment the
slate~$s$ with an ad that usually receives a lot of clicks, we can
expect an increase of the number of clicks.

Let the \emph{invariant variables} $\upsilon$ be all observed
variables that are not direct or indirect effects of variables
affected by the intervention under consideration. This definition
implies that the distribution of the invariant variables is not
affected by the intervention. Therefore the values $\upsilon_i$ of the
invariant variables sampled during the actual experiment are also
representative of the distribution of the invariant variables under the
counterfactual conditions.

We can leverage a priori knowledge to construct a
predictor~$\zeta(\omega)$ of the quantity~$\ell(\omega)$ whose
counterfactual expectation $Y^*$ is sought. We assume that the
predictor $\zeta(\omega)$ depends only on the invariant variables or
on variables that depend on the invariant variables through known
functional dependencies. Given sampled values $\upsilon_i$ of the
invariant variables, we can replay both the original and manipulated
structural equation model as explained in section~\ref{s-replay} and
obtain samples $\zeta_i$ and $\zeta^*_i$ that respectively follow the
actual and counterfactual distributions

\smallskip
Then, regardless of the quality of the predictor, 
\begin{eqnarray}
   Y^* ~=~ \int_\omega \ell(\omega)\,\Qa{\omega} 
    & = & 
     \int_\omega \zeta(\omega)\,\Qa{\omega} 
     ~+~ \int_\omega \inPar{\ell(\omega)-\zeta(\omega)}\,\Qa{\omega}  \nonumber\\
    & \approx & \frac{1}{n}\sum_{i=1}^{n} \zeta^*_i
     ~+~ \frac{1}{n}\sum_{i=1}^{n} \inPar{\ell(\omega_i)-\zeta_i}\,w(\omega_i)\,.
  \label{eq-doubly-robust}
\end{eqnarray}
The first term in this sum represents the counterfactual expectation
of the predictor and can be accurately estimated by averaging the
simulated counterfactual samples $\zeta_i^*$ without resorting to
potentially large importance weights.  The second term in this sum
represents the counterfactual expectation of the residuals
$\ell(\omega)-\zeta(\omega)$ and must be estimated using importance
sampling.  Since the magnitude of the residuals is hopefully smaller
than that of $\ell(\omega)$, the variance of
$\inPar{\ell(\omega)-\zeta(\omega)}\,w(\omega)$ is reduced and the
importance sampling estimator of the second term has improved
confidence intervals. The more accurate the predictor $\zeta(\omega)$,
the more effective this variance reduction strategy.

\begin{figure}
\center
\includegraphics[width=.6\linewidth]{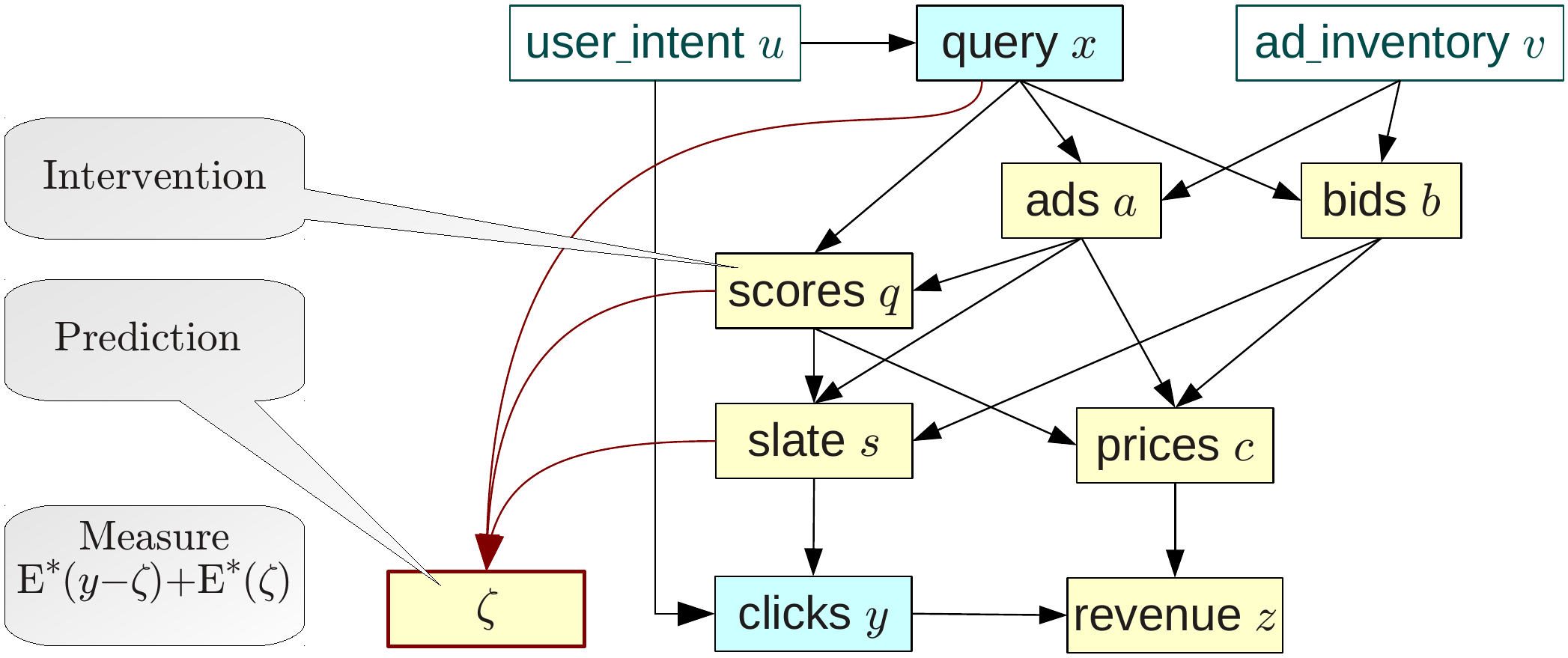}
\caption{\label{fig-semp} Leveraging a predictor. Yellow nodes
represent known functional relations in the structural equation model.
We can estimate the counterfactual expectation
$Y^*$ of the number of clicks per page as the sum of the
counterfactual expectations of a predictor $\zeta$, which is easy to
estimate by replaying empirical data, and $y-\zeta$, which has to be
estimated by importance sampling but has reduced variance.}
\end{figure}

This variance reduction technique is in fact identical to the doubly
robust contextual bandit evaluation technique
of~\citet{dudik-2012}. Doubly robust variance reduction has also been
extensively used for causal inference applied to
biostatistics~\citep[see][]{robins-2000,bang-robins-2005}. We
subjectively find that viewing the predictor as a component of the
causal graph (figure~\ref{fig-semp}) clarifies how a well designed
predictor can leverage prior knowledge. For instance, in order to
estimate the counterfactual performance of the ad placement system, we
can easily use a predictor that runs the ad auction and simulate the
user clicks using a click probability model trained offline.

\begin{figure}
\center
\includegraphics[width=.6\linewidth]{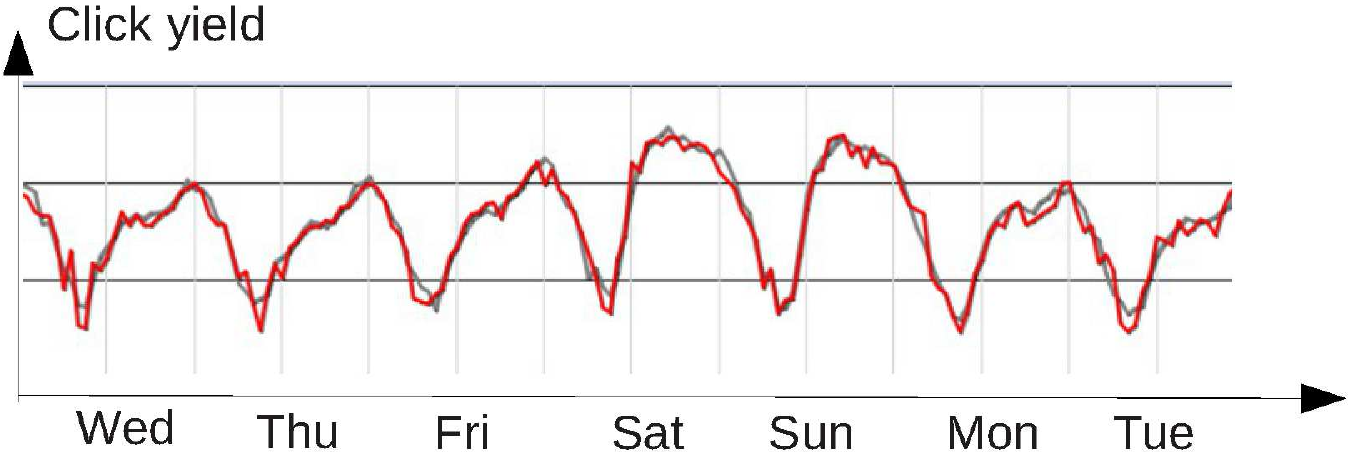}
\caption{\label{fig-dailycycle} 
The two plots show the hourly click yield for two variants of the ad
placement engine. The daily variations dwarf the
differences between the two treatments.}
\end{figure}


\subsection{Invariant Predictors}
\label{s-invariant-predictors}

In order to evaluate which of two interventions is most likely to
improve the system, the designer of a learning system often seeks to
estimate a \emph{counterfactual difference}, that is, the difference
$Y^+-Y^*$ of the expectations of a same quantity $\ell(\omega)$ under
two different counterfactual distributions~$\P^+\inPar{\omega}$
and~$\Qa{\omega}$.  These expectations are often affected by variables
whose value is left unchanged by the interventions under
consideration. For instance, seasonal effects can have very large
effects on the number of ad clicks (figure~\ref{fig-dailycycle}) but
affect $Y^+$ and $Y^*$ in similar ways.

Substantially better confidence intervals on the difference $Y^+-Y^*$
can be obtained using an \emph{invariant predictor}, that is,
a predictor function that depends only on invariant
variables~$\upsilon$ such as the time of the day. Since the invariant
predictor $\zeta(\upsilon)$ is not affected by the interventions
under consideration,
\begin{equation}
\label{eq-invariant-zeta}
   \int_\omega \zeta(\upsilon)\,\Qa{\omega}
   = \int_\omega \zeta(\upsilon)\,\P^+\inPar{\omega}\,.
\end{equation}
Therefore
\begin{eqnarray*}
Y^+-Y^* &=& \int_\omega \zeta(\upsilon)\,\P^+\inPar{\omega} 
            + \int_\omega \inPar{\ell(\omega)-\zeta(\upsilon)}\,\P^+\inPar{\omega} \\
        & & \quad - \int_\omega \zeta(\upsilon)\,\Qa{\omega}
            - \int_\omega \inPar{\ell(\omega)-\zeta(\upsilon)}\,\Qa{\omega} \\
       &\approx& \frac{1}{n} \sum_{i=1}^{n}~
             \big( \ell(\omega_i) - \zeta(\upsilon_i) \big) \:
             \frac{ P^+(\omega_i)-P^*(\omega_i) } {P(\omega_i)} \,.
\end{eqnarray*}
This direct estimate of the counterfactual difference~$Y^+-Y^*$
benefits from the same variance reduction effect
as~\eqref{eq-doubly-robust} without need to estimate the
expectations~\eqref{eq-invariant-zeta}.  Appendix~\ref{a-differences}
provide details on the computation of confidence intervals for
estimators of the counterfactual differences.
Appendix~\ref{a-derivatives} shows how the same approach can be used
to compute \emph{counterfactual derivatives} that describe the
response of the system to very small interventions.


\section{Learning}
\label{s-learning}

The previous sections deal with the identification and the measurement of
interpretable signals that can justify the actions of human decision makers.
These same signals can also justify the actions of machine learning
algorithms. This section explains why optimizing a counterfactual estimate is
a sound learning procedure.


\subsection{A Learning Principle}
\label{s-single-design-principle}

\begin{figure}
\center
\includegraphics[width=.67\linewidth]{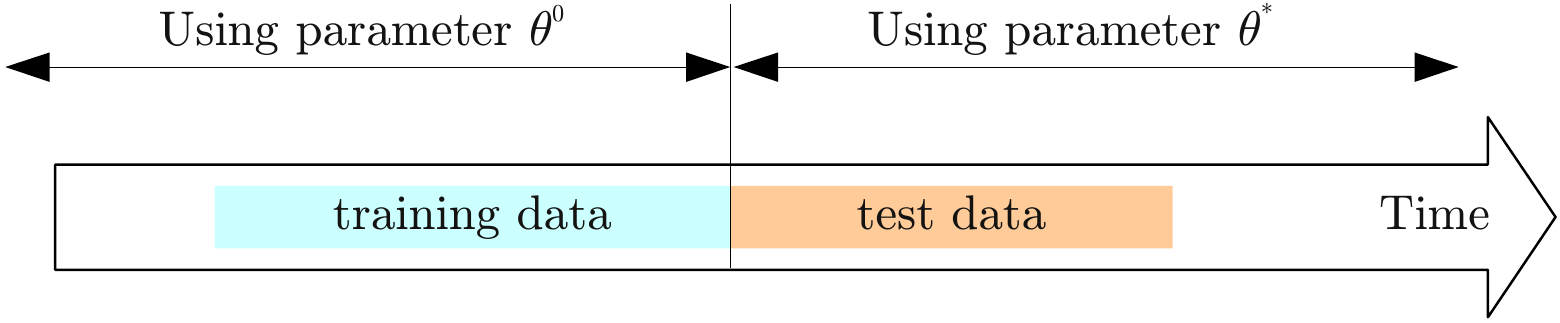}
\caption{Single design -- A preferred parameter value~$\thetastar$
  is determined using randomized data collected in the past.
  Test data is collected after loading~$\thetastar$ into the system.}
\label{fig-singledesign}
\end{figure}

We consider in this section interventions that depend on a
parameter $\theta$.  For instance, we might want to know what the
performance of the ad placement engine would have been if we had used
different values for the parameter $\theta$ of the click scoring model. 
Let~$\P^\theta\inPar{\omega}$ denote the
counterfactual Markov factorization associated with this
intervention. Let~$Y^\theta$ be the counterfactual expectation of
$\ell(\omega)$ under distribution
$\P^\theta$. Figure~\ref{fig-singledesign} illustrates our simple
learning setup.  Training data is collected from a single experiment
associated with an initial parameter value $\theta^0$ chosen using
prior knowledge acquired in an unspecified manner. A preferred
parameter value~$\thetastar$ is then determined using the training
data and loaded into the system.  The goal is of course to observe a
good performance on data collected during a test period that takes
place after the switching point.

The isolation assumption introduced in section~\ref{s-isolation} states that
the exogenous variables are drawn from an unknown but fixed joint probability
distribution. This distribution induces a joint distribution
$\Pa{\omega}$ on all the variables $\omega$ appearing in the structural
equation model associated with the parameter $\theta$. Therefore, if the
\emph{isolation assumption} \emph{remains valid during the test period}, the
test data follows the same distribution $\P^\thetastar(\omega)$ that would
have been observed during the training data collection period if the system
had been using parameter~$\thetastar$ all along.

We can therefore formulate this problem as the optimization of
the expectation~$Y^\theta$ of the reward $\ell(\omega)$ with 
respect to the distribution~$\P^\theta(\omega)$
\begin{equation}
\label{eq-goal-of-learning} 
  \max_\theta ~ Y^\theta 
     ~=~ \int_\omega \: \ell(\omega) \: \P^\theta(\omega)
\end{equation}
on the basis of a finite set of training examples $\omega_1,\dots,\omega_n$
sampled from $\Pa{\omega}$. 

However, it would be unwise to maximize the estimates obtained using
approximation~\eqref{eq-general-reweighting} because they could reach a
maximum for a value of~$\theta$ that is poorly explored by the actual
distribution. As explained in section~\ref{s-interpreting-confidence}, the gap
between the upper and lower bound of inequality \eqref{eq-bias-bound} reveals
the uncertainty associated with insufficient exploration.  Maximizing an
empirical estimate~$\estY^\theta$ of the lower bound~$\capY^\theta$ ensures
that the optimization algorithm finds a trustworthy answer

\begin{equation}
\label{eq-singledesign-principle}
    \thetastar ~=~ \argmax_\theta \estY^\theta~.
\end{equation}

We shall now discuss the statistical basis of this learning
principle.\footnote{The idea of maximizing the lower bound may surprise
readers familiar with the UCB algorithm for multi-armed
bandits \citep{auer-2002}. UCB performs exploration by maximizing the upper
confidence interval bound and updating the confidence intervals online.
Exploration in our setup results from the active system randomization during
the offline data collection. See also section~\ref{s-sequential-design}.}


\subsection{Uniform Confidence Intervals}
\label{s-uniform}

As discussed in section~\ref{s-confidence}, 
inequality~\eqref{eq-bias-bound},
\[
   \capY^\theta ~\le~ Y^\theta  \:\le~ 
           \capY^\theta + M (1-\capW^\theta) ~,
\]
where 
\begin{eqnarray*}
   \capY^\theta ~=~ \int_\omega \ell(\omega) \: \capw(\omega) \: \Pa{\omega} 
  &\approx& 
   \estY^\theta ~=~ \frac{1}{n} \sum_{i=1}^n
   \ell(\omega_i)\:\capw(\omega_i)~,
  \\[-.3ex]
   \capW^\theta ~=~ \int_\omega \capw(\omega) \: \Pa{\omega}
  &\approx&
   \estW^\theta ~=~\frac{1}{n} \sum_{i=1}^n \capw(\omega_i) ~,
\end{eqnarray*}
leads to confidence intervals
\eqref{eq-conf-dual} of the form
\begin{equation}
\label{eq-non-uniform-conf-interval}
   \forall\delta>0,~\forall\theta \quad 
   \bbbP\inBrace{ ~ \estY^\theta - \epsilon_R 
           ~\le~ Y^\theta \:\le~
       \estY^\theta + M(1 - \estW^\theta + \xi_R) 
          + \epsilon_R ~ } 
    \ge 1-\delta \,.
\end{equation}
Both~$\epsilon_R$ and~$\xi_R$ converge to zero in inverse proportion to the
square root of the sample size~$n$.  They also increase at most linearly in
$\log\delta$ and depend on both the capping bound~$R$ and the
parameter~$\theta$ through the empirical variances (see
appendix~\ref{a-confidence}.)

Such confidence intervals are insufficient to provide guarantees for a
parameter value~$\thetastar$ that depends on the sample.  In fact, the
optimization~\eqref{eq-singledesign-principle} procedure is likely to select
values of $\theta$ for which the inequality is violated. We therefore seek
uniform confidence intervals \citep{vapnik-chervonenkis-1968}, 
simultaneously valid for all values of~$\theta$.
\begin{itemize}
\item
  When the parameter~$\theta$ is chosen from a finite
  set~$\mathcal{F}$, applying the union bound to the ordinary 
  intervals~\eqref{eq-non-uniform-conf-interval} immediately
  gives the uniform confidence interval\::
  \[
     \bbbP\inBrace{\:\forall\theta\in\mathcal{F},~
       \estY^\theta\!-\!\epsilon_R 
           \le Y^\theta \le
       \estY^\theta\!+\! M(1\!-\!\estW^\theta\!+\!\xi_R) 
          \!+\!\epsilon_R \: } 
      \ge 1\!-\!|\mathcal{F}|\,\delta \,.
  \]
\item
  Following the pioneering work of~\citeauthor{vapnik-chervonenkis-1968},
  a~broad choice of mathematical tools have been developed to construct
  uniform confidence intervals when the set~$\mathcal{F}$ is infinite.
  For instance, appendix~\ref{a-uniform-bernstein} leverages
  uniform empirical Bernstein bounds~\citep{maurer-pontil-2009} 
  and obtains the uniform confidence interval
  \begin{equation}
   \label{eq-uniform-conf-interval}
     \bbbP\inBrace{\:\forall\theta\in\mathcal{F},~
       \estY^\theta\!-\!\epsilon_R 
           \le Y^\theta \le
       \estY^\theta\!+\! M(1\!-\!\estW^\theta\!+\!\xi_R) 
          \!+\!\epsilon_R \: } 
      \ge 1\!-\!\mathcal{M}(n)\,\delta \,,
  \end{equation}
  where the growth function~$\mathcal{M}(n)$ measures the capacity 
  of the family of functions
  \begin{equation}
   \label{eq-growth-function-family}  
     \inBrace{~ f_\theta: \omega \mapsto \ell(\omega)\capw(\omega)
              ~,~~ g_\theta: \omega \mapsto \capw(\omega)
              ~,~~\forall \theta\in\mathcal{F} ~} ~.
  \end{equation}
  Many practical choices of $\Qa{\omega}$ lead to 
  functions~$\mathcal{M}(n)$ that grow polynomially with the sample size.
  Because both $\epsilon_R$ and $\xi_R$ are 
  $\mathcal{O}(n^{-{1}/{2}}\log\delta)$, they converge to zero
  with the sample size when one maintains the confidence 
  level $1\!-\!\mathcal{M}(n)\,\delta$ equal to a predefined constant.
\end{itemize}

\begin{figure}
\center
\includegraphics[width=.6\linewidth]{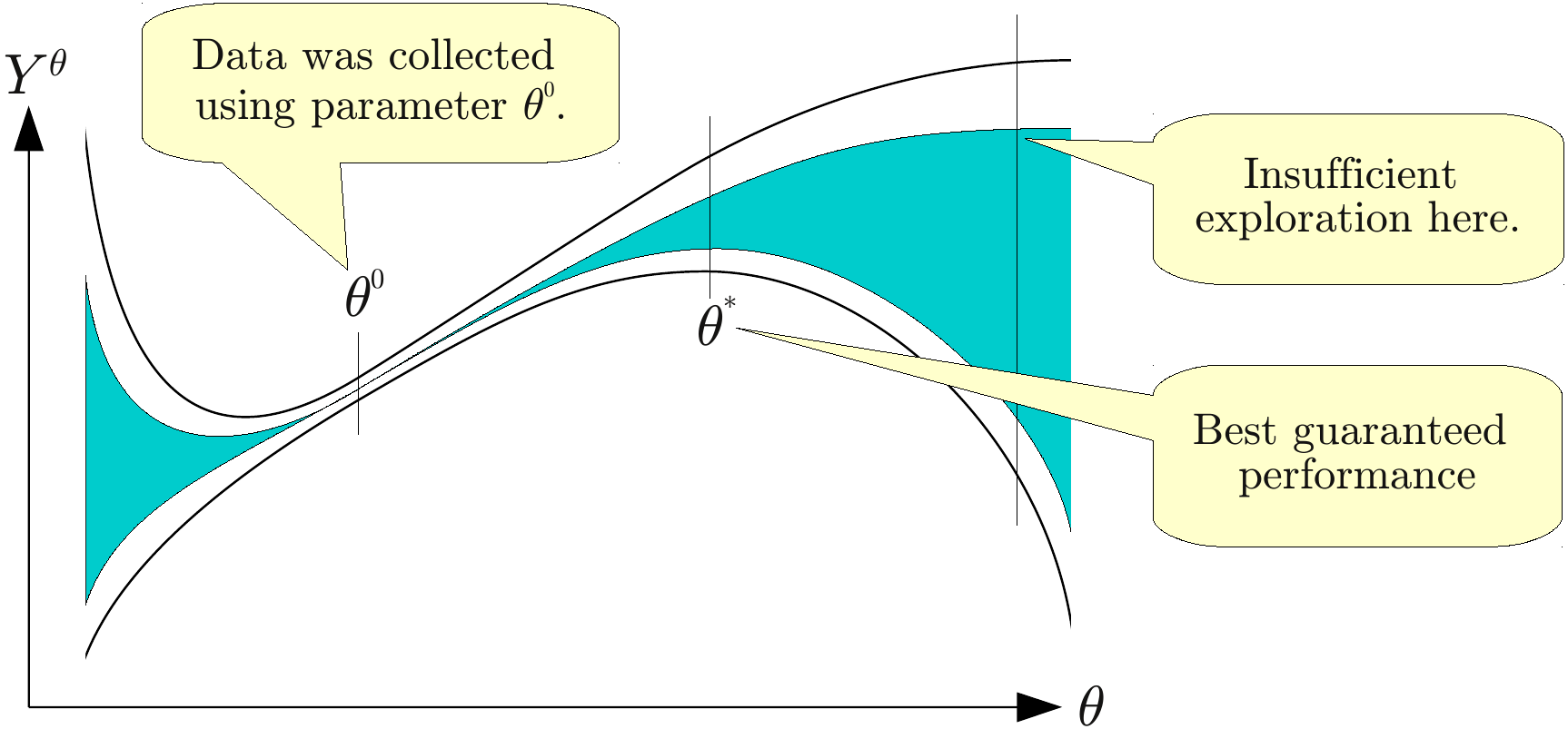}
\caption{\label{fig-learningbound}\relax The uniform inner confidence
  interval reveals where the best guaranteed~$Y^\theta$ is reached
  and where additional exploration is needed.}
\end{figure}

\noindent
The intepretation of the inner and outer confidence intervals
(section~\ref{s-interpreting-confidence}) also applies to the
uniform confidence interval~\eqref{eq-uniform-conf-interval}.  When
the sample size is sufficiently large and the capping bound~$R$ chosen
appropriately, the inner confidence interval reflects the upper and
lower bound of inequality~\eqref{eq-bias-bound}. 

The uniform confidence interval therefore ensures that ~$Y^\thetastar$ is close
to the maximum of the lower bound of inequality~\eqref{eq-bias-bound} which
essentially represents the best performance that can be guaranteed using
training data sampled from~$\Pa{\omega}$. Meanwhile, the upper bound of this
same inequality reveals which values of~$\theta$ could potentially offer
better performance but have been insufficiently probed by the sampling
distribution (figure~\ref{fig-learningbound}.)


\subsection{Tuning Ad Placement Auctions}
\label{s-exp-alphamlr}

We now present an application of this learning principle to the optimization
of auction tuning parameters in the ad placement engine.  Despite increasingly
challenging engineering difficulties, comparable optimization procedures can
obviously be applied to larger numbers of tunable parameters.

\begin{figure}[t]
\center
\includegraphics[width=.5\linewidth]{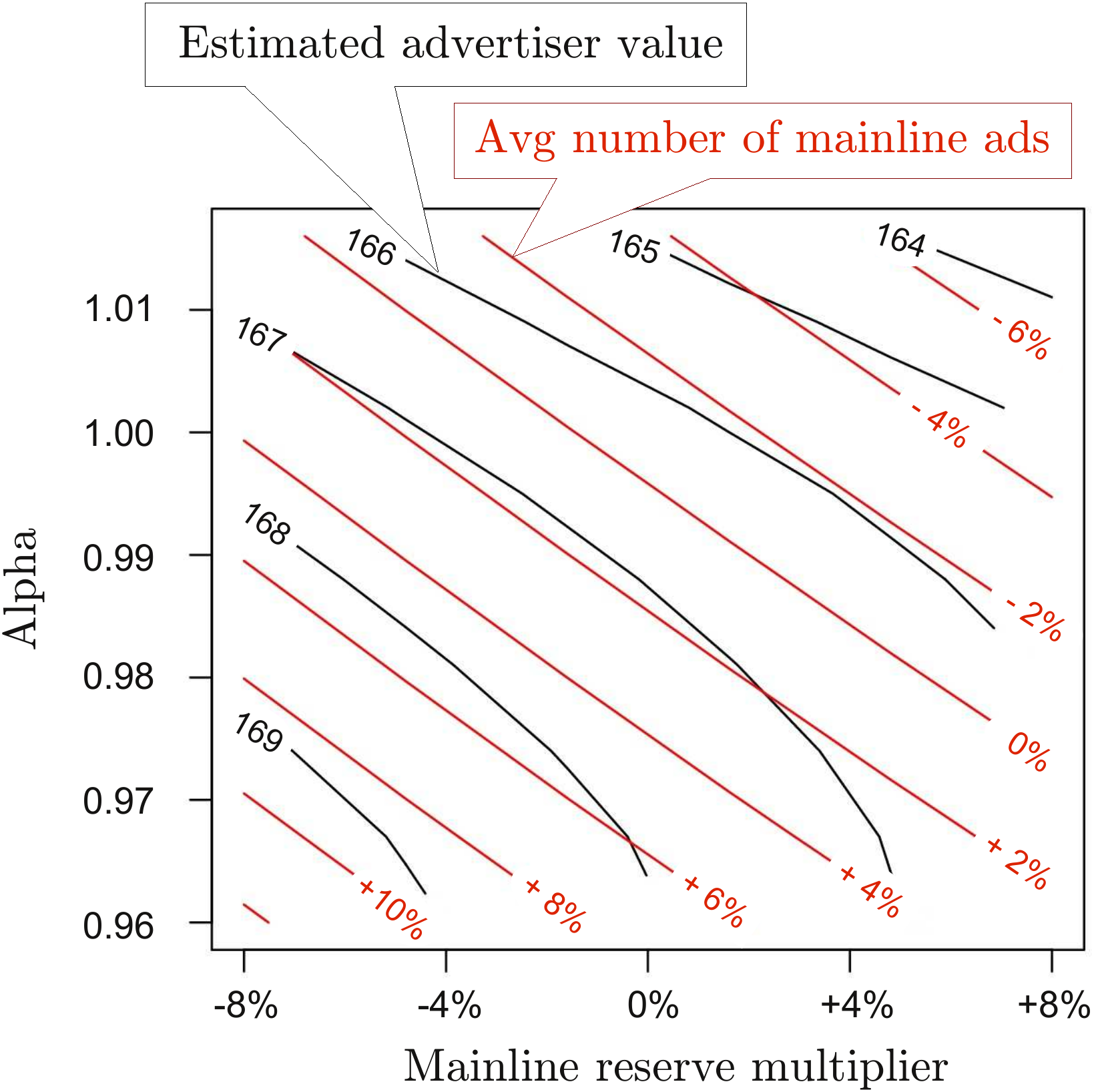}
\caption{Level curves associated with the average number of mainline ads per
  page (red curves, from~$-6\%$ to~$+10\%$) and the average estimated
  advertisement value generated per page (black curves, arbitrary units
  ranging from~$164$ to~$169$) that would have been observed for a certain
  query cluster if we had changed the mainline reserves by the multiplicative
  factor shown on the horizontal axis, and if we had applied a squashing
  exponent~$\alpha$ shown on the vertical axis to the estimated click
  probabilities~$q_{i,p}(x)$.}
\label{fig-exp-alphamlr}
\end{figure}

\citet{lahaie-mcafee-2011} propose to account for the uncertainty
of the click probability estimation by introducing a squashing
exponent~$\alpha$ to control the impact of the estimated probabilities
on the rank scores.  Using the notations introduced in
section~\ref{s-adplacement}, and assuming that the estimated
probability of a click on ad~$i$ placed at position $p$ after
query~$x$ has the form \mbox{$q_{ip}(x)=\gamma_p\:\beta_i(x)$} 
(see appendix~\ref{a-greedyadplacement}), they redefine the
rank-score~$r_{ip}(x)$ as:
\[
   r_{ip}(x) = \gamma_p\:b_i\:\beta_i(x)^\alpha~.
\]
Using a squashing exponent~$\alpha<1$ reduces the contribution of the
estimated probabilities and increases the reliance on the bids~$b_i$
placed by the advertisers.  

Because the squashing exponent changes the rank-score scale, it is
necessary to simultaneously adjust the reserves in order to display
comparable number of ads. In order to estimate the counterfactual
performance of the system under interventions affecting both the
squashing exponent and the mainline reserves, we have collected data
using a random squashing exponent following a normal distribution, and
a mainline reserve multiplier following a log-normal distribution as
described in section~\ref{s-exp-mlr}. Samples describing 12 million
search result pages were collected during four consecutive weeks.

Following \citet{charles-2012b}, we consider separate squashing
coefficients~$\alpha_k$ and mainline reserve multipliers~$\rho_k$ per query
cluster~$k\in\{1..K\/\}$, and, in order to avoid negative user or advertiser
reactions, we seek the auction tuning parameters~$\alpha_k$ and~$\rho_k$ that
maximize an estimate of the advertisement value\footnote{\relax The value of
  an ad click from the point of view of the advertiser. The advertiser payment
  then splits the advertisement value between the publisher and the
  advertiser.}  subject to a global constraint on the average number of ads
displayed in the mainline. Because maximizing the advertisement value instead
of the publisher revenue amounts to maximizing the size of the advertisement
pie instead of the publisher slice of the pie, this criterion is less likely
to simply raise the prices without improving the ads. Meanwhile the constraint
ensures that users are not exposed to excessive numbers of mainline ads.

We then use the collected data to estimate bounds on the
counterfactual expectations of the advertiser value and the
counterfactual expectation of the number of mainline ads per
page. Figure~\ref{fig-exp-alphamlr} shows the corresponding level
curves for a particular query cluster. We can then run a simple
optimization algorithm and determine the optimal auction tuning
parameters for each cluster subject to the global mainline footprint
constraint. Appendix~\ref{a-derivatives} describes how to estimate
off-policy counterfactual derivatives that greatly help the numerical
optimization.

The obvious alternative \citep[see][]{charles-2012b} consists of
replaying the auctions with different parameters and simulating the user
using a click probability model. However, it may be unwise to rely on
a click probability model to estimate the best value of a squashing
coefficient that is expected to compensate for the uncertainty of the
click prediction model itself. The counterfactual approach described
here avoids the problem because it does not rely on a click prediction
model to simulate users. Instead it estimates the counterfactual
peformance of the system using the actual behavior of the users
collected under moderate randomization.


\subsection{Sequential Design}
\label{s-sequential-design}

\begin{figure}
\center
\includegraphics[width=.67\linewidth]{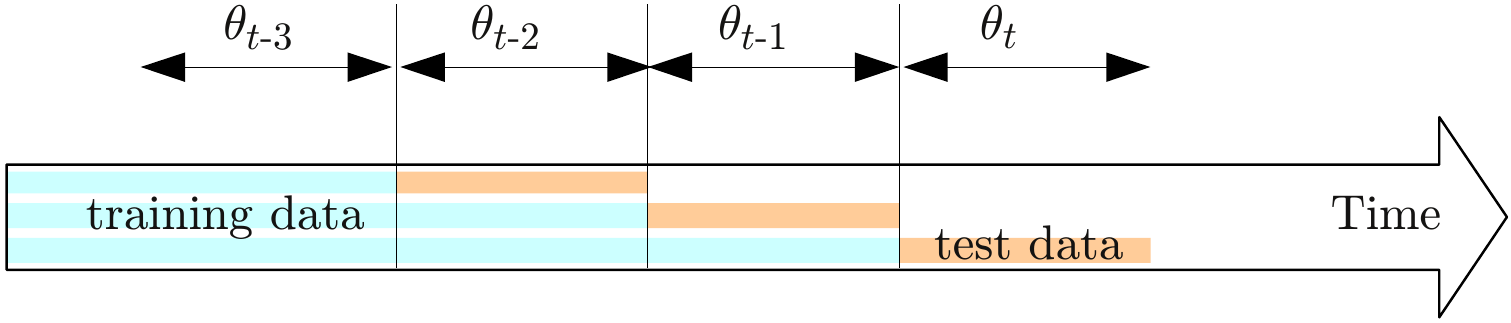}
\caption{Sequential design -- The parameter $\theta_t$ of each data
  collection experiment is determined using data collected during
  the previous experiments.}
\label{fig-fullseqdesign}
\end{figure}

Confidence intervals computed after a first randomized data collection
experiment might not offer sufficient accuracy to choose a final value of
the parameter~$\theta$. It is generally unwise to simply collect additional
samples using the same experimental setup because the current data already
reveals information (figure~\ref{fig-learningbound}) that can be used to
design a better data collection experiment. Therefore, it seems natural to
extend the learning principle discussed in
section~\ref{s-single-design-principle} to a sequence of data collection
experiments. The parameter $\theta_t$ characterizing the $t$-th experiment is
then determined using samples collected during the previous experiments
(figure~\ref{fig-fullseqdesign}).

Although it is relatively easy to construct convergent sequential
design algorithms, reaching the \emph{optimal} learning performance is
notoriously difficult~\citep{wald-1945} because the selection
of parameter~$\theta_t$ involves a trade-off between exploitation,
that is, the maximization of the immediate reward $Y^{\theta_t}$, and
exploration, that is, the collection of samples potentially leading to
better $Y^\theta$ in the more distant future.

The optimal exploration exploitation trade-off for multi-armed bandits
is well understood~\citep{gittins-1989,auer-2002,audibert-2007}
because an essential property of multi-armed bandits makes the
analysis much simpler: the outcome observed after performing a
particular action brings no information about the value of other
actions. Such an assumption is both unrealistic and pessimistic. For
instance, the outcome observed after displaying a certain ad in
response to a certain query brings very useful information about the
value of displaying similar ads on similar queries.

Refined contextual bandit approaches~\citep{slivkins-2011}
account for similarities in the context and action spaces but do not
take advantage of all the additional opportunities expressed by
structural equation models. For instance, in the contextual bandit
formulation of the ad placement problem outlined in
section~\ref{s-bandits-and-rl}, actions are pairs~$(s,c)$ describing
the ad slate~$s$ and the corresponding click prices~$c$, policies
select actions by combining individual ad scores in very specific
ways, and actions determine the rewards through very specific
mechanisms. 

Meanwhile, despite their suboptimal asymptotic properties, heuristic
exploration strategies perform surprisingly well during the time span
in which the problem can be considered stationary. Even in the simple
case of multi-armed bandits, excellent empirical results have been
obtained using Thompson sampling~\citep{chapelle-li-2012} or fixed
strategies~\citep{vermorel-mohri-2005,kuleshov-precup-2010}.
Leveraging the problem structure seems more important in practice than
perfecting an otherwise sound exploration strategy.

Therefore, in the absence of sufficient theoretical guidance, it is
both expedient and practical to maximizing~$\estY^{\theta}$ at each
round, as described in section~\ref{s-single-design-principle},
subject to additional ad-hoc constraints ensuring a minimum level of
exploration.

\section{Equilibrium Analysis}
\label{s-equilibrium}

All the methods discussed in this contribution rely on the isolation assumption
presented in section~\ref{s-isolation}. This assumption lets us interpret the
samples as repeated independent trials that follow the pattern defined by the
structural equation model and are amenable to statistical analysis.

The isolation assumption is in fact a component of the counterfactual
conditions under investigation. For instance, in
section~\ref{s-exp-mlr}, we model single auctions
(figure~\ref{fig-causalgraph}) in order to empirically determine how
the ad placement system would have performed if we had changed the
mainline reserves \emph{without incurring a reaction from the users or
the advertisers}.

\smallskip
Since the future publisher revenues depend on the continued satisfaction of
users and advertisers, lifting this restriction is highly desirable.
\begin{itemize}
\item
  We can in principle work with larger structural equation models. For
  instance, figure~\ref{fig-manysem} suggests to thread single auction models
  with additional causal links representing the impact of the displayed ads on
  the future user goodwill. However, there are practical limits on the number
  of trials we can consider at once.  For instance, it is relatively easy to
  simultaneously model all the auctions associated with the web pages served
  to the same user during a thirty minute web session.  On the other hand, it is
  practially impossible to consider several weeks worth of auctions in order
  to model their accumulated effect on the continued satisfaction of users and
  advertisers.
\item
  We can sometimes use problem-specific knowledge to construct alternate
  performance metrics that anticipate the future effects of the feedback
  loops. For instance, in section~\ref{s-exp-alphamlr}, we optimize the
  advertisement value instead of the publisher revenue. Since this alternative
  criterion takes the advertiser interests into account, it can be viewed 
  as a heuristic proxy for the future revenues of the publisher.
\end{itemize}

\noindent

This section proposes an alternative way to account for such feedback loops
using the \emph{quasistatic equilibrium} method familiar to physicists:
we assume that the publisher changes the parameter~$\theta$ so slowly that the
system remains at equilibrium at all times. Using data collected while the
system was at equilibrium, we describe empirical methods to determine how an
infinitesimal intervention $\d\theta$ on the model parameters would have
displaced the equilibrium:
\begin{quotation}
\hyphenpenalty2000
\noindent\llap{``}\emph{\relax How would the system have performed during the
  data collection period if a small change~$\d\theta$ had been applied to the
  model parameter~$\theta$ and the equilibrium had been reached before the
  data collection period.}''
\end{quotation}
A learning algorithm can then update~$\theta$ to improve selected performance metrics.

\subsection{Rational Advertisers}
\label{s-rational-advertisers}

The ad placement system is an example of game where each actor
furthers his or her interests by controlling some aspects of the
system: the publisher controls the placement engine parameters, the
advertisers control the bids, and the users control the clicks.

As an example of the general quasi-static approach, this section
focuses on the reaction of \emph{rational advertisers} to small
changes of the scoring functions driving the ad placement system.
Rational advertisers always select bids that maximize their economic
interests.  Although there are more realistic ways to model
advertisers, this exercise is interesting because the auction theory
approaches also rely on the rational advertiser assumption (see
section~\ref{s-adplacement}). This analysis seamlessly integrates the
auction theory and machine learning perspectives.

\begin{figure}[t]
\center
\includegraphics[width=.75\linewidth]{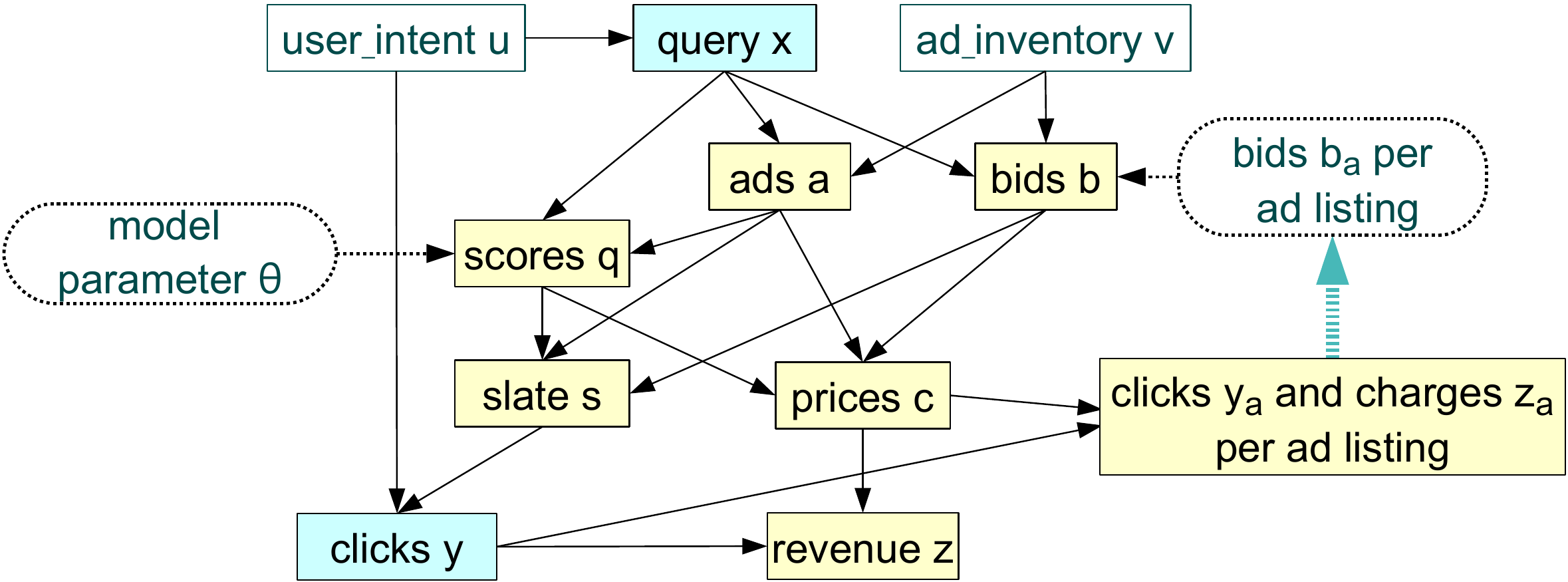}
\caption{\label{fig-advertiserloop} Advertisers select the bid amounts~$b_a$
  on the basis of the past number of clicks~$y_a$ and the past prices~$z_a$
  observed for the corresponding ads.}
\end{figure}

\begin{figure}
\center
\includegraphics[width=.52\linewidth]{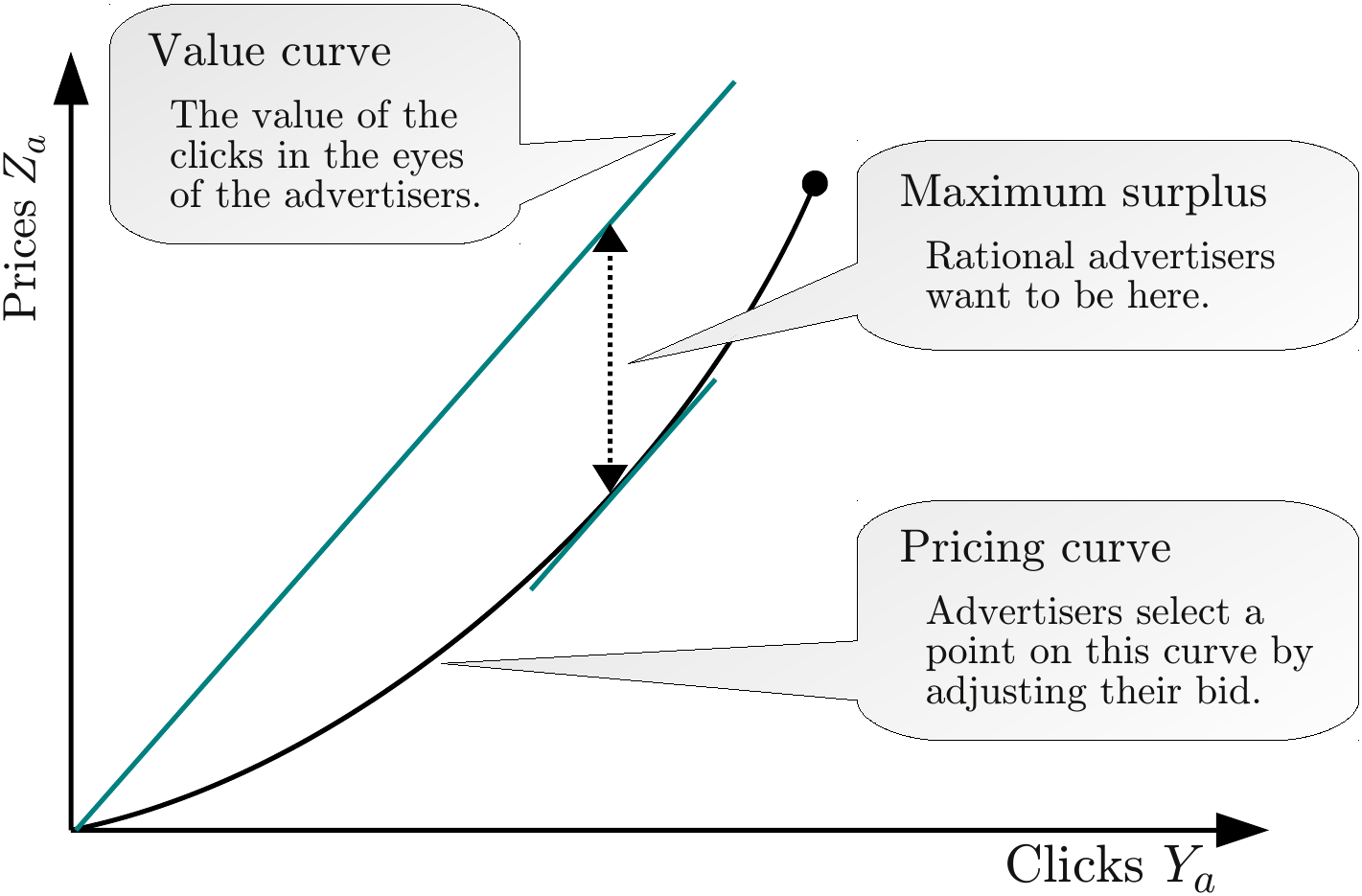}
\caption{\label{fig-pricing} Advertisers control the expected number of
  clicks~$Y_a$ and expected prices~$Z_a$ by adjusting their
  bids~$b_a$. Rational advertisers select bids that maximize the difference
  between the value they see in the clicks and the price they pay.}
\end{figure}

\smallskip 

As illustrated in figure~\ref{fig-advertiserloop}, we treat the bid
vector~$\b=(b_1\dots b_A)\in[0,b_{\max}]^A$ as the parameter of the
conditional distribution~$\P^\b(b|x,v)$ of the bids associated with
the eligible ads.\footnote{Quantities measured when a feedback causal
system reaches equilibrium often display conditional independence
patterns that cannot be represented with directed acyclic
graphs~\citep{lauritzen-richardson-2002,dash-2003}. Treating the
feedback loop as parameters instead of variables works around this
difficuly in a manner that appears sufficient to perform the
quasi-static analysis.}  The variables~$y_a$ in the structural
equation model represents the number of clicks received by ads
associated with bid~$b_a$. The variables~$z_a$ represents the amount
charged for these clicks to the corresponding advertiser. The
advertisers select their bids~$b_a$ according to their anticipated
impact on the number of resulting clicks~$y_a$ and on their
cost~$z_a$.

Following the pattern of the perfect information assumption (see
section~\ref{s-adplacement}), we assume that the advertisers 
eventually acquire full knowledge of the expectations
\[
  Y_a(\theta,\b) = \int_\omega y_a~\P^{\theta\!,\b}(\omega)
  \quad\text{and}\quad
  Z_a(\theta,\b) = \int_\omega z_a~\P^{\theta\!,\b}(\omega) ~.
\]
Let $V_a$ denote the value of a click for the corresponding advertiser.  
Rational advertiser seek to maximize the difference between the value
they see in the clicks and the price they pay to the publisher,
as illustrated in figure~\ref{fig-pricing}.  
This is expressed by the utility functions
\begin{equation}
\label{eq-surplus}
     U_a^\theta(\b) ~=~ V_a \: Y_a(\theta,\b) - Z_a(\theta,\b) ~.
\end{equation}

Following~\citet{athey-nekipelov-2010}, we argue that the injection of
smooth random noise into the auction mechanism changes the discrete
problem into a continous problem amenable to standard differential
methods. Mild regularity assumption on the densities
probability~$\P^\b(b|x,v)$ and~$\P^\theta(q|x,a)$ are in fact
sufficient to ensure that the expectations~$Y_a(\theta,\b)$
and~$Z_a(\theta,\b)$ are continuously differentiable functions 
of the distribution parameters~$\b$ and~$\theta$. 
Further assuming that utility functions $U_a^\theta(\b)$ are
diagonally quasiconcave, \citeauthor{athey-nekipelov-2010}
establish the existence of a unique Nash equilibrium
\begin{equation}
\label{eq-nash-equilibrium}
    \forall a\quad b_a \in \ArgMax_b~ 
       U^\theta_a( b_1,\dots,b_{a-1},b,b_{a+1},\dots,b_A) 
\end{equation}
characterized by its first order Karush-Kuhn-Tucker conditions
\begin{equation}
\label{eq-rational-equilibrium}
  \forall a\qquad V_a \, \pd{Y_a}{b_a} - \pd{Z_a}{b_a} ~~
  \left\{\begin{array}{ll}
     \le0 & \text{if $b_a=0$,} \\
     \ge0 & \text{if $b_a=b_{\max}$,} \\
     =0 & \text{if $0<b_a<b_{\max}$.}
  \end{array}\right.
\end{equation}

We use the first order equilibrium
conditions~\eqref{eq-rational-equilibrium} for two related purposes.
Section~\ref{sec-equilibrium-values} explains how to complete the
advertiser model by estimating the values~$V_a$.
Section~\ref{sec-equilibrium-linear-response} estimates how the
equilibrium bids and the system performance metrics respond to a small
change $\d\theta$ of the model parameters.

Interestingly, this approach remains sensible when key assumptions of
the equilibrium model are violated. The perfect information assumption
is unlikely to hold in practice. The quasi-concavity of the utility
functions is merely plausible. However, after observing the operation
of the stationary ad placement system for a sufficiently long time, it
is reasonable to assume that the most active advertisers have tried
small bid variations and have chosen locally optimal ones. Less active
advertisers may leave their bids unchanged for longer time periods, but
can also update them brutally if they experience a significant change in
return on investment.  Therefore it makes sense to use data collected
when the system is stationary to estimate advertiser values~$V_a$ that
are consistent with the first order equilibrium conditions. 
We then hope to maintain the conditions that each advertisers 
had found sufficiently attractive, by first estimating how a 
small change~$\d\theta$ displaces this posited local equilibrium,
then by using performance metrics that take this displacement
into account.

\subsection{Estimating advertiser values}
\label{sec-equilibrium-values}

We first need to estimate the partial derivatives appearing in the
equilibrium condition~\eqref{eq-rational-equilibrium}. These
derivatives measure how the expectations $Y_a$ and $Z_a$ would have
been changed if each advertiser had placed a slighly different
bid~$b_a$. Such quantities can be estimated by randomizing the bids
and computing on-policy counterfactual derivatives as explained in
appendix~\ref{a-derivatives}. Confidence intervals can be derived with
the usual tools.

Unfortunately, the publisher is not allowed to directly randomize the
bids because the advertisers expect to pay prices computed using the
bid they have specified and not the potentially higher bids resulting
from the randomization. However, the publisher has full control on the
estimated click probabilities~$q_{i,p}(x)$. Since the
rank-scores~$r_{i,p}(x)$ are the products of the bids and the
estimated click probabilities (see section~\ref{s-adplacement}), a
random multiplier applied to the bids can also be interpreted as a
random multiplier applied to the estimated click probabilities. Under
these two interpretations, the same ads are shown to the users, but
different click prices are charged to the advertisers. Therefore, the
publisher can simultaneously charge prices computed as if the
multiplier had been applied to the estimated click probabilities, and
collect data as if the multiplier had been applied to the bid.  This
data can then be used to estimate the derivatives.

Solving the first order equilibrium equations then yields estimated
advertiser values $V_a$ that are consistent with the observed
data.\footnote{This approach is of course related to the value
estimation method proposed by \citet{athey-nekipelov-2010} but
strictly relies on the explicit randomization of the scores. In
contrast, practical considerations force~\citeauthor{athey-nekipelov-2010} 
to rely on the apparent noise and hope that the noise model accounts 
for all potential confounding factors.}
\[
    V_a \approx  \pd{Y_a}{b_a} \Big/ \pd{Z_a}{b_a}
\]
There are however a couple caveats:
\begin{itemize}
\item
   The advertiser bid $b_a$ may be too small to cause ads to be
   displayed.  In the absence of data, we have no means to estimate a
   click value for these advertisers.
\item
   Many ads are not displayed often enough to obtain accurate estimates
   of the partial derivatives $\pd{Y_a}{b_a}$ and $\pd{Z_a}{b_a}$.
   This can be partially remediated by smartly aggregating the data of
   advertisers deemed similar.
\item
   Some advertisers attempt to capture all the available ad
   opportunities by placing extremely high bids and hoping to pay
   reasonable prices thanks to the generalized second price rule.
   Both partial derivatives $\pd{Y_a}{b_a}$ and $\pd{Z_a}{b_a}$ are
   equal to zero in such cases. Therefore we cannot recover $V_a$ by
   solving the equilibrium
   equation~\eqref{eq-rational-equilibrium}. It is however possible to
   collect useful data by selecting for these advertisers a maximum
   bid $b_{\max}$ that prevents them from monopolizing the eligible ad
   opportunities. Since the equilibrium condition is an inequality
   when $b_a=b_{\max}$, we can only determine a lower bound of the
   values $V_a$ for these advertisers.
\end{itemize}
These caveats in fact underline the limitations of the advertiser
modelling assumptions. When their ads are not displayed often enough,
advertisers have no more chance to acquire a full knowledge of the
expectations $Y_a$ and $Z_a$ than the publisher has a chance to
determine their value. Similarly, advertisers that place extremely
high bids are probably underestimating the risk to occasionally
experience a very high click price. A more realistic model of the
advertiser information acquisition is required to adequately handle
these cases.

\subsection{Estimating the equilibrium response}
\label{sec-equilibrium-linear-response}

Let $\calA$ be the set of the \emph{active advertisers}, that is, the
advertisers whose value can be estimated (or lower bounded) with sufficient
accuracy. Assuming that the other advertisers leave their bids unchanged, we
can estimate how the active advertisers adjust their bids in response to an
infinitesimal change $\d\theta$ of the scoring model parameters. This is
achieved by differentiating the equilibrium
equations \eqref{eq-rational-equilibrium}:
\begin{equation}
  \def\ap{{a^\prime}}
  \label{eq-rational-equilibrium-system}
  \forall \ap\in\calA,\quad
   0 ~ = ~ 
    \inPar{V_\ap\,\pdd{Y_\ap}{b_\ap}{\theta}-\pdd{Z_\ap}{b_\ap}{\theta}}
    \:\d\theta + \sum_{a\in\calA}
    \inPar{V_\ap\,\pdd{Y_\ap}{b_\ap}{b_a}-\pdd{Z_\ap}{b_\ap}{b_a} }
    \:\d b_a ~.
\end{equation}
The partial second derivatives must be estimated as described in
appendix~\ref{a-derivatives}. Solving this linear system of
equations then yields an expression of the form
\[
    \d b_a ~=~ \Xi_a \, \d\theta \,.
\]
This expression can then be used to estimate how any counterfactual
expectation~$Y$ of interest changes when the publisher applies an
infinitesimal change~$\d\theta$ to the scoring parameter~$\theta$ and the
active advertisers $\calA$ rationally adjust their bids $b_a$ in response:
\begin{equation}
\label{eq-rational-advertiser-total-variation}
   \d Y = \inPar{\, \pd{Y}{\theta} \:+\: \sum_a \,\Xi_a\,\pd{Y}{b_a} \,} \, \d\theta~.
\end{equation}

Although this expression provides useful information, one should remain aware
of its limitations. Because we only can estimate the reaction of active
advertisers, expression~\eqref{eq-rational-advertiser-total-variation}
does not includes the potentially positive reactions of advertisers who did
not bid but could have. Because we only can estimate a lower bound of their
values, this expression does not model the potential reactions of advertisers
placing unrealistically high bids. Furthermore, one needs to be very cautious
when the system~\eqref{eq-rational-equilibrium-system} approaches
singularities. Singularities indicate that the rational advertiser
assumption is no longer sufficient to determine the reactions of certain
advertisers. This happens for instance when advertisers cannot find bids that
deliver a satisfactory return. The eventual behavior of such advertisers then
depends on factors not taken in consideration by our model.

To alleviate these issues, we could alter the auction mechanism in a
manner that forces advertisers to reveal more information, and we
could enforce policies ensuring that the 
system~\eqref{eq-rational-equilibrium-system} remains safely
nonsingular. We could also design experiments revealing the impact of
the fixed costs incurred by advertisers participating into new
auctions. Although additional work is needed to design such
refinements, the quasistatic equilibrium approach provides a generic
framework to take such aspects into account.

\subsection{Discussion}

The rational advertiser assumption is the cornerstone of seminal works
describing simplified variants of the ad placement problem using
auction theory \citep{varian-2007,edelman-2007}. More sophisticated
works account for more aspects of the ad placement problem, such as
the impact of click prediction learning
algorithms~\citep{lahaie-mcafee-2011}, the repeated nature of the ad
auctions~\citep{bergemann-said-2010}, or for the fact that advertisers
place bids valid for multiple
auctions~\citep{athey-nekipelov-2010}. Despite these advances, it
seems technically very challenging to use these methods and account
for all the effects that can be observed in practical ad placement
systems.

We believe that our counterfactual reasoning framework is best viewed
as a modular toolkit that lets us apply insights from auction theory
and machine learning to problems that are far more complex than those
studied in any single paper. For instance, the quasi-static
equilibrium analysis technique illustrated in this section extends
naturally to the analysis of multiple simultaneous causal feedback
loops involving additional players:
\begin{itemize}
\item
  The first step consists in designing ad-hoc experiments to identify
  the parameters that determine the equilibrium equation of each
  player.  In the case of the advertisers, we have shown how to use
  randomized scores to reveal the advertiser values. In the case of
  the user feedback, we must carefully design experiments that reveal
  how users respond to changes in the quality of the displayed ads.
\item
  Differentiating all the equilibrium equations yields a linear system
  of equations linking the variations of the parameter under our
  control, such as $d\theta$, and all the parameters under the control
  of the other players, such as the advertiser bids, or the user
  willingness to visit the site and click on ads. Solving this system
  and writing the total derivative of the performance measure gives
  the answer to our question.
\end{itemize}

Although this programme has not yet been fully realized, the existence
of a principled framework to handle such complex interactions is
remarkable. Furthermore, thanks to the flexibility of the causal
inference frameworks, these techniques can be infinitely adapted to
various modeling assumptions and various system complexities.


\section{Conclusion}
 
Using the ad placement example, this work demonstrates the central role of
causal inference \citep{pearl-2000,spirtes-1993} for the design of learning
systems interacting with their environment. Thanks to importance sampling
techniques, data collected during randomized experiments gives precious cues
to assist the designer of such learning systems and useful signals to drive
learning algorithms.

Two recurrent themes structure this work. First, we maintain a sharp
distinction between the learning algorithms and the extraction of the
signals that drive them. Since real world learning systems often
involve a mixture of human decision and automated processes, it makes
sense to separate the discussion of the learning signals from the
discussion of the learning algorithms that leverage them. Second, we
claim that the mathematical and philosophical tools developed for the
analysis of physical systems appear very effective for the analysis of
causal information systems and of their equilibria. These two themes
are in fact a vindication of cybernetics \citep{wiener-1948}.

\vskip5ex
\section*{Acknowledgements}

We would like to acknowledge extensive discussions with Susan Athey,
Miroslav Dud\'{\i}k, Patrick Jordan, John Langford, Lihong Li,
Sebastien Lahaie, Shie Mannor, Chris Meek, Alex Slivkins, and Paul
Viola.  We also thank the Microsoft adCenter RnR team for giving us
the invaluable opportunity to deploy these ideas at scale and prove
their worth.  Finally we gratefully acknowledge the precious comments
of our JMLR editor and reviewers.


\vskip6ex

\section*{Appendices}
\addcontentsline{toc}{section}{Appendices}
\setcounter{subsection}{0}
\def\thesection{Appendix}
\def\thesubsection{\Alph{subsection}}


\bigskip
\subsection{Greedy Ad Placement Algorithms}
\label{a-greedyadplacement}

Section~\ref{s-adplacement} describes how to select and place ads on a web
page by maximizing the total rank-score~\eqref{eq-maxrankscore}.  Following
{\citep{varian-2007,edelman-2007}}, we assume that the click probability
estimates are expressed as the product of a positive position term $\gamma_p$
and a positive ad term $\beta_i(x)$. The rank-scores can therefore be written
as $r_{i,p}(x) = \gamma_p b_i \beta_i(x)$.  We also assume that the policy
constraints simply state that a web page should not display more than one ad
belonging to any given advertiser. The discrete maximization problem is then
amenable to computationally efficient greedy algorithms.

Let us fix a layout $L$ and focus on the inner maximization problem.
Without loss of generality, we can renumber the positions such that
\[
   L=\{1,2,\dots N\} \quad\text{and}\quad \gamma_1\ge\gamma_2\ge\dots\ge0\,.
\]
and write the inner maximization problem as
\[
        \max_{i_1,\dots,i_N} ~ \calR_L(i_1,\dots,i_N)  
             ~=~ \sum_{p\in L} r_{i_p,p}(x) 
\]
subject to the policy constraints and reserve constraints
 $r_{i,p}(x)\ge R_p(x)$.

Let $S_i$ denote the advertiser owning ad $i$.  The set of ads is then
partitioned into subsets $\calI_s = \{i:S_i=s\}$ gathering the ads
belonging to the same advertiser $s$. The ads that maximize the product
$b_i\beta_i(x)$ within set $\calI_s$ are called the best ads for advertiser
$s$.  If the solution of the discrete maximization problem contains one ad
belonging to advertiser $s$, then it is easy to see that this ad must be one
of the best ads for advertiser $s$: were it not the case, replacing the
offending ad by one of the best ads would yield a higher $\calR_L$ without
violating any of the constraints.  It is also easy to see that one could
select any of the best ads for advertiser $s$ without changing $\calR_L$.

Let the set $\calI^*$ contain exactly one ad per advertiser,
arbitrarily chosen among the best ads for this advertiser.  
The inner maximization problem can then be simplified as:
\[
        \max_{i_1,\dots,i_N \in \calI^*} ~ \calR_L(i_1,\dots,i_N) 
             ~=~ \sum_{p\in L} \gamma_p \: b_{i_p} \: \beta_{i_p}(x) 
\]
where all the indices $i_1,\dots,i_N$ are distinct,
and subject to the reserve constraints. 

Assume that this maximization problem has a solution $i_1,\dots,i_N$,
meaning that there is a feasible ad placement solution for the layout $L$.
For $k=1\dots N$, let us define $I^*_k\subset \calI^*$ as
\[ 
     I^*_k = \ArgMax_{i\in \calI^* \setminus \{ i_1,\dots,i_{k-1} \}}
                  ~~~ b_i \beta_i(x) \,.
\]
It is easy to see that $I^*_k$ intersects $\{i_{k},\dots,i_N\}$ because, were
it not the case, replacing $i_k$ by any element of $I^*_k$ would increase
$\calR_L$ without violating any of the constraints.  Furthermore it is easy
to see that $i_k\in I^*_k$ because, were it not the case, there would be $h>k$
such that $i_h\in I^*_k$, and swapping $i_k$ and $i_h$ would increase 
$\calR_L$ without violating any of the constraints. 

Therefore, if the inner maximization problem admits a solution, we can compute
a solution by recursively picking $i_1,\dots,i_N$ from $I^*_1, I^*_2, \dots,
I^*_N$.  This can be done efficiently by first sorting the $b_i \beta_i(x)$ in
decreasing order, and then greedily assigning ads to the best positions
subject to the reserve constraints. This operation has to be repeated for all
possible layouts, including of course the empty layout.

The same analysis can be carried out for click prediction estimates expressed
as arbitrary monotone combination of a position term $\gamma_p(x)$ and an ad
term $\beta_i(x)$, as shown, for instance, by \citet{graepel-2010}.


\bigskip
\subsection{Confidence Intervals}
\label{a-confidence}

Section~\ref{s-confidence} explains how to obtain improved confidence
intervals by replacing the unbiased importance sampling
estimator~\eqref{eq-naive-mc} by the clipped importance sampling
estimator~\eqref{eq-capped-mc}.  This appendix provides details 
that could have obscured the main message.

\subsubsection{Outer confidence interval}
\label{a-outer-confidence}

We first address the computation of the outer confidence
interval~\eqref{eq-conf-outer} which describes how the 
estimator $\estY^*$ approaches the clipped expectation $\capY^*$.
\[
   \capY^* = \int_\omega \ell(\omega)\,\capw(\omega) ~\Pa{\omega} ~~\approx~~ 
   \estY^* = \frac{1}{n} \sum_{i=1}^{n} \ell(\omega_i)\,\capw(\omega_i) \,.
\]
Since the samples $\ell(\omega_i)\,\capw(\omega_i)$ are independent and
identically distributed, the central limit theorem
\citep[\eg.,][section~17.4]{cramer-1946} states that the empirical average
$\estY^*$ converges in law to a normal distribution of mean
$\capY^*=\bbbe[\ell(\omega)\,\capw(\omega)]$ and variance
$\capV=\var[\ell(\omega)\,\capw(\omega)]$.  
Since this convergence usually
occurs quickly, it is widely accepted to write
\[
  \bbbP\inBrace{ ~ \estY^*-\epsilon_R \le \capY^* \le 
                 \estY^*+\epsilon_R ~ } \ge 1-\delta  \,,
\]     
with 
\begin{equation}
\label{eq-outer-width-clt}
   \def\erfi{{\mathrm{erf}\raisebox{1.6ex}{\!$\scriptstyle-\!1$}\!}}
   \epsilon_R ~=~ \erfi(1-\delta) ~ \sqrt{2\,\capV} ~.
\end{equation}
and to estimate the variance $\capV$ using the 
sample variance $\widehat{V}$
\[
    \capV ~\approx~ \estV = \frac{1}{n-1} 
        \sum_{i=1}^{n} \inPar{ \ell(\omega_i)\,\capw(\omega_i) - \estY^* }^2\,.
\]
This approach works well when the ratio ceiling $R$ is relatively small.
However the presence of a few very large ratios makes the
variance estimation noisy and might slow down 
the central limit convergence.

The first remedy is to bound the variance more rigorously.
For instance, the following bound results from 
\citep[theorem~10]{maurer-pontil-2009}.
\[
   \bbbP\inBrace{ 
     \sqrt{\rule{0pt}{2.2ex}\capV} ~>~ \sqrt{\estV} ~+~
           (M-m)R \sqrt{\frac{2\log(2/\delta)}{n-1}}
          } \le \delta
\]
Combining this bound with \eqref{eq-outer-width-clt} gives a confidence
interval valid with probability greater than $1-2\delta$.  
Although this approach eliminates the potential problems related to the
variance estimation, it does not address the potentially slow 
convergence of the central limit theorem. 

The next remedy is to rely on \emph{empirical Bernstein bounds} 
to derive rigorous confidence intervals that leverage both
the sample mean and the sample variance
\citep{audibert-2007,maurer-pontil-2009}.

\begin{theorem}[Empirical Bernstein bound]
\label{thm-empirical-bernstein}
\emph{\citep[thm~4]{maurer-pontil-2009}}\\
Let $X,X_1,X_2,\dots,X_n$ be \mbox{i.i.d.} random variable
with values in $[a,b]$ and let $\delta>0$.
Then, with probability at least $1-\delta$,
\[
   \bbbe[X] - M_n ~\le~
      \sqrt{\frac{2 \,V_n \log(2/\delta)}{n}}
      + (b-a) \, \frac{7 \log(2/\delta)}{3(n-1)} ~,
\]
where $M_n$ and $V_n$ respectively are 
the sample mean and variance
\[
   M_n =\frac{1}{n}\sum_{i=1}^{n} X_i ~,\qquad
   V_n =\frac{1}{n-1}\sum_{i=1}^{n} (X_i-M_n)^2 ~.
\]
\end{theorem}
\medskip

Applying this theorem to both $\ell(\omega_i)\,\capw(\omega_i)$ 
and $-\ell(\omega_i)\,\capw(\omega_i)$ provides
confidence intervals that hold for for the worst possible distribution 
of the variables $\ell(\omega)$ and $\capw(\omega)$.
\[
  \bbbP\inBrace{~ \estY^*-\epsilon_R \le \capY^* \le 
                 \estY^*+\epsilon_R ~ } \ge 1-2\delta  
\]
where
\begin{equation}
\label{eq-outer-width-bernstein}
   \epsilon_R ~=~ 
      \sqrt{\frac{2\,\estV\log(2/\delta)}{n}}
      + M\,R \, \frac{7 \log(2/\delta)}{3(n-1)} .
\end{equation}

Because they hold for the worst possible distribution,
confidence intervals obtained in this way are less tight than
confidence intervals based on the central limit theorem.
On the other hand, thanks to the Bernstein bound, 
they remains reasonably competitive, and they provide
a much stronger guarantee.

\subsubsection{Inner confidence interval}
\label{a-inner-confidence}

Inner confidence intervals are derived from 
inequality~\eqref{eq-bias-bound} which bounds 
the difference between the counterfactual expectation~$Y^*$ and
the clipped expectation~$\capY^*$\::
\[
  0 ~\le~ Y^*-\capY^* \:\le~ M\,\inPar{1-\capW^*}\,.
\]
The constant $M$ is defined by assumption~\eqref{eq-bounded-integrand}.  
The first step of the derivation
consists in obtaining a lower bound of $\capW^*-\estW^*$
using either the central limit theorem or an empirical Bernstein bound.

For instance, applying theorem~\ref{thm-empirical-bernstein}
to $-\capw(\omega_i)$ yields
\[
  \bbbP\inBrace{~ \capW^* ~\ge~ \estW^* - 
         \sqrt{\frac{2 \,\estV_w \log(2/\delta)}{n}}
       - R \, \frac{7 \log(2/\delta)}{3(n-1)}} ~\ge~ 1-\delta
\] 
where $\estV_w$ is the sample variance of the clipped weights
\[
   \estV_w ~=~ \frac{1}{n-1} \sum_{i=1}^{n} 
      \inPar{ \capw(\omega_i) - \estW^* }^2\,.
\]
Replacing in inequality~\eqref{eq-bias-bound} 
gives the outer confidence interval
\[
  \bbbP\inBrace{ ~ \capY^* ~\le~ Y^* \:\le~ \capY^* + M(1-\estW^*+\xi_R) ~ }
     \ge 1-\delta \,.
\]
with
\begin{equation}
\label{eq-inner-width-bernstein}
  \xi_R ~=~ \sqrt{\frac{2 \,\estV_w \log(2/\delta)}{n}}
           + R \, \frac{7 \log(2/\delta)}{3(n-1)} ~.
\end{equation}
Note that~$1-\estW+\xi_R$ can occasionally be negative. 
This occurs in the unlucky cases where the confidence interval 
is violated, with probability smaller than~$\delta$. 

\smallskip
Putting together the inner and outer confidence intervals,
\begin{equation}
\label{eq-conf-dual-bernstein}
   \bbbP\inBrace{ ~ \estY^* - \epsilon_R \le Y^* \le
               \estY^* + M(1-\estW^*+\xi_R) + \epsilon_R ~ } \ge 1-3\delta \,,
\end{equation}
with $\epsilon_R$ and $\xi_R$ computed
as described in expressions~\eqref{eq-outer-width-bernstein}
and~\eqref{eq-inner-width-bernstein}.

\bigskip
\subsection{Counterfactual Differences}
\label{a-differences}

We now seek to estimate the difference $Y^+-Y^*$ of the expectations
of a same quantity~$\ell(\omega)$ under two different counterfactual
distributions~$\P^+\inPar{\omega}$ and~$\Qa{\omega}$.  These
expectations are often affected by variables whose value is left
unchanged by the interventions under consideration. For instance,
seasonal effects can have very large effects on the number of ad
clicks.  When these variables affect both $Y^+$ and $Y^*$ in similar
ways, we can obtain substantially better confidence intervals for the
difference~$Y^+-Y^*$.

In addition to the notation~$\omega$ representing all the variables in
the structural equation model, we use notation~$\upsilon$ to
represent all the variables that are not direct or indirect
effects of variables affected by the interventions under
consideration.

Let $\zeta(\upsilon)$ be a known function believed to be a good predictor of
the quantity~$\ell(\omega)$ whose counterfactual expectation is sought.
Since $\Qa{\upsilon}=\Pa{\upsilon}$, the following equality holds
regardless of the quality of this prediction: 
\begin{eqnarray}
\label{eq-invariant-predictor}
Y^* ~=~ \int_\omega \ell(\omega)~\Qa{\omega} 
    &=& \int_\upsilon \zeta(\upsilon)~ \Qa{\upsilon}
    ~+\: \int_\omega \big[\ell(\omega)-\zeta(\upsilon)\big]
                   ~\Qa{\omega} \nonumber\\
    &=& \int_\upsilon \zeta(\upsilon)~ \Pa{\upsilon}
    ~+\: \int_\omega \big[\ell(\omega)-\zeta(\upsilon)\big]
                   ~w(\omega)~\Pa{\omega}~.
\end{eqnarray}
Decomposing both~$Y^+$ and~$Y^*$ in this way and computing the difference,
\[
    Y^+\!\!-Y^* 
    ~=~ \int_\omega ~ [\ell(\omega)-\zeta(\upsilon)] \, 
           \Delta w(\omega) ~ \Pa{\omega}
    ~\approx~ \frac{1}{n} \sum_{i=1}^{n} \:
           \big[\, \ell(\omega_i) - \zeta(\upsilon_i) \,\big] \, 
           \Delta w(\omega_i) \,,
\]
\begin{equation}
\label{eq-difference}
 \text{with}\qquad  \Delta w(\omega) 
              ~=~ \frac{\P^+\inPar{\omega}}{\Pa{\omega}} - 
                  \frac{\Qa{\omega}}{\Pa{\omega}}
              ~=~\frac{\P^+\inPar{\omega}-\Qa{\omega}}{\Pa{\omega}}\,. 
\end{equation}

\medskip
The outer confidence interval size is reduced if the variance of the
residual~\mbox{$\ell(\omega)-\zeta(\upsilon)$} is smaller than the variance of
the original variable~$\ell(\omega)$.  For instance, a suitable predictor
function $\zeta(\upsilon)$ can significantly capture the seasonal click yield
variations regardless of the interventions under consideration.
Even a constant predictor function can considerably change the variance of the
outer confidence interval. Therefore, in the absence of better predictor, we
still can (\,and always should\,) center the integrand using a constant
predictor.

\medskip

\begingroup
  \def\u{{\upsilon}}
  \def\ou{{\omega{\scriptstyle\setminus}\upsilon}}
  \def\ouc{{\,\ou\,|\u}}
  \def\bhi{{\cal B_{\rm hi}}}
  \def\blo{{\cal B_{\rm lo}}}
  \def\ebhi{{\widehat{\cal B}_{\rm hi}}}
  \def\eblo{{\widehat{\cal B}_{\rm lo}}}
  \def\evhi{{\widehat{V}_{\rm hi}}}
  \def\evlo{{\widehat{V}_{\rm lo}}}

The rest of this appendix describes how to construct confidence intervals for
the estimation of counterfactual differences.  Additional bookkeeping is
required because both the weights $\Delta w(\omega_i)$ and the
integrand \mbox{$\ell(\omega)-\zeta(\upsilon)$} can be positive or negative.
We use the notation $\upsilon$ to represent the variables of the structural
equation model that are left unchanged by the intervention under
considerations.  Such variables satisfy the relations $\Qa{\u}=\Pa{\u}$ and
$\Qa{\omega}=\Qa{\ouc}\,\Pa{\u}$, where we use notation $\ou$ to denote all
remaining variables in the structural equation model. An invariant predictor
is then a function $\zeta(\u)$ that is believed to be a good predictor of
$\ell(\omega)$.  In particular, it is expected that
$\var[\ell(\omega)-\zeta(\u)]$ is smaller than $\var[\ell(\omega)]$.

\subsubsection{Inner confidence interval with dependent bounds}
\label{a-upsilon-dependent-bounds}

We first describe how to construct finer inner confidence intervals by using
more refined bounds on $\ell(\omega)$.  In particular, instead of the simple
bound~\eqref{eq-bounded-integrand}, we can use bounds that 
depend on invariant variables:
\[
   \forall\omega \qquad  
     m \le m(\upsilon) \le \ell(\omega) \le M(\upsilon) \le M \,.
\]
The key observation is the equality
\[
  \bbbe\inBrack{w^*(\omega)|\u} ~=~ \int_\ou w^*(\omega)~\P(\ouc)
  ~=~ \int_\ou \frac{P^*(\ouc)\,P(\u)}{P(\ouc)\,P(\u)} ~ P(\ouc)
  ~=~ 1\,.
\]
We can then write
\begin{eqnarray*}
  Y^*-\capY^* 
     &=& \int_\omega \big[w^*(\omega)-\capw^*(\omega)\big]
             ~ \ell(\omega) ~ \Pa{\omega} 
     ~\le~ \int_\upsilon \bbbe\inBrack{\,w^*(\omega)-\capw^*(\omega)\,|\,\u\,}
            ~ M(\upsilon) ~ \P(\u) \\
     &=& \int_\upsilon \inPar{\,1-\bbbe\inBrack{\capw^*(\omega)|\u}\,}
            ~ M(\upsilon)~ \P(\u) 
     ~=~ \int_\omega \inPar{\,1-\capw^*(\omega)\,} 
            ~ M(\upsilon) ~ \P(\omega) 
     ~=~ \bhi ~.
\end{eqnarray*}
Using a similar derivation for the lower bound $\blo$, 
we obtain the inequality
\[
     \blo ~\le~ Y^*-\capY^* \:\le ~\bhi
\]
With the notations
{\small\[
 \def\ds{\displaystyle}
 \begin{array}{cc}
   \ds\eblo = \frac{1}{n}\sum_{i=1}^{n} (1-\capw^*(\omega_i))\,m(\u_i)~, &
   \ds\ebhi = \frac{1}{n}\sum_{i=1}^{n} (1-\capw^*(\omega_i))\,M(\u_i)~,\\[3ex]
   \ds\evlo = \frac{1}{n\!-\!1}\sum_{i=1}^{n} 
     \inBrack{\,(1\!-\!\capw^*(\omega_i))\,m(\u_i)-\eblo\,}^2~,  & 
   \ds\evhi = \frac{1}{n\!-\!1}\sum_{i=1}^{n} 
     \inBrack{\,(1\!-\!\capw^*(\omega_i))\,M(\u_i)-\ebhi\,}^2~, \\[3.3ex]
   \ds\xi_{\rm lo} ~=~
        \sqrt{\frac{2 \,\evlo\log(2/\delta)}{n}}
         + |m| R \, \frac{7 \log(2/\delta)}{3(n-1)} ~, &
   \ds\xi_{\rm hi} ~=~
        \sqrt{\frac{2 \,\evhi\log(2/\delta)}{n}}
         + |M| R \, \frac{7 \log(2/\delta)}{3(n-1)} ~,
 \end{array}
\]}\relax
two applications of theorem~\ref{thm-empirical-bernstein}
give the inner confidence interval:
\[
   \bbbP\inBrace{~
      \capY^* + \eblo - \xi_{\rm lo} ~\le~ Y^* ~\le~ 
      \capY^* + \ebhi + \xi_{\rm hi} ~} ~\ge~ 1-2\delta~.
\]

\subsubsection{Confidence Intervals for Counterfactual Differences}
\label{a-bounds-for-differences}

We now describe how to leverage invariant predictors in order to
construct tighter confidence intervals for the difference 
of two counterfactual expectations.
\[
    Y^+-Y^* ~\approx~ \frac{1}{n} \sum_{i=1}^{n} \:
    \big[\, \ell(\omega_i) - \zeta(\upsilon_i) \,\big] \, \Delta w(\omega_i)
    \quad\text{~~with~~}  \Delta w(\omega) =  
      \frac{\P^+\inPar{\omega}-\Qa{\omega}}{\Pa{\omega}}\,.
\]
Let us define the reweigthing ratios
$w^+(\omega)={\P^+(\omega)}/{\Pa{\omega}}$ and
$w^*(\omega)={\P^*(\omega)}/{\Pa{\omega}}$,
their clipped variants~$\capw^+(\omega)$ and~$\capw^*(\omega)$,
and the clipped centered expectations 
\[
   \capY^{+}_c = \int_\omega [\ell(\omega)-\zeta(\upsilon)]
                       \,\capw^+(\omega)\,\Pa{\omega} 
   \quad\text{and}\quad
   \capY^{*}_c = \int_\omega [\ell(\omega)-\zeta(\upsilon)]
                       \,\capw^*(\omega)\,\Pa{\omega} \,.
\]
\medskip
The outer confidence interval is obtained by applying the
techniques of section~\ref{a-outer-confidence} to
\[
   \capY^+_c-\capY^*_c ~=~ \int_\omega ~
    [\:\ell(\omega)-\zeta(\upsilon)\:]\,[\:\capw^+(\omega)-\capw^*(\omega)\:] 
    ~ \Pa{\omega} \,.
\]

Since the weights $\capw^+-\capw^*$ can be positive or negative, adding or
removing a constant to~$\ell(\omega)$ can considerably change the variance of
the outer confidence interval. This means that one should \emph{always} use a
predictor. Even a \emph{constant predictor} can vastly improve the outer
confidence interval difference. 

\medskip

The inner confidence interval is then obtained by writing the difference
\begin{eqnarray*}
  \inPar{Y^+\!-Y^*} - \inPar{\capY^+_c\!-\capY^*_c}
  &=& \int_\omega \big[\,\ell(\omega)-\zeta(\upsilon)\,\big]~
                 \big[w^+(\omega)-\capw^+(\omega)\big]\:\Pa{\omega} \\
  &-& \int_\omega \big[\,\ell(\omega)-\zeta(\upsilon)\,\big]~
                 \big[w^*(\omega)-\capw^*(\omega)\big]\:\Pa{\omega}
\end{eqnarray*}
and bounding both terms by leveraging $\upsilon$--dependent bounds
on the integrand:
\[
   \forall\omega\qquad  -M ~\le\: -\zeta(\upsilon) ~\le~ 
     \ell(\omega)-\zeta(\upsilon) ~\le~ M-\zeta(\upsilon) ~\le~ M ~.
\]
This can be achieved as shown in section~\ref{a-upsilon-dependent-bounds}.


\bigskip
\subsection{Counterfactual Derivatives}
\label{a-derivatives}

We now consider interventions that depend on a continuous parameter $\theta$.
For instance, we might want to know what the performance of the ad placement
engine would have been if we had used a parametrized scoring model.  
Let $\P^\theta\inPar{\omega}$ represent the counterfactual Markov
factorization associated with this intervention.  Let~$Y^\theta$ be the
counterfactual expectation of $\ell(\omega)$ under distribution $\P^\theta$.

Computing the derivative of~\eqref{eq-invariant-predictor} 
immediately gives
\[
   \pd{Y^\theta}{\theta}
     ~=~ \int_w ~ \big[\, \ell(\omega) - \zeta(\upsilon) \,\big] 
           \:  w^\prime_\theta(\omega) ~ \Pa{\omega} 
     ~\approx~  \frac{1}{n} \sum_{i=1}^{n} ~ 
                  \big[\, \ell(\omega_i) - \zeta(\upsilon_i) \,\big] 
           \:  w^\prime_\theta(\omega_i) 
\]
\begin{equation}
\label{eq-counter-gradient}
   \text{with}\quad
     w_\theta(\omega) = \frac{\P^\theta(\omega)}{\Pa{\omega}} 
   \qquad\text{and}\quad
     w^\prime_\theta(\omega) 
      = \pd{w_\theta(\omega)}{\theta} 
      = w_\theta(\omega) \: \pd{\log\P^\theta(\omega)}{\theta} ~.
\end{equation}

Replacing the expressions $\Pa{\omega}$ and $\P^\theta(\omega)$ by the
corresponding Markov factorizations gives many opportunities to
simplify the reweighting ratio $w^\prime_\theta(\omega)$. The term
$w_\theta(\omega)$ simplifies as shown in~\eqref{eq-general-weights}.
The derivative of $\log\P^\theta(\omega)$ depends only on the factors
parametrized by~$\theta$. Therefore, in order to evaluate
$w^\prime_\theta(\omega)$, we only need to know the few factors
affected by the intervention.

Higher order derivatives can be estimated using the same approach.
For instance,
\[
   \pdd{Y^\theta}{\theta_i}{\theta_j}
     ~=~ \int_w ~ \big[\, \ell(\omega) - \zeta(\upsilon) \,\big] 
           \:  w^\pprime_{ij}(\omega) ~ \Pa{\omega} 
     ~\approx~  \frac{1}{n} \sum_{i=1}^{n} ~ 
                  \big[\, \ell(\omega_i) - \zeta(\upsilon_i) \,\big] 
           \:  w^\pprime_{ij}(\omega_i) 
\]
\begin{equation}
\label{eq-counter-hessian}
   \text{with}\quad
     w^\pprime_{ij}(\omega) 
      = \pdd{w_\theta(\omega)}{\theta_i}{\theta_j}
      = w_\theta(\omega) 
         \:\pd{\log\P^\theta(\omega)}{\theta_i}
         \:\pd{\log\P^\theta(\omega)}{\theta_j}
          + w_\theta(\omega) 
         \:\pdd{\log\P^\theta(\omega)}{\theta_i}{\theta_j} ~.
\end{equation}
The second term in $w^\pprime_{ij}(\omega)$ vanishes when $\theta_i$ and
$\theta_j$ parametrize distinct factors in 
$\P^\theta(\omega)$.

\subsubsection{Infinitesimal Interventions and Policy Gradient}
\label{s-policy-gradient}

Expression~\eqref{eq-counter-gradient} becomes particularly attractive
when $\Pa{\omega}=\P^\theta(\omega)$, that is, when one seeks
derivatives that describe the effect of an infinitesimal intervention
on the system from which the data was collected. The resulting
expression is then identical to the celebrated \emph{policy
gradient} \citep{aleksandrov-1968,glynn-1987,williams-1992} which
expresses how the accumulated rewards in a reinforcement learning
problem are affected by small changes of the parameters of the policy
function.
\[
    \pd{Y^\theta}{\theta}
       ~=~ \int_\omega \big[\,\ell(\omega)-\zeta(\upsilon)\,\big] 
               \: w^\prime_{\theta}(\omega) \:  \P^\theta(\omega)
        ~\approx~ \frac{1}{n} \sum_{i=1}^{n} 
               \big[\,\ell(\omega_i)-\zeta(\upsilon_i)\,\big] 
               \: w^\prime_{\theta}(\omega_i)
\]
\begin{equation}
\label{eq-policy-gradient}
   \text{where $\omega_i$ are sampled i.i.d. from $\P^\theta$ and~}
      w^\prime_{\theta}(\omega) ~=~ \pd{\log \P^\theta(\omega)}{\theta}\,.
\end{equation}

Sampling from $\P^\theta(\omega)$ eliminates the potentially large
ratio~$w_\theta(\omega)$ that usually plagues importance sampling
approaches. Choosing a parametrized distribution that depends smoothly on
$\theta$ is then sufficient to contain the size of the
weights~$w^\prime_\theta(\omega)$.  Since the weights can be positive or
negative, centering the integrand with a prediction function~$\zeta(\upsilon)$
remains very important. Even a constant predictor~$\zeta$ can substantially
reduce the variance
\begingroup
\def\w{{w^\prime_\theta}}
\def\o{{(\omega)}}
\begin{eqnarray*}
   \var[\,(\ell(\omega)-\zeta)\,\w(\omega)\,] 
      &=& \var[\,\ell(\omega)\,\w(\omega) - \zeta\,\w(\omega)\,] \\ 
      & & \hskip-8em ~=~ \var[\ell(\omega)\,\w(\omega)] 
           - 2\,\zeta\,\cov[\,\ell(\omega)\,\w(\omega),\,\w(\omega)\,] 
           + \zeta^2\,\var[\w(\omega)] 
\end{eqnarray*}
whose minimum is reached for 
$ \displaystyle ~
  \zeta = \frac{\cov[\ell\o \w\o, \w\o]}{\var[\w\o]} 
    = \frac{\bbbe[\ell\o \w\o^2]}{\bbbe[ \w\o^2]} ~ . $
\endgroup

\medskip
We sometimes want to evaluate expectations under a counterfactual distribution
that is too far from the actual distribution to obtain reasonable confidence
intervals. Suppose, for instance, that we are unable to reliably estimate
which click yield would have been observed if we had used a certain
parameter~$\theta^*$ for the scoring models.  We still can estimate how
quickly and in which direction the click yield would have changed if we had
slightly moved the current scoring model parameters~$\theta$ in the direction
of the target~$\theta^*$.  Although such an answer is not as good as a
reliable estimate of~$Y^{\theta^*}$, it is certainly better than no answer.

\subsubsection{Off-Policy Gradient}
\label{s-off-policy-gradient}

We assume in this subsection that the parametrized probability
distribution~$\P^\theta(\omega)$ is regular enough to ensure that all
the derivatives of interest are defined and that the
event~\mbox{$\{w_\theta(\omega)\!=\!R\}$} has probability zero.
Furthermore, in order to simplify the exposition, the following
derivation does not leverage an invariant predictor function.

Estimating derivatives using data sampled from a distribution
$\Pa{\omega}$ different from $\P^\theta(\omega)$ is more challenging
because the ratios~$w_\theta(\omega_i)$ in
equation~\eqref{eq-counter-gradient} can take very large values.
However it is comparatively easy to estimate the derivatives of lower
and upper bounds using a slightly different way to clip the weights.
Using notation $\bbbone(x)$ represent the indicator function,
equal to one if condition $x$ is true and zero otherwise,
let us define respectively the clipped weights~$\capwz$ and 
the capped weights~$\capwm$:
\[
  \capwz(\omega) =
     w_\theta(\omega)\,\bbbone\{ \Qa{\omega}<R\,\Pa{\omega} \}
  \quad\text{and}\quad
  \capwm(\omega) =
     \min\{w_\theta(\omega),\,R \} ~.
\]

Although section~\ref{s-confidence} illustrates the use of clipped weights,
the confidence interval derivation can be easily extended to the capped weights.
Defining the capped quantities
\begin{equation}
\label{eq-maxcap-defs}
   \capY^\theta = \int_\omega \ell(\omega) \: \capwm(\omega) \: \Pa{\omega} 
   \quad\text{and}\quad
   \capW^\theta = \int_\omega \capwm(\omega) \: \Pa{\omega}
\end{equation}
and writing
\begin{eqnarray*}
   0 ~\le~ Y^\theta - \capY^\theta &=& 
     \int_{\omega\in\Omega\setminus\Omega_R} \hskip-1.5em 
           \ell(\omega)\,\inPar{\,\Qa{\omega} - R\,\Pa{\omega} \,} \\
                       &\le& 
     M\,\Big(\,1-\Qa{\Omega_R} - R\,\Pa{\Omega\setminus\Omega_R}\,\Big)
                       ~=~ 
     M\,\inPar{\,1-\int_\omega \capwm(\omega)\,\Pa{\omega}\,}
\end{eqnarray*}
yields the inequality
\begin{equation}
\label{eq-maxcap-bounds}
   \capY^\theta ~\le~ Y^\theta  \:\le~ 
           \capY^\theta + M (1-\capW^\theta) ~.
\end{equation}
In order to obtain reliable estimates of the derivatives of these upper and
lower bounds, it is of course sufficient to obtain reliable estimates of the
derivatives of~$\capY^\theta$ and~$\capW^\theta$.  By separately considering
the cases~$w_\theta(\omega)<R$ and~$w_\theta(\omega)>R$, 
we easily obtain the relation
\[
  \capwmp(\omega)~=~\pd{\capwm(\omega)}{\theta} 
    ~=~ \capwz(\omega)\,\pd{\log P^\theta(\omega)}{\theta} 
    \quad\text{when}~~w_\theta(\omega)\neq R
\]
and, thanks to the regularity assumptions, we can write
\begin{eqnarray*}
    \pd{\capY^\theta}{\theta} 
    &=& \int_\omega 
           \ell(\omega) \: \capwmp(\omega)\:\Pa{\omega}
    ~\approx~ \frac{1}{n} \sum_{i=1}^{n} \:
           \ell(\omega_i) \: \capwmp(\omega_i) ~,\\
    \pd{\capW^\theta}{\theta} 
    &=& \int_\omega \capwmp(\omega)\: \Pa{\omega} 
    ~\approx~ \frac{1}{n} \sum_{i=1}^{n} \:
           \: \capwmp(\omega_i) \,,
\end{eqnarray*}
Estimating these derivatives is considerably easier than using
approximation~\eqref{eq-counter-gradient} because they involve the bounded
quantity~$\capwz(\omega)$ instead of the potentially large
ratio~$w_\theta(\omega)$.  It is still necessary to choose a
sufficiently smooth sampling distribution~$\Pa{\omega}$ to limit the
magnitude of~$\partial{\log\P^\theta}\!/\partial{\theta}$.

Such derivatives are very useful to drive optimization algorithms. Assume for
instance that we want to find the parameter~$\theta$ that maximizes the
counterfactual expectation~$Y^\theta$ as illustrated in
section~\ref{s-exp-alphamlr}. Maximizing the estimate obtained using
approximation~\eqref{eq-general-reweighting} could reach its maximum for a
value of~$\theta$ that is poorly explored by the actual
distribution. Maximizing an estimate of the lower bound
\eqref{eq-maxcap-bounds} ensures that the optimization algorithm finds a
trustworthy answer.

\bigskip
\subsection{Uniform empirical Bernstein bounds}
\label{a-uniform-bernstein}

\def\calX{{\mathcal{X}}}
\def\calF{{\mathcal{F}}}
\def\calN{{\mathcal{N}}}
\def\calM{{\mathcal{M}}}
\def\bx{{\mathbf{x}}}

This appendix reviews the uniform empirical Bernstein bound
given by \citet{maurer-pontil-2009} and describes how it can
be used to construct the uniform confidence 
interval~\eqref{eq-uniform-conf-interval}.
The first step consists of characterizing the size of a family~$\calF$ of
functions mapping a space~$\mathcal{X}$ into the
interval~$[a,b]\subset\bbbr$. Given~$n$
points~$\bx=(x_1{\dots}x_n)\in\calX^n$, the
trace~$\calF(\bx)\in\bbbr^n$ is the set of vectors
$\big(f(x_1),\dots,f(x_n)\big)$ for all functions $f\in\calF$.

\begin{definition}[Covering numbers, etc.]
Given $\varepsilon>0$, the covering number~$\calN(\bx,\varepsilon,\calF)$ is
the smallest possible cardinality of a subset $C\subset\calF(\bx)$ satisfying
the condition
\[ \forall v\in\calF(\bx)~~~ \exists c\in C ~~~ 
       \max_{i=1\dots n} |v_i-c_i|\,\le\,\varepsilon~,
\]
and the growth function $\calN(n, \varepsilon,\calF)$ is
\[
   \calN(n, \varepsilon,\calF) ~=~
     \sup_{\bx\in\calX^n} \calN(\bx, \varepsilon,\calF) ~.
\]
\end{definition}
Thanks to a famous combinatorial
lemma~\citep{vapnik-chervonenkis-1968,vapnik-chervonenkis-1971,sauer-1972},
for many usual parametric families~$\calF$, the growth
function~$\calN(n,\varepsilon,\calF)$ increases at most
polynomially{\footnote{\relax For a simple proof of this fact, slice $[a,b]$
    into intervals~$S_k$ of maximal width $\varepsilon$ and apply the
    lemma to the family of indicator functions
    $(x_i,S_k)\mapsto \bbbone\inBrace{f(x_i)\in{S_k}}$.  
} with both~$n$ and~$1/\varepsilon$.

\begin{theorem}[Uniform empirical Bernstein bound]
\label{thm-uniform-bernstein}
\emph{\citep[thm~6]{maurer-pontil-2009}}\\
Let $\delta\in(0,1)$, $n>=16$.
Let~$X,X_1,\dots,X_n$ be \mbox{i.i.d.} random variables with values in~$\calX$.
Let~$\calF$ be a set of functions mapping~$\calX$ into~$[a,b]\subset\bbbr$
and let~$\calM(n)=10\,\calN(2n,\calF,1/n)$.
Then we probability at least $1-\delta$,
\[
   \forall f\in\calF, ~~ \bbbe\inBrack{f(X)} - M_n ~ \le ~
     \sqrt{\frac{18\,V_n\,\log(\calM(n)/\delta)}{n}} +
      (b-a)\frac{15\,\log(\calM(n)/\delta)}{n-1}~,
\]
where $M_n$ and $V_n$ respectively are 
the sample mean and variance
\[
   M_n =\frac{1}{n}\sum_{i=1}^{n} f(X_i) ~,\qquad
   V_n =\frac{1}{n-1}\sum_{i=1}^{n} (f(X_i)-M_n)^2 ~.
\]
\end{theorem}

\noindent
The statement of this theorem emphasizes its similarity with the
non-uniform empirical Bernstein bound
(theorem~\ref{thm-empirical-bernstein}). 
Although the constants are less attractive, the uniform bound 
still converges to zero when $n$ increases,
provided of course that $\calM(n)=10\,\calN(2n,\calF,1/n)$ grows
polynomially with~$n$.

\smallskip
Let us then define the family of functions
\[
  \calF = \inBrace{~f_\theta: \omega \mapsto \ell(\omega)\capwm(\omega)
              ~,~~ g_\theta: \omega \mapsto \capwm(\omega)
              ~,~~\forall \theta\in\mathcal{F} ~} ~,
\]
and use the uniform empirical Bernstein bound to derive an outer
inequality similar to~\eqref{eq-outer-width-bernstein} and an inner
inequality similar to~\eqref{eq-inner-width-bernstein}. The theorem
implies that, with probability $1-\delta$, both inequalities are simultaneously 
true for all values of the parameter $\theta$. The uniform confidence
interval~\eqref{eq-uniform-conf-interval} then follows directly.

\vskip6ex
\bibliography{counterfactuals}

\end{document}